%% file: main.tex
\documentclass[10pt,twocolumn,letterpaper]{article}
\usepackage[pagenumbers]{wacv} 


\definecolor{wacvblue}{rgb}{0.21,0.49,0.74}
\usepackage[pagebackref,breaklinks,colorlinks,allcolors=wacvblue]{hyperref}

\usepackage{colortbl}
\usepackage{mathtools}
\usepackage{tikz,pgfplots}
\usepackage{caption}
\usepackage[symbol]{footmisc}

\input{math_commands.tex}

\newcommand*{\stddev}[2]{\genfrac{}{}{0pt}{}{\textstyle #1}{\color{gray} \scriptscriptstyle \pm #2}}
\newcommand*{\meanrelrand}[2]{\genfrac{}{}{0pt}{}{\textstyle #1}{\color{gray} \scriptstyle #2}}
\newcommand*{\methodpub}[2]{\genfrac{}{}{0pt}{}{\textstyle #1 \hfill}{\color{gray} \scriptstyle #2 \hfill}}
\newcommand*{\methodlat}[2]{\textstyle #1  {\color{gray} \scriptstyle #2}}
\newcommand*{\xlabel}[2]{{\textstyle #1}{\color{gray} \scriptstyle #2}}
\newcommand{\stddevrow}{\rule{0pt}{3.3ex}}

\definecolor{note}{rgb}{0.7,0.1,0.1}
\definecolor{rowgray}{rgb}{0.975,0.975,0.975}

\def\wacvPaperID{555} 
\def\confName{WACV}
\def\confYear{2026}

\title{Zero-Shot Coreset Selection via Iterative Subspace Sampling}

\author{Brent A. Griffin$^\text{\scriptsize 1}$\footnotemark  \qquad Jacob Marks$^\text{\scriptsize 1}$\footnotemark \qquad Jason J. Corso$^\text{\scriptsize 1,2}$ \\
		$^\text{\scriptsize 1~}$Voxel51 \qquad $^\text{\scriptsize 2~}$University of Michigan
}

\begin{document}

\twocolumn[{
	\maketitle
	\begin{center}
		\centering
		\includegraphics[width=0.93\textwidth]{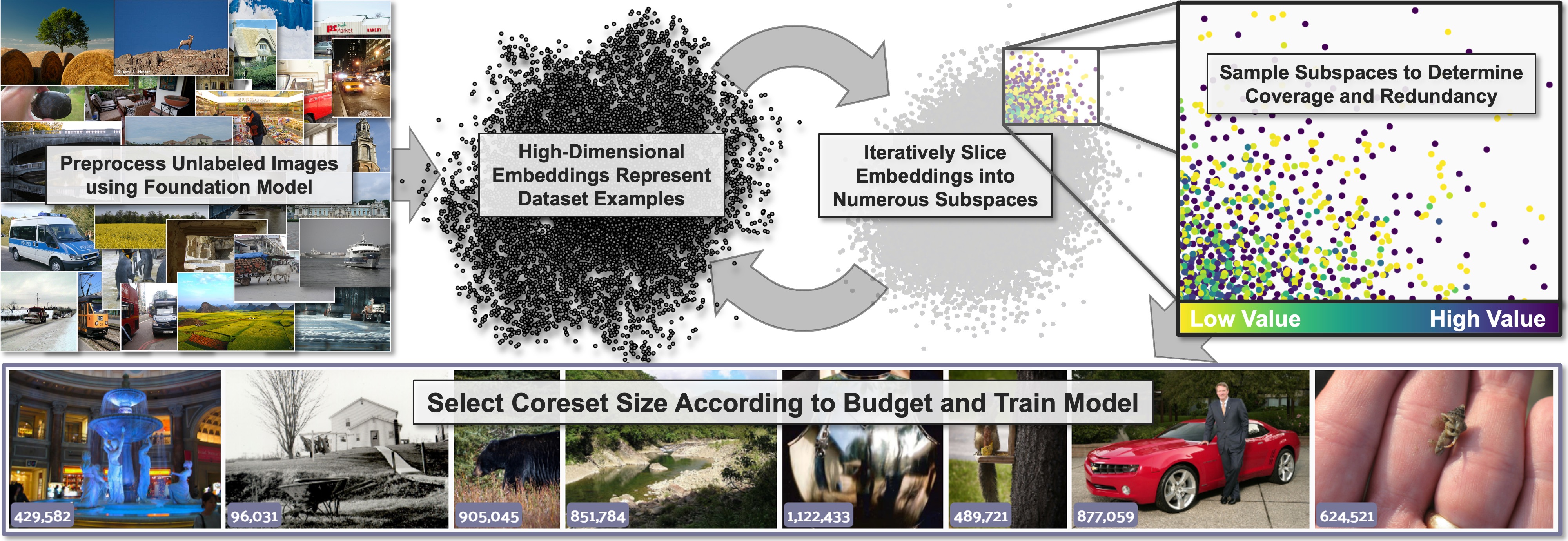}
		\vspace{-0.5em}
		\captionof{figure}{\textbf{Overview}.
			We use foundation models to generate high-dimensional embeddings for unlabeled candidate images (left).
			We then iteratively slice the full embedding space into subspaces (center), which we sample to find examples covering large regions of the resulting subspace distributions while penalizing redundancy (right).
			Finally, we output a coreset of data to train models for a given budget (bottom).}
	\label{fig:front}
\end{center}
\vspace{0.25em}
}]

\begin{abstract}
	Deep learning increasingly relies on massive data with substantial storage, annotation, and training costs.
	To reduce costs, coreset selection finds a representative subset of data to train models while ideally performing on par with the full data training.
	To maximize performance, current state-of-the-art coreset methods select data using dataset-specific ground truth labels and training.
	However, these methodological requirements prevent selection at scale on real-world, unlabeled data.
	To that end, this paper addresses the selection of coresets that achieve state-of-the-art performance but without using any labels or training on candidate data. 
	Instead, our solution, \textbf{Z}ero-Shot \textbf{Core}set Selection via Iterative Subspace Sampling (ZCore), uses previously-trained foundation models to generate zero-shot, high-dimensional embedding spaces to interpret unlabeled data.
	ZCore then iteratively quantifies the relative value of all candidate data based on coverage and redundancy in numerous subspace distributions.
	Finally, ZCore selects a coreset sized for any data budget to train downstream models.
	We evaluate ZCore on four datasets and outperform several state-of-the-art label-based methods, especially at low data rates that provide the most substantial cost reduction.\footnotetext[1]{~Corresponding email: {\tt brent@voxel51.com}.}\footnotetext[2]{~Work completed while Jacob Marks was at Voxel51.}~On ImageNet, ZCore selections for 10\% training data achieve a downstream validation accuracy of 53.99\%, which outperforms prior label-based methods and removes annotation and training costs for 1.15 million images.\footnote[3]{~Code available at \url{https://github.com/voxel51/zcore}.} 
\end{abstract}

\vspace{-1.35em}
\section{Introduction}

The computational cost to train a single state-of-the-art deep learning model in various fields doubles every 3.4 months in the deep learning era due to increasingly large models and datasets \cite{openai18, Zhao_2023_WACV}.
Since the introduction of AlexNet \cite{alexnet}, groundbreaking models in computer vision like ViT and DALL-E all rely on massive datasets for training \cite{dosovitskiy2021an,ramesh2022hierarchical}.
However, there are substantial costs to collecting, storing, transmitting, and pre-processing such a vast amount of data.
Furthermore, training models on vast datasets introduces yet another substantial cost for computation, sometimes hundreds of thousands of GPU hours to achieve satisfactory performance, which frustrates applications requiring repeat training over datasets such as hyparameter optimization \cite{pmlr-v37-maclaurin15,pmlr-v108-lorraine20a} and neural architecture search \cite{JMLR:v20:18-598,pmlr-v115-li20c}.

Increasing data efficiency mitigates these deep learning challenges.
Specifically, coreset selection reduces the training set size by selecting a pruned subset that contains only valuable examples (the \textit{core set}), such that downstream models trained on the coreset achieve similar performance to those trained on the original, full dataset \cite{NIPS2011_2b6d65b9}.

Coreset selection methods have demonstrated impressive results using median class values \cite{xia2023moderate}, diverse coverage of importance scores \cite{zheng2023coveragecentric}, and gradient dynamics during training \cite{Zhang_2024_CVPR}.
Notably, to achieve maximum performance in experiments, these and other state-of-the-art coreset methods increasingly use data labels and training prior to selection \cite{He_2024_CVPR,deepcore22}. 
However, the majority of real-world data are unlabeled. 
Moreover, labeling all candidate images to consider them for selection is cost prohibitive, with annotation taking anywhere between 7 seconds per bounding box to 1.5 hours for full semantic segmentation \cite{JaGr13,Cordts_2016_CVPR}.

\textit{Unlabeled} coreset selection uniquely reduces label costs by pruning away data prior to \textit{any} labeling.
If the downstream model is self-supervised, no labels are required.
If the downstream model is supervised, \textit{only} the coreset is labeled.
Notably, some coreset methods use self-supervised learning in place of label-based training \cite{Sorscher_NEURIPS2022,zheng2025elfs}, but this still requires substantial time and compute to make selections at scale.
Furthermore, coupling selection with training on a single model architecture decreases generalization.

To that end, this paper addresses the problem of selecting coresets that achieve state-of-the-art performance \textit{without} using labels or training, enabling low-cost data selection at scale.
Our approach, \textbf{Z}ero-Shot \textbf{Core}set Selection via Iterative Subspace Sampling (ZCore), uses previously-trained foundation models to process images in a single forward pass, generating a high-dimensional embedding space representation of all unlabeled data.
ZCore then iteratively slices the full embedding space down to numerous low-dimensional subspaces that are sampled via Triangular distribution to estimate each data example's value based on \textit{coverage} and \textit{redundancy} (see \cref{fig:front}).
Each subspace contains a unique combination of foundation model features and reduces runtime relative to full embedding space processing.
Notably, our default ZCore implementation uses 818,560 diverse subspaces and completes coreset selections for all possible prune rates on CIFAR100 in 381.6\textrm{s} using a standard laptop.
Our contributions are:
\begin{enumerate}
	\item Formalizing and contrasting labeled and unlabeled coreset selection, with the later substantially reducing data- \textit{and} label-based costs for efficient deep learning at scale.
	\item Developing the ZCore method, which selects coresets from unlabeled data with efficient computation at scale using a novel iterative subspace sampling technique. 
	\item Evaluating ZCore with the state-of-the-art using ten prior coreset methods and four datasets.
	Results show that ZCore performs best in multiple low-data rate cases and overall outperforms all label-based methods save one, without ZCore using labels or training for selection.
	\item  Evaluating ZCore under constant algorithmic settings across a dataset and coreset selection scale greater than three orders of magnitude. To our knowledge, there is no precedent for this selection range of experiments and generalization in the coreset selection literature.
\end{enumerate}
From these results, ZCore establishes a new state-of-the-art.

\section{Related Work}

\noindent {\bf Dataset Distillation} is similar to coreset selection in that it supports data-efficient deep learning.
On a functional level, the objectives of many coreset methods also apply to dataset distillation.
However, as opposed to selecting a subset of \textit{existing} data for a coreset, dataset distillation aims to generate a much smaller dataset with \textit{synthetic} examples that yield the same performance as the larger initial dataset \cite{datadistillcomp24}.
Notable dataset distillation methods generate synthetic examples relative to the initial dataset by matching gradients \cite{zhao2021dataset}, differentiable Siamese augmentation for better synthesis \cite{pmlr-v139-zhao21a}, aligning features \cite{Wang_2022_CVPR}, multi-step parameter matching \cite{Cazenavette_2022_CVPR}, and embedding space distribution matching \cite{Zhao_2023_WACV}.
These dataset distillation methods are remarkable for creating small but effective synthetic training datasets.
Alternatively, to utilize real-world data, our current work selects and evaluates coresets from already existing data.

\noindent {\bf Active Learning}
is another research area that supports data-efficient deep learning by enabling learning algorithms to choose their own data to improve performance with less training \cite{Bu12}, which is especially useful when large portions of data are unlabeled and labeling data is expensive \cite{BeEtAl18}.
In fact, active learning encompasses the particularly hard problem of starting selection without initial labeled examples, i.e., the cold start problem \cite{ColdStart98}.
Notably, some recent active learning methods focus on the importance of coverage diversity in data selection \cite{Ash2020Deep,NEURIPS2021_64254db8}.
However, these methods iteratively train and select data on an increasing set for a specific model.
Alternatively, our work selects data without initial labels for a \textit{single}, \textit{model-agnostic} training phase.

\noindent {\bf Coreset Selection}
prunes datasets down to a smaller, valuable \textit{core set} to reduce costs and enable data-efficient deep learning.
A basic solution to find the optimal coreset is to search through and train on every subset to find the best corresponding model performance. 
However, this simple approach is NP-hard, leading to the development of many alternative coreset methods.
Early coreset methods aim for a consistent data distribution to the full dataset \cite{NIPS2011_2b6d65b9,pmlr-v37-bachem15}, e.g., greedily adding one sample at a time to match embedding space centers \cite{coreherd}.
Other coreset methods can be broadly categorized as selecting by optimization \cite{pmlr-v37-wei15,yang2023dataset}, coverage or diversity \cite{sener2018active,zheng2023coveragecentric,NEURIPS2023_04768210}, semantic redundancy \cite{abbas2023semdedup,Slyman_2024_CVPR}, and importance criteria \cite{toneva2018an,NEURIPS2023_3abe23bf}.
Recent coreset innovations address ongoing challenges like application on a wide range of dataset sizes \cite{xia2023moderate}, making selections on data with label errors \cite{NEURIPS2023_ebb6bee5}, and best use of full training dynamics \cite{Zhang_2024_CVPR,zheng2025elfs}.

\begin{table}
	\setlength{\tabcolsep}{0.5pt}
	\small
	\footnotesize
	\centering
	\caption{Comparison of {\bf Data and Procedural Requirements}.}
	\label{tab:rel}
	\vspace{-1.25em}
	\begin{tabular}{l c  c c c c c}
		\noalign{\global\arrayrulewidth=0.4mm} \hline \noalign{\global\arrayrulewidth=0.2mm}
		\noalign{\global\arrayrulewidth=0.4mm} \arrayrulecolor{white}\hline \noalign{\global\arrayrulewidth=0.2mm} \arrayrulecolor{gray}
		\multicolumn{1}{r}{Selects Coreset \textit{Without}}: & &   Training   &&   Ground  & &  Prune   \\
		& &   on & & Truth &  & Specific  \\
		\multicolumn{1}{c}{State-of-the-art Coreset Methods} & &   Data & &  Labels &  &  Tuning \\
		\noalign{\global\arrayrulewidth=0.4mm} \arrayrulecolor{white}\hline \noalign{\global\arrayrulewidth=0.2mm} \arrayrulecolor{gray}
		\cline{1-1} \cline{3-3} \cline{5-5}  \cline{7-7}
		{\bf ZCore (ours)} & & \bf Yes & & \bf Yes & & \bf Yes \rule{0pt}{2ex} \\ 
		\cline{1-1} \cline{3-3} \cline{5-5}  \cline{7-7}
		$\methodlat{\text{Prototypes$_{\text{\tiny SS}}$ \cite{Sorscher_NEURIPS2022}}}{\text{\tiny ~NeurIPS 2022}}$ &  & No & & \bf Yes & & \bf Yes \rule{0pt}{2ex} \\
		\cline{1-1} \cline{3-3} \cline{5-5}  \cline{7-7}
		$\methodlat{\text{ELFS \cite{zheng2025elfs}}}{\text{\tiny ~ICLR 2025}}$ &  & No & & \bf Yes & & No \rule{0pt}{2ex} \\
		\cline{1-1} \cline{3-3} \cline{5-5}  \cline{7-7}
		$\methodlat{\text{Dyn-Unc \cite{He_2024_CVPR}}}{\text{\tiny ~CVPR WS `24}}$,	$\methodlat{\text{Moderate \cite{xia2023moderate}}}{\text{\tiny ~ICLR 2023}}$&  & No & & No & & \bf Yes \rule{0pt}{2ex} \\
		\cline{1-1} \cline{3-3} \cline{5-5}  \cline{7-7}
		$\methodlat{\text{TDDS \cite{Zhang_2024_CVPR}}}{\text{\tiny ~CVPR 2024}}$,  $\methodlat{\text{Coverage \cite{zheng2023coveragecentric}}}{\text{\tiny ~ICLR 2023}}$ & & No & & No & &No\rule{0pt}{2ex} \\
		\noalign{\global\arrayrulewidth=0.4mm} \arrayrulecolor{white}\hline \noalign{\global\arrayrulewidth=0.2mm}
		\noalign{\global\arrayrulewidth=0.4mm} \arrayrulecolor{black}\hline \noalign{\global\arrayrulewidth=0.2mm}
	\end{tabular}
\end{table}

\begin{figure}
	\centering
	\includegraphics[width=0.42\textwidth]{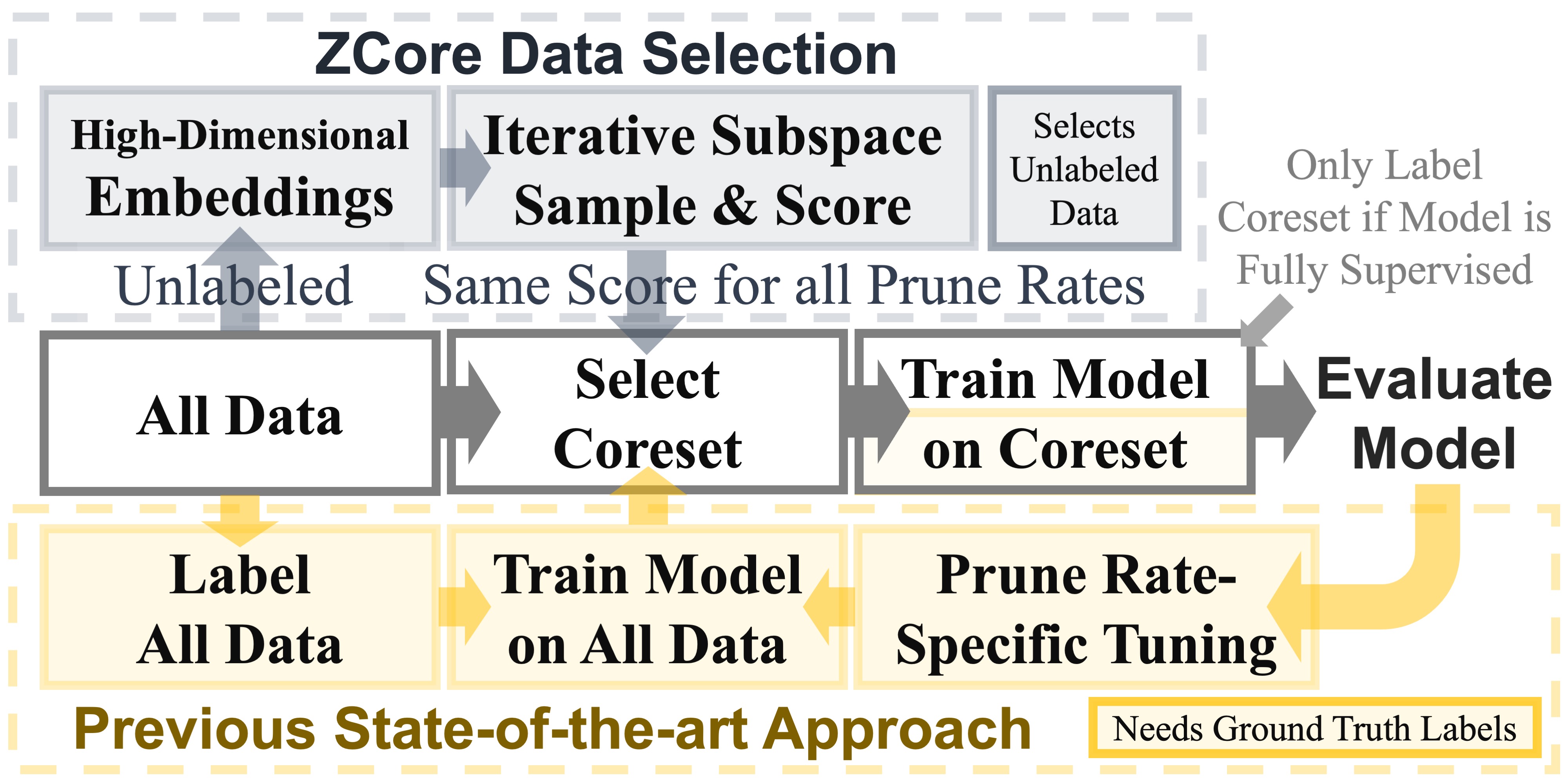}
	\vspace{-0.9em}
	\captionof{figure}{\textbf{Coreset and Model Train Workflow Comparison}.}
	\label{fig:overview}
\end{figure}

Our current work is inspired by the success of previous coreset selection methods.
Importantly, recent state-of-the-art methods have increasingly improved coreset performance (especially at low data rates) through the use of labels and/or training on the larger initial dataset (see \cref{tab:rel}, \cref{fig:overview}).
Instead, our current work maintains or improves state-of-the-art coreset selection performance using only unlabeled data and no dataset- or architecture-specific training.
Fundamentally, this coreset approach broadens applicability to new data and models while reducing label error sensitivity, labeling costs, and computation at scale.

\section{Problem Formulation}

We first define the problem of \textit{labeled} coreset selection. 
Formally, we are given a labeled dataset $\sS^\text{\tiny L} = \{(\rvx_i, y_i)\}^N_{i=1}$ with $N$ examples drawn i.i.d.~from an underlying distribution $P$, where $\rvx_i$ are the data and $y_i$ is the ground truth label for each example. 
The goal is to select a subset of $\sS^\text{\tiny L}$ to reduce future storage and training consumption while closely maintaining the performance of full dataset training. We denote this \textit{coreset} as $\sS^\text{\tiny C} = \{(\rvx_i, y_i)\}^n_{i=1} \subset \sS^\text{\tiny L}$, which has $n$ examples and a \textit{prune rate} of $1-\frac{n}{N}$. We formulate coreset selection as  \cite{sener2018active}:
\begin{align}
	\argmin_{\sS^{\text C} \subset \sS^\text{L} |1- \frac{n}{N} \geq p} \E_{\rvx, y \sim P} [ l(\rvx, y; f_{(\sS^\text{C})}) ],
	\label{eq:core}
\end{align}
where $p$ is a prune rate set \textit{before} training, $l$ is the loss function, and $f_{(\sS^\text{C})}$ is a model trained on $\sS^\text{\tiny C}$.
Notably, many SOTA methods select $\sS^\text{\tiny C}$ by assigning an importance score to each example \cite{Zhang_2024_CVPR, zheng2025elfs}, which we denote as  $\vs \in \sR^N$.

We now define the problem of \textit{unlabeled} coreset selection for data- and \textit{label}-efficient deep learning.
Formally, given an unlabeled dataset $\sS = \{(\rvx_i)\}^N_{i=1}$, the goal is to select $\sS^\text{\tiny C} \subset \sS$ without using \textit{any} ground-truth label $y_i$.
The motivation for this change is that it is preventative to label an entire massive dataset when much of the data will be pruned.
We formulate unlabeled coreset selection by replacing $\sS^\text{\tiny C} \subset \sS^\text{\tiny L}$ with $\sS^\text{\tiny C} \subset \sS$ in \cref{eq:core}.
Notably, after selecting  $\sS^\text{\tiny C}$, \textit{if} the downstream model $f_{(\sS^\text{C})}$ is supervised, we use only $n$ labels for the coreset as $\sS^\text{\tiny C} = \{(\rvx_i, y_i)\}^n_{i=1}$. 

Along with the aforementioned benefits of coreset selection, unlabeled coreset selection uniquely increases scale and reduces labeling costs.
First, while we can use any $\rvx_i$ from a labeled dataset $\sS^\text{\tiny L}$, we can also extensibly sample and consider more examples $\rvx$ from the underlying distribution $P$ without any annotation or labeling requirements. 
This extension enables us to source coresets from a much larger initial dataset.
In effect, unlabeled coreset selection extends selection to the majority of unlabeled, real-world data.
Second, for supervised model training, we only label the $n$ coreset examples after they are selected, so there is a $N-n$ reduction in labeling costs relative to label-based selection.
On ImageNet, unlabeled selection at a 90\% prune rate removes label requirements for 1.15 million images.

Finally, we likewise note that selecting examples without training on candidate data substantially improves runtime.

\section{Zero-Shot Coreset Selection via \\ Iterative Subspace Sampling}
\label{sec:method}

We use the unlabeled coreset selection formulation to develop a new method of \textbf{Z}ero-Shot \textbf{Core}set Selection via Iterative Subspace Sampling (ZCore).
In place of label- or training-based selection, ZCore first uses foundation models to generate a zero-shot, high-dimensional embedding space representation of the full dataset (\cref{sec:embed}). 
ZCore then samples varying, lower-dimensional embedding subspaces to efficiently determine which examples provide valuable coverage across unique feature combinations (\cref{sec:cover}).
Subsequently, ZCore determines which examples in proximity to those providing coverage are redundant (\cref{sec:red}).
Finally, ZCore uses the coverage and redundancy metrics to iteratively sample and score each candidate training example to determine final coreset selections (\cref{sec:proc}).

\subsection{Foundational Embedding Representation}
\label{sec:embed}

ZCore selects data using a zero-shot, high-dimensional embedding space representation of unlabeled dataset~$\sS$.
We generate embeddings using a foundation model denoted as $f(\cdot) = g(h(\cdot))$, where $h$ is the model component mapping input data to hidden representations at the penultimate layer and $g$ maps the embedding space to a previously learned output $f$.
We use $h(\rvx_i)  \in \sR^M$ to generate an $M$-dimension \textit{embedding space} for input data $\sS = \{(\rvx_i)\}^N_{i=1}$ denoted as
\begin{align}
	\mZ = [h(\rvx_1), \cdots, h(\rvx_N)] \in \sR^{N \times M}.
	\label{eq:Z}
\end{align}

Embedding space $\mZ$ lets us use the previously learned hidden representation of $h$ as a zero-shot alternative to label- or training-based coreset selection.
In general, examples corresponding to different classes or concepts map to different regions of the feature-based embedding space.
Thus, we quantify the importance of each example in terms of relative coverage and redundancy in $\mZ$ as a representation of the underlying data distribution $\rvx, y \sim P$ in \cref{eq:core}.

\noindent {\bf {\small Remarks on} $\mZ$}: In \cref{sec:eval} experiments, we generate all model embeddings in advance using off-the-shelf weights for a ResNet18 \cite{HeEtAl16} and CLIP ViT-L-14 model \cite{clip}, which we concatenate as $h(\rvx_i) = \big[\genfrac{}{}{0pt}{}{h^\text{RN18}(\rvx_i)}{h^\text{CLIP}(\rvx_i)}\big] \in \sR^{\text{1,280}}$.
As we will show in \cref{sec:eff}, relative to coreset methods using full dataset training for 60-200 epochs, generating embeddings takes much less time as it only requires one forward pass per sample, a subcomponent ($h$) of the full model architecture, and no training-based back propagation or metric tracking. 

\begin{figure}[t!]
	\centering
	\begin{minipage}{0.25\textwidth}
		\input{resnet18_0_paper.tex}
	\end{minipage}%
	\begin{minipage}{0.25\textwidth}
		\input{open_0_paper.tex}
		\vspace{1.175em}
	\end{minipage}%
	\vspace{-2em}
	\caption{{\bf Comparison of embeddings and sampling techniques}. ResNet18 (left) and CLIP (right) are the first-dimension model embeddings for 50,000 CIFAR100 train set examples, while each corresponding distribution type is sampled 50,000 times.
	}
	\label{fig:dist}
\end{figure}

\subsection{Coverage of Embedding Subspaces}
\label{sec:cover}

Our first objective for coreset selection is to select examples that maximize coverage of embedding space $\mZ$.
To quantify coverage, we develop a Monte Carlo-inspired sampling technique \cite{Ulam_49}, which estimates the relative contribution of each candidate training example $\rvx_i \in \sS$ in covering a carefully-designed distribution over the embedding space.

We assume a Triangular distribution over each embedding space dimension $j \in \{1,\cdots,M\}$ using 
\begin{equation}
	\begin{aligned}
			\rz_j \sim p(\rx, j) \coloneq \qquad \qquad \qquad \qquad \qquad \qquad \qquad ~~~~~ \\ \begin{cases}
				\frac{ 2(\rx - \evz^\text{min}_j) }{(\evz^\text{max}_j - \evz^\text{min}_j)(\evz^\text{med}_j - \evz^\text{min}_j)} & \text{\small for}~ \evz^\text{\tiny min}_j \leq \rx < \evz^\text{\tiny med}_j \\[5pt]
				\frac{ 2(\evz^\text{max}_j - \rx) }{(\evz^\text{max}_j - \evz^\text{min}_j)(\evz^\text{max}_j - \evz^\text{med}_j)} & \text{\small for}~ \evz^\text{\tiny med}_j \leq \rx \leq \evz^\text{\tiny max}_j 
			\end{cases},
			\\
			\rvz \coloneq [\rz_1, \cdots, \rz_M]^\intercal \in \sR^M, 
		\end{aligned}
		\label{eq:s}
	\end{equation}
	where $\rvz$ is a full random sample from the distribution of $\mZ$ across all $M$ dimensions, $\vz^\text{\tiny min} =\{\text{min}(\mZ _{:,j})\}^M_{j=1} \in \sR^M$ is the minimum $\mZ$ value for each embedding dimension, and $\vz^\text{\tiny med}, \vz^\text{\tiny max} \in \sR^M$ are the corresponding median and maximum $\mZ$ values.
	In practice, Triangular distribution \cref{eq:s} robustly covers both exponential- (ResNet) and Gaussian-shaped (CLIP) embedding distributions (see \cref{fig:dist}). 
	Relative to uniform or Gaussian, the Triangular distribution uniquely achieves all objectives of: providing sufficient coverage for densely populated regions of the embedding space, covering outliers, and not sampling empty space.
	
	We increase sample efficiency over $\mZ \in \sR^{N \times M}$ by reducing its dimensionality to $\sR^{N \times m}$ subspaces using
	\begin{equation}
		\begin{aligned}
			\mD \coloneq & [\1_{\ervd_1},\cdots, \1_{\ervd_m}] \in \sN^{M \times m}, \\
			\hat{\mZ} \coloneq & \mZ \mD \in \sR^{N \times m},
		\end{aligned}
		\label{eq:zhat}
	\end{equation}
	where $\mD$ linearly maps $\mZ$ to $m$ reduced embedding dimensions ($m \leq M$), $\{\ervd_1,\cdots,\ervd_m\} \in \sN$ is a set of $m$ random indices chosen without replacement from $\{1,\cdots,M\}$, and $\1_{i}$ is a one-hot vector with $i$-th element equal to 1.
	We use \cref{eq:zhat} to generate numerous subspaces $\hat{\mZ}$ that are computationally efficient, diverse, and cover $\mZ$ in the aggregate.
	We also use \cref{eq:zhat} in \cref{eq:s} to reduce the dimension of random sampling from $\rvz \in \sR^M$ to $\hat{\rvz} \coloneq \rvz \mD \in \sR^m$.
	
			\begin{figure}
		\centering
		\includegraphics[width=0.32\textwidth]{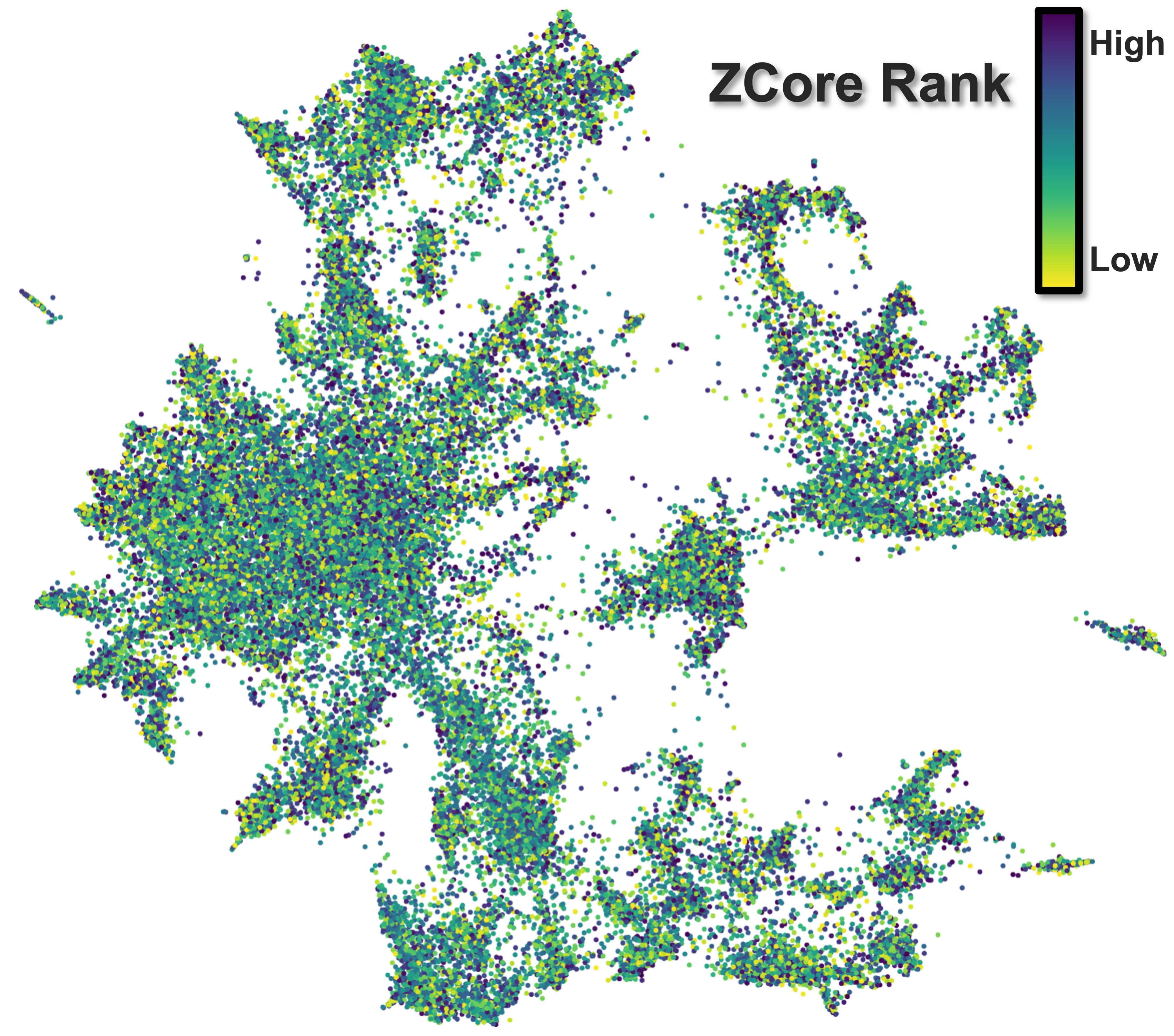}
		\vspace{-0.75em}
		\caption{\textbf{ZCore Coreset Rank Visualization} for CIFAR100. 
			Model embeddings and 2D approximation of the full embedding space generated using the FiftyOne Library and UMAP \cite{moore2020fiftyone,mcinnes2018umap-software}.
		}
		\label{fig:cifar100score}
	\end{figure} 

	We quantify coverage for each random subspace sample $\hat{\rvz}$ by finding the closest \textit{existing} dataset example
	\begin{align}
		\argmin_{i}   || \hat{\rvz} - \hat{\mZ}_{i}||_1,
		\label{eq:mces}
	\end{align}
	where we denote $k$ as the solution to $i$ in \cref{eq:mces} and $\hat{\mZ}_{k}$ is the dataset example closest to $\hat{\rvz}$.
	Finally, we quantify an importance score specifically for coverage ($\vs^\text{\tiny C}$) as 
	\begin{equation}
		\begin{aligned}
			\evs^\text{\tiny C}_i \coloneq & \begin{cases}
				1 & \text{for}~ i = k \\
				0 & \text{otherwise}
			\end{cases}, \\
			\vs^\text{\tiny C} \coloneq & [\evs^\text{\tiny C}_1, \cdots, \evs^\text{\tiny C}_N] \in \sR^N,
		\end{aligned}
		\label{eq:sc}
	\end{equation}
	where $\vs^\text{C}$ adds to the estimated embedding coverage value for dataset example $k$. 
	We repeat our process of randomly selecting $\mD$, sampling $\hat{\rvz}$, and adding coverage for the closest examples across many iterations to extend the coverage score across all examples in $\sS$.
	Unlike random sampling, our coverage score rewards hard examples that individually occupy large, unique, low-density areas of the embedding space (see \cref{fig:cifar100score}), which improves coreset performance.
	
	\noindent {\bf {\small Remarks on} $m$}: In \cref{sec:eval} experiments, we choose $m=2$ (s.t. $\mD\in\sR^{M \times 2}$) random embedding dimensions per sample $\hat{\rvz}$, which decreases runtime, generates ${M \choose 2} \approx \frac{M^2}{2}$ unique 2D subspaces from $\mZ$, and optimizes ZCore selection performance (see \cref{tab:ablate} and \cref{fig:dimruntime} in Supplementary).

	\subsection{Removing Embedding Subspace Redundancy}
	\label{sec:red}
	
	To avoid redundant coreset selection in embedding subspaces, we develop a redundancy metric that operates subsequently to each coverage solution $k$ in \cref{eq:mces}.
	Specifically, for each coverage example $\hat{\mZ}_{k}$, we quantify a redundancy penalty for the set of  $\sK \in \sN^\alpha$ nearest neighbors as
	\begin{equation}
		\begin{aligned}
			\evv^\text{\tiny R}_i \coloneq & \begin{cases}
				{( || \hat{\mZ}_{k} - \hat{\mZ}_i ||_1 )}^{\text{-}\beta} & \text{for}~ i\in\sK \\
				0 & \text{otherwise}
			\end{cases}, \\
			\vs^\text{\tiny R} \coloneq & [\evs^\text{\tiny R}_1, \cdots, \evs^\text{\tiny R}_N] = \frac{\vv^\text{\tiny R}}{||\vv^\text{\tiny R}||_1} =  \frac{[\evv^\text{\tiny R}_1, \cdots, \evv^\text{\tiny R}_N]}{||\vv^\text{\tiny R}||_1} \in \sR^N,
		\end{aligned}
		\label{eq:vr}
	\end{equation}
	where exponent $\beta$ determines the rate of penalty changes between $\hat{\mZ}_{k}$'s $\alpha$ nearest neighbors with varying distance $|| \hat{\mZ}_{k} - \hat{\mZ}_i ||_1$ and $||\vv^\text{\tiny R}||_1 \in \sR$ normalizes our redundancy score $\vs^\text{\tiny R}$ so that the coverage and redundancy scores for each sample iteration are balanced as $||\vs^\text{\tiny R}||_1 = ||\vs^\text{\tiny C}||_1 = 1$.
	
	\noindent {\bf {\small Remarks on} $\alpha,\beta$}: In \cref{sec:eval} experiments, we choose $\alpha=\text{1,000}$ to decrease runtime of \cref{eq:vr} on large datasets while still reaching many examples per iteration, and we choose $\beta=4$ to ensure that primarily the closest neighbors to each $\hat{\mZ}_{k}$ are substantially estimated as redundant.

	\subsection{Iterative Sampling for Coreset Selection}
	\label{sec:proc}
	
	Using the embedding subspace sampling process ($\mD$ and $\hat{\mZ}$ in \cref{eq:zhat}, $\hat{\rvz}$ and $k$ in \cref{eq:mces}) and subsequent coverage and redundancy metrics ($\vs^\text{\tiny C}$ in \cref{eq:sc}, $\vs^\text{\tiny R}$ in \cref{eq:vr}), we define our final coreset importance score $\vs \in \R^N$ as
	\begin{align}
				\vs \coloneq \rvs + \sum_{t=1}^{T} \vs^\text{\tiny C}_t(k_t) - \vs^\text{\tiny R}_t(k_t),
		\label{eq:score}
	\end{align}
	where $\rvs  = \{ \rs_i \sim \mathcal{U}[0,1] \} ^N_{i=1}$ is a random initialization for each example, 
	$k_t$ is the example solution in \cref{eq:mces} at iteration $t$ for subspace $\hat{\mZ}_t$ and sample $\hat{\rvz}_t$, 
	$\vs^\text{\tiny C}_t(k_t)$ and $\vs^\text{\tiny R}_t(k_t)$ are the corresponding coverage and redundancy scores for $k_t$, 
	and $T$ is the total number of sample and score iterations.
	Notably, $\rvs$ promotes inclusion of a few examples outside of our primary coverage and redundancy metrics and each iteration $t$ is independent, enabling us parallelize and accelerate ZCore score computation.

	Finally, after computing score $\vs$ to estimate the relative value of each example in unlabeled dataset $\sS$, we select the $n$ examples with highest scores (corresponding to ``High" rank in \cref{fig:cifar100score}) as our final coreset for downstream model training.
	Notably, we use the same $\vs$ for all prune rates.

	For experiments in \cref{sec:eval}, we also use $\vs$ to weight the loss and gradient for downstream model training using
	\begin{align}
		\vw = \frac{\vs - \text{min}(\vs)}{\text{max}(\vs) - \text{min}(\vs)},
	\end{align}
	where $\vw = [\evw_1, \cdots, \evw_N]^\intercal \in \sR^N$, $w_i \in [0,1]$, and the loss is scaled each batch by the mean $\evw_i$ score corresponding to the specific training examples in that batch.
	Essentially, we already determine how valuable each example is for coreset selection and want to influence model training accordingly.
	
	\begin{figure*} [t!]
		\begin{minipage}{0.33\textwidth}
			\input{CIFAR10_paper_bar.tex}
			\vspace{-1.3em}
		\end{minipage}%
		\begin{minipage}{0.335\textwidth}
			\input{CIFAR100_paper_bar.tex}
		\end{minipage}%
		\begin{minipage}{0.33\textwidth}
			\input{imagenet_paper.tex}
		\end{minipage}%
		\vspace{-0.925 em}
		\caption{Comparison of coreset selection methods using downstream model validation on {\bf CIFAR10}, {\bf CIFAR100}, and {\bf ImageNet}. 
			Method results with solid lines select coreset data using labels and training, method results with dotted lines select coreset data using self-supervised training, and method results with dashed lines select coreset data without labels \textit{or} training.
			The $x$-axis is in log scale for the number of coreset examples used for model training.
			Corresponding result tables for each dataset are provided in the Supplementary Material.
		}
		\label{fig:cifar}
	\end{figure*}
	
	\section{Evaluation}
	\label{sec:eval}
	
	\subsection{Experimental Setup}

	\noindent {\bf Datasets}. 
	We evaluate ZCore selections for downstream model training and subsequent validation accuracy on four image classification datasets: CIFAR10 \cite{cifar}, CIFAR100, ImageNet \cite{imagenet}, and EuroSAT \cite{eurosat}.
	Notably, full dataset sizes span from 1.3 \textrm{M} to 2,700 examples and coreset sizes span from 896,817 to 270 examples (over three orders of magnitude, see \cref{tab:data} in Supplementary).
	EuroSAT has no explicit training set, so we create ``four" datasets using 80/20, 40/60, 20/80, and 10/90 training/validation splits to experiment with decreasing dataset scale in the same distribution of satellite images.
	All reported accuracy results are for model validation after training on a specific coreset.
	
	\noindent {\bf Network Training}. 
	We use two different network models and training regimes to evaluate coresets.
	For CIFAR10, CIFAR100, and EuroSAT, we train a ResNet18 model on selected coresets for 200 epochs with a batch size of 128. 
	For ImageNet, we alternatively train a ResNet32 model for 60 epochs with a batch size of 256.
	Following the protocol of \cite{Zhang_2024_CVPR}, we use an SGD optimizer with momentum 0.9, weight decay 0.0005, and a learning rate of 0.1 that decays with the cosine annealing scheduler via PyTorch \cite{NEURIPS2019_bdbca288}.
	After model training, we use the model's validation accuracy to quantitatively evaluate coreset selection performance.

	\noindent {\bf ZCore \& Baselines}. 
	We implement ZCore using the \cref{sec:method} formulation with constant parameter settings across all datasets and prune rates.
	We compare ZCore against the state-of-the-art using ten previous coreset selection methods:
	{\bf Entropy} \cite{Coleman2020Selection}, {\bf Forgetting} \cite{toneva2018an}, {\bf EL2N} \cite{Paul_NEURIPS2021}, {\bf AUM} \cite{Pleiss_NEURIPS2020}, {\bf Moderate} \cite{xia2023moderate}, {\bf Dyn-Unc} \cite{He_2024_CVPR}, {\bf TDDS} \cite{Zhang_2024_CVPR}, {\bf Prototypes$_{\text{\tiny Sup.}}$} \cite{Sorscher_NEURIPS2022}, {\bf Prototypes$_{\text{\tiny SS}}$} \cite{Sorscher_NEURIPS2022}, and {\bf ELFS} \cite{zheng2025elfs}.
	ZCore is the only method to make selections without labels or data training (see \cref{fig:overview}) aside from {\bf Random}, which selects examples with uniform random sampling. 
	In place of label-based training, Prototypes$_{\text{\tiny SS}}$ and ELFS make selections after self-supervised training on candidate data.
	Detailed method descriptions and runtime \& label efficiency comparisons are provided in the Supplementary Material.

\subsection{Quantitative Coreset Performance Comparison}
	\label{sec:quant}
	\noindent {\bf CIFAR Datasets}. 
	We provide coreset selection results for CIFAR10 (10 classes) and CIFAR100 (100 classes) in \cref{fig:cifar} (50,000 images each).
	Full dataset training on the ResNet18 model achieves 95.23\% (CIFAR10) and 78.21\% (CIFAR100) accuracy.
	Despite making selections without labels or training, ZCore achieves the second best overall performance, staying relatively competitive with SOTA label- and training-based TDDS selections all the way through the lowest 10\% data rate.
	Meanwhile, most coreset methods exhibit a substantial performance drop at 30\% data selection on CIFAR100 and at 20\% selection on CIFAR10.
	
	\noindent {\bf ImageNet Dataset}. 
	We provide coreset selection results for ImageNet (1,000 classes) in \cref{fig:cifar}.
	Full dataset training (1.28\textrm{M} images) on the ResNet32 model achieves 73.54\% accuracy.
	ZCore and TDDS coreset selections achieve the best overall performance, establishing consistent performance on a larger, more difficult dataset.
	Notably, ZCore selects the best performing coreset at the 10\% data selection rate, reducing costs for 1.15\textrm{M} pruned images.

	\noindent {\bf Decreasing Dataset Size on EuroSAT}. 
	We provide coreset selection results for all EuroSAT dataset splits in \cref{fig:euro}, which evaluates coreset selection at a decreasing scale on satellite images (all 10 classes).
	Full dataset training on the ResNet18 model achieves 98.59\% (EuroSAT80, 21,600 images), 98.20\% (EuroSAT40, 10,800 images), 97.36\% (EuroSAT20, 5,400 images), and 93.64\% accuracy (EuroSAT10, 2,700 images).  
	Overall, ZCore cuts much of the performance gap between Random and TDDS selections.
	For the 10\% selection rates, ZCore outperforms TDDS on EuroSAT40 but has a lower accuracy than TDDS and Random on EuroSAT20 and EuroSAT10, where the pruned coresets only have 540 and 270 training examples.
	Notably, unlike TDDS, ZCore uses constant parameter settings across all prune rates.
	On the other hand, ZCore small dataset performance improves with alternative settings (e.g., reducing  number of neighbors for redundancy in \cref{eq:vr}). 
	
\begin{figure*}
	\begin{minipage}{0.25\textwidth}
		\input{EuroSAT80.tex}
	\end{minipage}%
	\begin{minipage}{0.25\textwidth}
		\input{EuroSAT40.tex}
	\end{minipage}%
	\begin{minipage}{0.25\textwidth}
		\input{EuroSAT20.tex}
	\end{minipage}%
	\begin{minipage}{0.25\textwidth}
		\input{EuroSAT10.tex}
	\end{minipage}%
	\vspace{-2.25em}
	\caption{
		Comparison of coreset methods using downstream model validation on 80/20, 40/60, 20/80, and 10/90 training/validation splits of {\bf EuroSAT}.
		Only TDDS selects data using labels and training.
		Corresponding results \cref{tab:euro} is provided in the Supplementary Material.
	}
	\label{fig:euro}
\end{figure*}

	\setlength{\tabcolsep}{2.35pt}
	\begin{table}[t!]
		\centering
		\caption{{\bf Comparison of Coreset Selection Efficiency} on CIFAR100 for 70\%, 30\%, and 10\% data rates. 
			Selection ``Total" is the runtime to generate all three coresets. 
			Runtimes use a standard M3 Max-equipped ``Laptop" or a single L40S ``GPU" on a Lambda Scalar.
			ELFS hardware requirements prohibit laptop experiments.
			Model training and full details provided in Supplementary \cref{tab:efficient}.
		}
		\vspace{-0.875em}
		\label{tab:eff}
		\small
		\scriptsize
		\begin{tabular}{l c r r r r c r c r r r }
			\noalign{\global\arrayrulewidth=0.4mm} \hline \noalign{\global\arrayrulewidth=0.2mm} \arrayrulecolor{gray}
			&& \multicolumn{6}{c}{\footnotesize ~~~~~~~~~~~~~~~~~~~~~Selection Runtime (\textrm{s})} && \multicolumn{3}{c}{\footnotesize Train Labels}\rule{0pt}{2.5ex}  \\
			\multicolumn{1}{c}{\footnotesize Setup} && \multicolumn{4}{c}{\footnotesize Components} && \multicolumn{1}{c}{\footnotesize Total} && \multicolumn{1}{c}{\footnotesize 70\%} &\multicolumn{1}{c}{\footnotesize 30\%} & \multicolumn{1}{c}{\footnotesize 10\%}\rule{0pt}{2.25ex} \\
			\noalign{\global\arrayrulewidth=0.4mm} \arrayrulecolor{white}\hline \noalign{\global\arrayrulewidth=0.2mm} \arrayrulecolor{gray}
			\cline{1-1} \cline{3-6} \cline{8-8} \cline{10-12}
			\multicolumn{1}{l}{\textcolor{gray}{ \small \bf }} && \multicolumn{3}{c}{\textcolor{gray}{\scriptsize ~~Embedding Generation}}  & \multicolumn{1}{c}{\textcolor{gray}{\scriptsize ZCore}}              \rule{0pt}{2.5ex} \vspace{-0.1em} \\ 
			\multicolumn{1}{l}{\textcolor{gray}{ \bf \footnotesize ZCore}} && \multicolumn{1}{c}{\textcolor{gray}{\scriptsize ResNet18}} & \multicolumn{1}{c}{\textcolor{gray}{\scriptsize CLIP}}  & \multicolumn{1}{c}{\textcolor{gray}{\scriptsize Parallel}}  & \multicolumn{1}{c}{\textcolor{gray}{\scriptsize ~Score}}          \rule{0pt}{0.0ex}  \\ 
			Laptop	&&  635	&  6,471	& 6,471	& 382~	&& \bf 6,853 &&	 35\textrm{K}	&  15\textrm{K} & \bf 5\textrm{K} \\
			GPU && 145	& 97.0	& 145	& 54.7~	&&  \bf 199.8 &&	35\textrm{K}	& 15\textrm{K} & \bf 5\textrm{K}  \\
			\multicolumn{1}{l}{\textcolor{gray}{ \small \bf }} && \multicolumn{1}{c}{\textcolor{gray}{\scriptsize Train}}  & \multicolumn{3}{c}{\textcolor{gray}{\scriptsize TDDS Score}}              \rule{0pt}{2.5ex} \vspace{-0.1em} \\ 
			\multicolumn{1}{l}{\textcolor{gray}{ \bf \footnotesize TDDS}} && \multicolumn{1}{c}{\textcolor{gray}{\scriptsize Dynamics}} & \multicolumn{1}{r}{\textcolor{gray}{\scriptsize 70\%}}  & \multicolumn{1}{c}{\textcolor{gray}{\scriptsize ~~~30\%}}  & \multicolumn{1}{r}{\textcolor{gray}{\scriptsize 10\%}}          \rule{0pt}{0.0ex}  \\ 
			Laptop	&& 188,295 &	27.0& 25.5 &	0.9~ && 188,296 &&	50\textrm{K}	& 50\textrm{K} & 50\textrm{K} \\
			GPU && 1,540 & 22.6	& 20.9	& 1.3~ &&	1,541 &&	50\textrm{K}	& 50\textrm{K} & 50\textrm{K} \\
			\multicolumn{1}{l}{\textcolor{gray}{ \small \bf }} && \multicolumn{1}{c}{\textcolor{gray}{\scriptsize ~~~~DINO}}  & \multicolumn{1}{c}{\textcolor{gray}{\scriptsize Deep}} & \multicolumn{1}{c}{\textcolor{gray}{\scriptsize Pseudo}}    & \multicolumn{1}{c}{\textcolor{gray}{\scriptsize ELFS}}         \rule{0pt}{2.5ex} \vspace{-0.1em} \\ 
			\multicolumn{1}{l}{\textcolor{gray}{ \bf \footnotesize ELFS}} && \multicolumn{1}{c}{\textcolor{gray}{\scriptsize ~~~~Embed}} & \multicolumn{1}{r}{\textcolor{gray}{\scriptsize Cluster}}  & \multicolumn{1}{c}{\textcolor{gray}{\scriptsize Train}}  & \multicolumn{1}{c}{\textcolor{gray}{\scriptsize Score}}          \rule{0pt}{0.0ex}  \\ 
			Laptop &&	---~ &---~~ &---~~ &---~~~	&& ---~~~ &&	 ---~	&  ---~ & ---~  \\
			GPU && 111	& 1,093	& 1,476& 	30.3~	&& 7,908 &&	 35\textrm{K}	&  15\textrm{K} & \bf 5\textrm{K}  \\
			\noalign{\global\arrayrulewidth=0.4mm} \arrayrulecolor{black}\hline \noalign{\global\arrayrulewidth=0.2mm}
		\end{tabular}
	\end{table}
	
	\begin{figure*}
		\centering
		\includegraphics[width=0.89\textwidth]{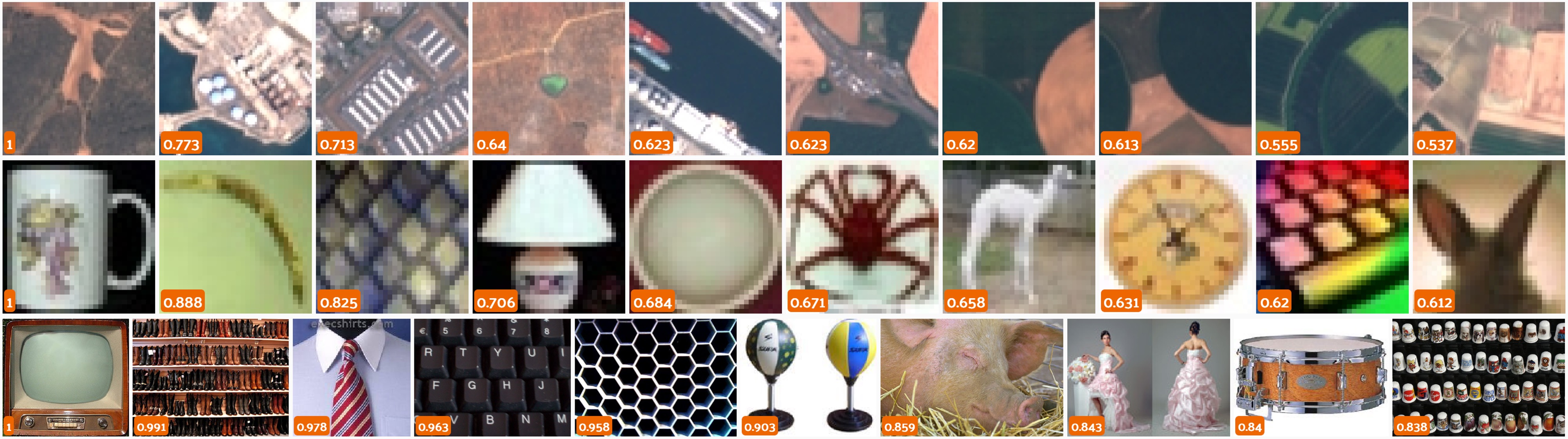}
		\vspace{-0.8em}
		\caption{{\bf Top-10 ZCore Selection Scores} (orange) for EuroSAT80 (top), CIFAR100 (middle), and ImageNet (bottom).
		}
		\label{fig:good}
	\end{figure*}

\subsection{Quantitative Coreset Efficiency Comparison}
\label{sec:eff}
We compare runtime and label efficiency of TDDS (labeled training-based selection), ELFS (label-free training-based selection), and ZCore (no labels \textit{or} training) in \cref{tab:eff}.

For runtime, ZCore selects the three CIFAR100 coresets 27.5$\times$ faster than TDDS on a laptop and 39.6$\times$ faster than ELFS on a GPU.
In reality, the relative time to implement ZCore is even faster.
Reason being, ZCore does not use hyperarameter searches to determine unique settings for each dataset and prune rate.
Instead, ZCore uses the exact same settings across all datasets and the exact same score across all prune rates.
On the other hand, both comparison methods perform parameter searches on \textit{each} dataset \textit{and} prune rate to determine their final dataset- and prune-specific settings.
It is cost-prohibitive to replicate the hyperparameter search of each comparison method for our experiments in \cref{tab:eff}, but as one example, ELFS's developers report it takes approximately 17 hours using four A6000 GPUs to find a single prune rate setting on ImageNet \cite{zheng2025elfs}.

For label efficiency, TDDS uses ground truth labels on all candidate data prior to selection.
On the other hand, ZCore and ELFS only use labels for downstream model training on the selected coreset.
Thus, TDDS requires 1.4, 3.3, and 10$\times$ more labels than ZCore to train models at the respective 70\%, 30\%, and 10\% data selection rates.

\subsection{Qualitative Results}
\label{sec:qual}
\begin{figure} [t!] 
	\centering
	\includegraphics[width=0.475\textwidth]{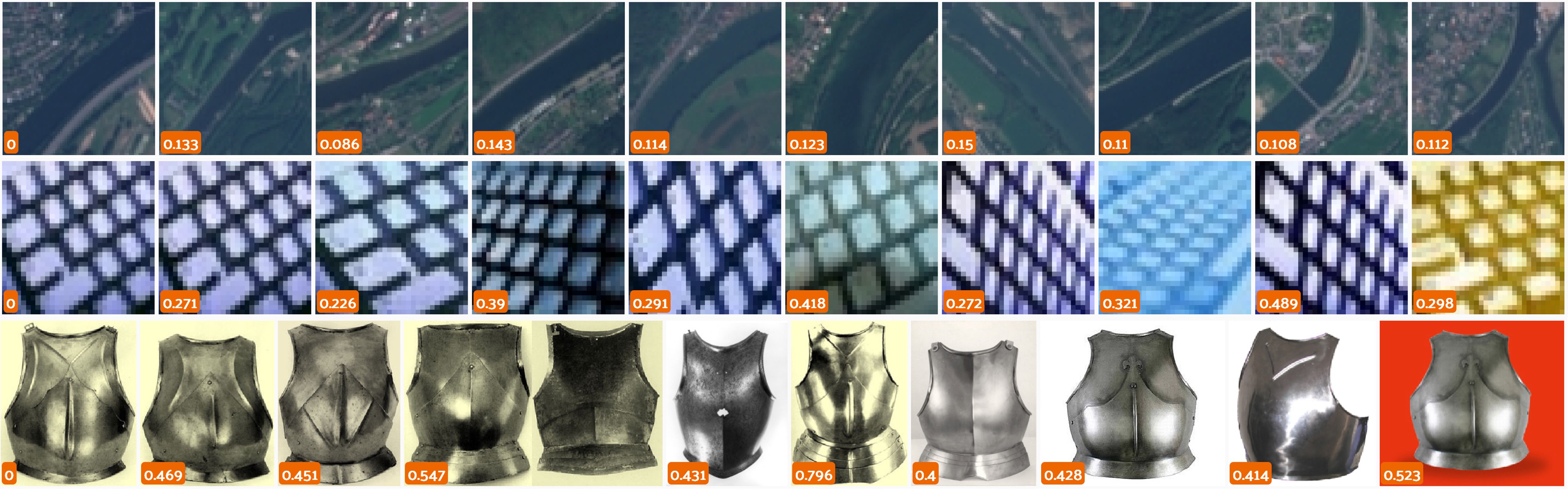}
	\vspace{-1.975em}
	\caption{{\bf Lowest ZCore Selection Score} (column 1) with nine nearest neighbors in the full embedding space for EuroSAT80 (top), CIFAR100 (middle), and ImageNet (bottom). 
	}
	\label{fig:bad}
\end{figure}

We show the highest ranked ZCore images for EuroSAT80, CIFAR100, and ImageNet in \cref{fig:good}.
Across all dataset sizes and settings, the highest-ranked images are all unique, demonstrating that the embedding-based coverage score successfully promotes diverse examples from the full dataset without labels or training.
In fact, ZCore achieves full class recall across all datasets and prune rates (full recall details and additional qualitative results in the Supplementary).
For the same three datasets, we also show the lowest-ranked ZCore images with their nearest embedding neighbors in \cref{fig:bad}.
For each dataset, the lowest-ranked images are highly redundant, which demonstrates that the embedding-based redundancy score is penalizing and subsequently pruning redundant examples from the full dataset.

\subsection{Ablation Study}
\label{sec:ablate}

We provide ZCore ablative results in \cref{tab:ablate}.
When using a single model to generate our embedding space ($\mZ$), ResNet18 outperforms CLIP, but neither perform as well as the standard concatenated setting.
We also test a DINOv2 ViT-B-14 model \cite{oquab2024dinov}, which shows that ZCore maintains near full performance even with a self-supervised zero-shot embedding space.
Gaussian sampling ($\rvz$) outperforms Uniform but does not match Triangular performance.
Given the close performance gap between Triangular and Gaussian sampling, we postulate that exploring additional sampling strategies is a promising area for future work.
Decreasing or increasing the subspace sampling dimension ($m$) leads to lower performance, with high-dimension samples performing worst in terms of \textit{both} accuracy and runtime (see \cref{fig:dimruntime} in Supplementary).
We postulate the accuracy decrease occurs because the $L_1$ distance metric in \cref{eq:mces} is less meaningful in higher-dimensional space \cite{Park_2024_CVPR}.
Nonetheless, the $L_1$ metric outperforms the $L_2$ and $\cos$ distance ablations.

For the importance score, we test multiple redundancy settings ($\alpha, \beta$) and find that ZCore performance is robust to these parameter changes.
Removing the redundancy score entirely ($\vs^\text{R}$) decreases performance more substantially than any other ablative configuration, validating our design choice to penalize nearest neighbors in each embedding subspace to reduce redundancy.
Replacing embedding coverage sample selection with a uniformly random sample $k$ also decreases performance, which validates our design choice to emphasize selection of examples that individually occupy larger, lower-density embedding areas. 
Finally, removing random initialization ($\rvs$) or the score loss weight $\vw$ from model training decreases performance.

We plot the runtime and accuracy performance of ZCore over a wide range of score iterations in \cref{fig:runtime}.
We measure coreset selection runtime using a M3 Max-equipped laptop.
The largest accuracy increase occurs when the coverage and redundancy score iterations ($T$) increase from 100 to 1,000, at which point, with the redundancy score reaching 1,000 neighbors per iteration, the score likely reaches most of the 50,000 CIFAR100 candidate training examples.
The accuracy peaks at the default 1\textrm{M} iteration setting then converges on a slightly lower accuracy with increasing iterations.
Notably, the default setting takes less than 400 \textrm{s} on a standard laptop, making ZCore a computationally efficient alternative to training-based coreset selection methods.

Additional ablative studies for ImageNet embedding models are provided in the Supplementary Material.

\section{Conclusion}

Recent state-of-the-art coreset selection methods improve performance in experiments by using labels and training on candidate data prior to selection.
However, the majority of real-world data are unlabeled.
To address this, we developed a coreset selection method, ZCore, that maintains state-of-the-art performance without labels or training.
Instead, ZCore selects data using a novel iterative subspace sampling technique that operates on a high-dimensional embedding space representation of unlabeled data.
In our experiments, we find that embeddings can be generated by a variety of previously-trained, off-the-shelf foundation models, including those originally trained with self-supervision.
Not training on candidate data reduces computation, and unlabeled selection enables data- \textit{and} label-efficient downstream model training on the selected coreset.

To evaluate our approach, we include ten state-of-the-art methods in experiments that select coresets, train downstream models, and evaluate subsequent validation accuracy on four datasets.
Our experiment scale ranges from full datasets of over a million images all the way down to pruned coresets of 270 training images.
In these experiments, our method outperforms all others save one, which uses full ground truth labels and model training on all data before coreset selection as well as prune rate-specific tuning on the validation set for configuration.
In contrast, ZCore uses constant algorithmic settings across all experiments, making it generalizable, and selects coresets without labels or training, making it much more efficient for coreset selection at the rapidly expanding scale of our current deep learning era.
From these results, ZCore establishes a new state-of-the-art approach to coreset selection.

We discuss future work in the Supplementary Material.

	\setlength{\tabcolsep}{5.1 pt}
	\begin{table}
		\centering
		\caption{Comparison of {\bf ZCore ablations} on CIFAR 100 with ResNet18. 
			Accuracy is the mean across 70\%, 50\%, 30\%, 20\%, and 10\% data rates over five repeat trials.
			``ResNet18, CLIP" is the default concatenated embedding space using both foundation models off-the-shelf.
			Additional details in Supplementary \cref{tab:fullablate}.
		}
		\vspace{-0.875em}
		\label{tab:ablate}
		\scriptsize
		\begin{tabular}{l c  l c c c c }
			\noalign{\global\arrayrulewidth=0.4mm} \hline \noalign{\global\arrayrulewidth=0.2mm}
			\noalign{\global\arrayrulewidth=0.4mm} \arrayrulecolor{white}\hline \noalign{\global\arrayrulewidth=0.2mm} \arrayrulecolor{gray}
			\multicolumn{1}{c}{ {\footnotesize \bf \color{gray} Full Method} / Ablation} & &  \multicolumn{1}{c}{\footnotesize Setting} &&\footnotesize Accuracy \rule{0pt}{2ex} \\
			\noalign{\global\arrayrulewidth=0.4mm} \arrayrulecolor{white}\hline \noalign{\global\arrayrulewidth=0.2mm} \arrayrulecolor{gray}
			\cline{1-1} \cline{3-3} \cline{5-5}  
			\bf \color{gray} ZCore& &  \bf \color{gray} ResNet18, CLIP && \bf \color{gray} 65.77 \rule{0pt}{2ex} \\ 
			Zero-Shot Embedding Space {\color{gray} ($\mZ$)} & &  ResNet18 \cite{HeEtAl16} & &  65.52 \rule{0pt}{2ex} \\
			& &  CLIP \cite{clip} & &  64.84 \rule{0pt}{2ex} \\
			& &  DINOv2 \cite{oquab2024dinov} &&  65.61 \rule{0pt}{2ex} \\
			\hline
			\bf \color{gray} ZCore& &  \bf \color{gray}  Triangular && \bf  \color{gray} 65.77 \rule{0pt}{2ex} \\ 
			Sampling Distribution {\color{gray} ($\rvz$)} & &  Gaussian  && 65.75 \rule{0pt}{2ex} \\
			& &   Uniform & &  64.84 \rule{0pt}{2ex} \\
			\hline
			\bf \color{gray} ZCore& & \bf \color{gray}  2  &&  \bf \color{gray} 65.77 \rule{0pt}{2ex} \\ 
			Subspace Sampling Dimensions {\color{gray} ($m$)}  & &   1 &&  64.59 \rule{0pt}{2ex} \\
			& &   3 &&  65.10 \rule{0pt}{2ex} \\
			\small  & & 10 &&  63.27 \rule{0pt}{2ex} \\
			\small  & & 100 &&  62.20 \rule{0pt}{2ex} \\ 
			\hline
			\bf \color{gray} ZCore& & \bf \color{gray}  \boldmath$L_1$ && \bf \color{gray} 65.77 \rule{0pt}{2ex} \\ 
			Distance Metric {\color{gray}(\cref{eq:mces}, \cref{eq:vr})} && $L_2$  &&  65.60 \rule{0pt}{2ex} \\
			& & $\cos$ &&   64.48 \rule{0pt}{2ex} \\ 
			\hline
			\bf \color{gray} ZCore& &  \bf \color{gray} 1,000 &&  \bf \color{gray} 65.77 \rule{0pt}{2ex} \\ 
			Number of Nearest  Neighbors {\color{gray}($\alpha$)} && 100&&  65.68 \rule{0pt}{2ex} \\
			&& 10,000&&   65.72 \rule{0pt}{2ex} \\ 
			\hline
			\bf \color{gray} ZCore& & \bf \color{gray}  4 &&  \bf \color{gray} 65.77 \rule{0pt}{2ex} \\ 
			Distance Penalty Exponent {\color{gray}($\beta$)} & &  3 &&   65.64 \rule{0pt}{2ex} \\
			& &  5 &&   65.74 \rule{0pt}{2ex} \\ 
			\hline
			\bf \color{gray} ZCore & &  \bf \color{gray} Full Score &&  \bf \color{gray} 65.77 \rule{0pt}{2ex} \\ 
			Random Coverage Sample {\color{gray}($k$)} & &  Partial Score && 64.74 \rule{0pt}{2ex} \\
			No Redundancy Score {\color{gray} ($\vs^\text{\tiny R}$)} && Partial Score && 61.71 \rule{0pt}{2ex} \\
			No Random Initialization {\color{gray} ($\rvs$)} & &  Partial Score && 65.66 \rule{0pt}{2ex} \\
			No Score Loss Weight {\color{gray} ($\vw$)} & &  Partial Score && 63.53 \rule{0pt}{2ex} \\
			\noalign{\global\arrayrulewidth=0.4mm} \arrayrulecolor{black}\hline \noalign{\global\arrayrulewidth=0.2mm}
		\end{tabular}
	\end{table}
	
	\begin{figure}
	\centering
	\begin{minipage}{0.5\textwidth}
		\input{nsample_accuracy_paper_vertical.tex}
	\end{minipage}%
	\\
	\vspace{-0.5em}
	\begin{minipage}{0.5\textwidth}
		\input{nsample_runtime_paper_vertical.tex}
	\end{minipage}%
	\vspace{-1.1em}
	\caption{Comparison of number of sample iterations vs. accuracy (top) and {\bf runtime} (bottom) on CIFAR100.
		Accuracy is the mean across 70\%, 50\%, 30\%, 20\%, and 10\% data selection rates over five repeat trials.
		Runtime is selection time using a standard laptop.
	}
	\label{fig:runtime}
\end{figure}

{
    \small
    \bibliographystyle{ieeenat_fullname}
    \bibliography{refzcore}
}

\newpage

\twocolumn[
\begin{center}
	\vspace{10mm}
	\Large\textbf{Zero-Shot Coreset Selection via Iterative Subspace Sampling} \\
	\vspace{7mm}
	\large{Supplementary Material}
	\vspace{6mm}
\end{center}
]

\section*{Reproducibility Statement}
We provide detailed method and experiment descriptions in \cref{sec:method,sec:eval} of the main paper.
We generate all ZCore experimental results from a single attempt of five consecutive trials with the exception of ImageNet, which is from a single attempt of one trial.
Our paper's code is publicly available at {\url{https://github.com/voxel51/zcore}}.

\setlength{\tabcolsep}{7.2pt}
\renewcommand{\arraystretch}{1}
\begin{table*}
	\centering
	\caption{Comparison of {\bf full training and coreset size} across all datasets. 
		ZCore uses constant algorithmic settings across all experiments.
	}
	\vspace{-0.875em}
	\label{tab:data}
	\small
	\begin{tabular}{l  c c c c c c c c c c c c}
		\noalign{\global\arrayrulewidth=0.4mm} \hline \noalign{\global\arrayrulewidth=0.2mm} \arrayrulecolor{gray}
		\multicolumn{1}{c}{} & &  && Number of && Full Dataset && \multicolumn{5}{c}{Percent of Dataset Selected for Coreset}  \rule{0pt}{2ex} \\ 
		\multicolumn{1}{c}{Dataset} && \multicolumn{1}{c}{Scale} & & Classes &&  Training Size && 70\% & 50\% & 30\% & 20\% & 10\%    \rule{0pt}{0ex} \\ 
		\noalign{\global\arrayrulewidth=0.4mm} \arrayrulecolor{white}\hline \noalign{\global\arrayrulewidth=0.2mm} \arrayrulecolor{gray}
		\cline{1-1} \cline{3-3} \cline{5-5} \cline{7-7} \cline{9-13}
		ImageNet	 && 	Large	 && 	 1,000 	 && 	 1,281,167 	 && 	 896,817 	 & 	 640,584 	 & 	 384,350 	 & 	 256,233 	 & 	 128,117 \rule{0pt}{2ex}	\\ \hline
		CIFAR100	 && 	Medium	 && 	 100 	 && 	 50,000 	 && 	 35,000 	 & 	 25,000 	 & 	 15,000 	 & 	 10,000 	 & 	 ~5,000 	\rule{0pt}{2ex} \\
		CIFAR10	 && 	Medium	 && 	 10 	 && 	 50,000 	 && 	 35,000 	 & 	 25,000 	 & 	 15,000 	 & 	 10,000 	 & 	 5,000 	\\
		EuroSAT80	 && 	Medium	 && 	 10 	 && 	 21,600 	 && 	 15,120 	 & 	 10,800 	 & 	 6,480 	 & 	 4,320 	 & 	 2,160 	\\ \hline
		EuroSAT40	 && 	Small	 && 	 10 	 && 	 10,800 	 && 	 7,560 	 & 	 5,400 	 & 	 3,240 	 & 	 2,160 	 & 	 1,080 	\rule{0pt}{2ex} \\
		EuroSAT20	 && 	Small	 && 	 10 	 && 	 5,400 	 && 	 3,780 	 & 	 2,700 	 & 	 1,620 	 & 	 1,080 	 & 	 540 	\\
		EuroSAT10	 && 	Small	 && 	 10 	 && 	 2,700 	 && 	 1,890 	 & 	 1,350 	 & 	 810 	 & 	 540 	 & 	 270 	\\
		\noalign{\global\arrayrulewidth=0.4mm} \arrayrulecolor{black}\hline \noalign{\global\arrayrulewidth=0.2mm}
	\end{tabular}
\end{table*}

\section*{Experiment Datasets \& External Baselines}
To confirm that ZCore selects high-performing coresets, we perform experiments with ten state-of-the-art methods to select coresets, train downstream models, and evaluate performance on four datasets.
Our experiments expand the recent work of Zhang et al.~\cite{Zhang_2024_CVPR}, which evaluated their coreset method with seven external baselines and three datasets.

Here, we provide detailed method descriptions for external baselines.
{\bf Entropy} selects examples with high entropy of predicted probabilities at the end of fully-supervised training on all candidate data \cite{Coleman2020Selection}.
{\bf Forgetting} selects examples that change to being misclassified after correct classification the most times during training \cite{toneva2018an}.
{\bf EL2N} selects examples with high gradient magnitude using the L2 norm of error vectors \cite{Paul_NEURIPS2021}.
{\bf AUM} selects examples with high area under the margin, i.e., the probability gap between the target class and the next largest class across all epochs \cite{Pleiss_NEURIPS2020}.
{\bf Moderate} selects examples closest to the median class value in the full dataset trained model embedding space \cite{xia2023moderate}.
{\bf Dyn-Unc} selects examples with high target class probability variance during training \cite{He_2024_CVPR}.
{\bf TDDS} selects examples with high projected gradient variance across many epochs \cite{Zhang_2024_CVPR}.
{\bf Prototypes$_{\text{\tiny Sup.}}$} selects examples based on difficulty, which is determined using $k$-means embedding clustering and the distance of each example to cluster centroids or \textit{prototypes} \cite{Sorscher_NEURIPS2022}. 
Notably, Prototypes$_{\text{\tiny Sup.}}$ uses a supervised model trained on candidate data to generate embeddings while {\bf Prototypes$_{\text{\tiny SS}}$} uses a self-supervised model \cite{Sorscher_NEURIPS2022}.
{\bf ELFS} uses an extensive framework of deep clustering on pretrained model embeddings to generate psuedo labels, subsequent full model training on pseudo labels, and finally selecting examples based on pseudo training dynamic-based scores without ground truth labels \cite{zheng2025elfs}.

ZCore is the only method to make selections without ground truth labels \textit{or} dataset training (see \cref{fig:overview} of the main paper) aside from {\bf Random}, which selects examples with uniform random sampling.
Furthermore we evaluate ZCore under constant algorithmic settings across a dataset and coreset selection scale greater than three orders of magnitude. 
Specifically, full dataset sizes span from 1.3 \textrm{M} to 2,700 examples and coreset sizes span from 896,817 to 270 examples (see \cref{tab:data}.). 
To our knowledge, there is no precedent for this selection range of experiments and generalization in the coreset selection literature.

We also considered several other selection methods as baselines for our experiments.
Unfortunately, as we will explain, these selection methods and many others are out of scope. 
Notably, our experiments use consistent prune rates for fair comparisons across coreset selection methods.
On the other hand, SemDeDup \cite{abbas2023semdedup}, which selects coresets to remove semantic duplicates, uses configuration parameters that result in a prune rate that is unknown until \textit{after} coreset selection experiments end.
Perhaps due to this variable prune rate, the SemDeDup paper includes only a single data point as an external methodological comparison.
GIGA constructs Bayesian coresets that scale log-likelihood optimality \cite{pmlr-v80-campbell18a}, but is originally applied to regression and would need to be modified for image classification.
Other selection baselines originally applied to natural language processing would similarly require modification for our experiments \cite{kirchhoff-bilmes-2014-submodularity,10.1145/3589302}, which would materially change their functionality.

\setlength{\tabcolsep}{3.1pt}
\begin{table*}[t!]
	\centering
	\caption{{\bf Comparison of Coreset Selection Efficiency} on CIFAR100 for 70\%, 30\%, and 10\% data rates. Selection ``Total" is the runtime to generate all three coresets. Notably, ZCore uses the same score across all settings, while TDDS uses a unique validation-tuned setting for each prune rate (see \cref{fig:overview}).
		ELFS also uses a unique hyperparameter-tuned setting for each prune rate and additionally requires full-data deep clustering, pseudo training dynamics generation, and score calculation for each prune rate.
		Runtimes use a standard M3 Max-equipped ``Laptop" or a single L40S ``GPU" on a Lambda Scalar.
		ELFS hardware requirements prohibit laptop experiments.
		Following the protocol of TDDS \cite{Zhang_2024_CVPR}, downstream model training at a 10\% data rate on a GPU takes longer than 30\% due to a reduced batch size.
	}
	\vspace{-0.875em}
	\label{tab:efficient}
	\small
	\begin{tabular}{l c r r r r c r c r r r c r r r c r r r r}
		\noalign{\global\arrayrulewidth=0.4mm} \hline \noalign{\global\arrayrulewidth=0.2mm} \arrayrulecolor{gray}
		\multicolumn{1}{c}{Compute} && \multicolumn{6}{c}{~~~~~~~~~~~~~~~~~~Selection Runtime (\textrm{s})} && \multicolumn{3}{c}{Model Train Time (\textrm{s})} && \multicolumn{3}{c}{Selection + Train Time (\textrm{s})} && \multicolumn{3}{c}{Total Labels}\rule{0pt}{2ex} \\
		\multicolumn{1}{c}{Hardware} && \multicolumn{4}{c}{Components} && \multicolumn{1}{c}{Total} && \multicolumn{1}{c}{70\%} &\multicolumn{1}{c}{30\%} & \multicolumn{1}{c}{10\%} && \multicolumn{1}{c}{70\%} &\multicolumn{1}{c}{30\%} & \multicolumn{1}{c}{10\%} \rule{0pt}{0ex} && \multicolumn{1}{c}{70\%} &\multicolumn{1}{c}{30\%} & \multicolumn{1}{c}{10\%} \\
		\noalign{\global\arrayrulewidth=0.4mm} \arrayrulecolor{white}\hline \noalign{\global\arrayrulewidth=0.2mm} \arrayrulecolor{gray}
		\cline{1-1} \cline{3-6} \cline{8-8} \cline{10-12} \cline{14-16} \cline{18-20}
		\multicolumn{1}{l}{\textcolor{gray}{ \small \bf }} && \multicolumn{3}{c}{\textcolor{gray}{\scriptsize ~~Embedding Generation}}  & \multicolumn{1}{c}{\textcolor{gray}{\scriptsize ZCore}}              \rule{0pt}{2.5ex} \vspace{-0.4em} \\ 
		\multicolumn{1}{l}{\textcolor{gray}{ \small \bf ZCore}} && \multicolumn{1}{c}{\textcolor{gray}{\scriptsize ResNet18}} & \multicolumn{1}{c}{\textcolor{gray}{\scriptsize CLIP}}  & \multicolumn{1}{c}{\textcolor{gray}{\scriptsize Parallel}}  & \multicolumn{1}{c}{\textcolor{gray}{\scriptsize ~Score}}          \rule{0pt}{0.0ex}  \\ 
		Laptop	&&  635	&  6,471	& 6,471	& 382~	&& \bf 6,853	&& 130,503	& 64,877 & 41,711	&& \bf 137,356	& \bf 71,730	& \bf 48,564	&& \bf 35\textrm{K}	& \bf 15\textrm{K}	& \bf 5\textrm{K} \\
		GPU && 145	& 97.0	& 145	& 54.7~	&&  \bf 199.8	&&	990	& 589	& 643	&& \bf 1,190 &	\bf 789	& \bf 843&& \bf 35\textrm{K}	& \bf 15\textrm{K}	& \bf 5\textrm{K} \\
		\multicolumn{1}{l}{\textcolor{gray}{ \small \bf }} && \multicolumn{1}{c}{\textcolor{gray}{\scriptsize Train}}  & \multicolumn{3}{c}{\textcolor{gray}{\scriptsize TDDS Score}}              \rule{0pt}{2.5ex} \vspace{-0.4em} \\ 
		\multicolumn{1}{l}{\textcolor{gray}{ \small \bf TDDS}} && \multicolumn{1}{c}{\textcolor{gray}{\scriptsize Dynamics}} & \multicolumn{1}{r}{\textcolor{gray}{\scriptsize 70\%}}  & \multicolumn{1}{c}{\textcolor{gray}{\scriptsize ~~~30\%}}  & \multicolumn{1}{r}{\textcolor{gray}{\scriptsize 10\%}}          \rule{0pt}{0.0ex}  \\ 
		Laptop	&& 188,295 &	27.0& 25.5 &	0.9~ && 188,296 && 130,503 & 64,877& 41,711&&	318,825	& 253,198 &	230,008 &&	50\textrm{K}	& 50\textrm{K} & 50\textrm{K} \\
		GPU && 1,540 & 22.6	& 20.9	& 1.3~ &&	1,541	&&	990	& 589	& 643&&	2,553	&2,150&	2,184	&&	50\textrm{K}	& 50\textrm{K} & 50\textrm{K} \\
		\multicolumn{1}{l}{\textcolor{gray}{ \small \bf }} && \multicolumn{1}{c}{\textcolor{gray}{\scriptsize ~~~~DINO}}  & \multicolumn{1}{c}{\textcolor{gray}{\scriptsize Deep}} & \multicolumn{1}{c}{\textcolor{gray}{\scriptsize Pseudo}}    & \multicolumn{1}{c}{\textcolor{gray}{\scriptsize ELFS}}         \rule{0pt}{2.5ex} \vspace{-0.4em} \\ 
		\multicolumn{1}{l}{\textcolor{gray}{ \small \bf ELFS}} && \multicolumn{1}{c}{\textcolor{gray}{\scriptsize ~~~~Embed}} & \multicolumn{1}{r}{\textcolor{gray}{\scriptsize Cluster}}  & \multicolumn{1}{c}{\textcolor{gray}{\scriptsize Train}}  & \multicolumn{1}{c}{\textcolor{gray}{\scriptsize Score}}          \rule{0pt}{0.0ex}  \\ 
		GPU && 111	& 1,093	& 1,476& 	30.3~	&& 7,908	&&	990	& 589	& 643&&	3,701 &	3,300	& 3,353	&&	\bf 35\textrm{K}	& \bf 15\textrm{K} & \bf 5\textrm{K} \\
		\multicolumn{15}{l}{\textcolor{gray}{ \small \bf TDDS / ZCore Runtime \& Label Requirement Ratio}} \rule{0pt}{3ex}  \\ 
		Laptop && ---~~~ &---~ &---~ &---~ && \bf 27.5		&&	\multicolumn{1}{c}{~~~~---}	& \multicolumn{1}{c}{~~~~---}	 & \multicolumn{1}{c}{~~~~---}	 &&	2.32	&3.53	&4.74		&& \bf 1.43	& \bf 3.33	&\bf 10.0 \\
		GPU    && ---~~~ &---~ &---~ &---~ && \bf 7.7		&&	\multicolumn{1}{c}{~~~~---}	& \multicolumn{1}{c}{~~~~---}	& \multicolumn{1}{c}{~~~~---}	 &&	2.14	&2.72	&2.59		&& \bf 1.43	& \bf 3.33	&\bf 10.0 \\
		\multicolumn{15}{l}{\textcolor{gray}{ \small \bf ELFS / ZCore Runtime Requirement Ratio}} \rule{0pt}{3ex}  \\ 
		GPU    && ---~~~ &---~ &---~ &---~ && \bf 39.6	&&	\multicolumn{1}{c}{~~~~---}	& \multicolumn{1}{c}{~~~~---}	& \multicolumn{1}{c}{~~~~---}	 &&	3.11 &	4.18 & 3.98		&& 1.00	& 1.00	& 1.00 \\
		\noalign{\global\arrayrulewidth=0.4mm} \arrayrulecolor{black}\hline \noalign{\global\arrayrulewidth=0.2mm}
	\end{tabular}
\end{table*}

\section*{Expanded Coreset Selection Efficiency Details}
We provide a detailed comparison of coreset selection efficiency in \cref{tab:efficient}.
In terms of runtime efficiency, ZCore selects data using embeddings generated from a single forward pass of off-the-shelf models, while TDDS and ELFS, which are representative of current SOTA methods, train on the entire dataset prior to coreset selection.
However, training on data prior to selection takes substantially \textit{more} time than any other process, whether using a laptop or GPU (see TDDS ``Train Dynamics," 52.3 hours and 1,540\textrm{s} respectively).
Furthermore, ELFS requires repeat full data deep clustering \textit{and} model training on pseudo labels to calculate scores for each prune rate.
On the other hand, generating embeddings prior to selection is a one-time and parallizable operation that takes substantially less time than deep clustering, pretraining, or downstream model training, whether on the laptop or GPU (see ZCore ``Embedding Generation," 1.8 hours and 145\textrm{s} respectively).

From the results in \cref{tab:efficient}, ZCore selects the three coresets 27.5$\times$ faster than TDDS on a laptop and 39.6$\times$ faster than ELFS on a GPU.
Importantly, as discussed in \cref{sec:eff}, the relative time to implement ZCore is even faster.
Reason being, ZCore uses the same settings across all datasets and the same score across all prune rates.
On the other hand, both comparison methods perform parameter searches on \textit{each} dataset \textit{and} prune rate to determine their final dataset- and prune-specific settings.
It is cost-prohibitive to replicate the hyperparameter search of each comparison method for our experiments in \cref{tab:efficient}, but ELFS's developers report that it takes approximately 17 hours using four A6000 GPUs to find a single prune rate setting on ImageNet \cite{zheng2025elfs}.

In terms of label efficiency, TDDS, like many SOTA methods, uses ground truth labels on all candidate data prior to coreset selection.
On the other hand, ZCore and ELFS select data without any ground truth labels.
Accordingly, labels are only required during downstream model training for the selected coreset if the model is fully-superivsed.
Thus, TDDS requires 1.43, 3.33, and 10$\times$ more labels than ZCore to train downstream models at the respective 70\%, 30\%, and 10\% data selection rates in \cref{tab:efficient}.

\setlength{\tabcolsep}{5.45pt}
\begin{table*}[t!]
	\centering
	\caption{{\bf ZCore Coreset Class Recall} after selection on all datasets. ZCore achieves 100\% class recall in all the main paper's experiments ({\bf 10-70\%}, middle), so we include a few minimal selection rates as an added challenge ({\bf 0.5-5\%}, right). Recall \% is mean across all trials.
	}
	\vspace{-0.875em}
	\label{tab:recall}
	\small
	\begin{tabular}{l  c c c c c c c c c c c c c c c}
		\noalign{\global\arrayrulewidth=0.4mm} \hline \noalign{\global\arrayrulewidth=0.2mm} \arrayrulecolor{gray}
		&& && && \multicolumn{5}{c}{Class Recall (\%) at} && \multicolumn{4}{c}{Class Recall (\%) at}  \rule{0pt}{2ex} \\ 
		\multicolumn{1}{c}{} & &  Number of && Full Dataset && \multicolumn{5}{c}{Original Data Selection Rates ({\bf 10-70\%}) } && \multicolumn{4}{c}{Minimal Data Rates ({\bf 0.5-5\%}) }  \rule{0pt}{0ex} \\ 
		\multicolumn{1}{c}{Dataset} && Classes &&  Training Size && \bf 70\% & \bf 50\% & \bf 30\% & \bf 20\% & \bf 10\%  && \bf 5\% & \bf 2\% & \bf 1\% & \bf 0.5\%  \rule{0pt}{0ex} \\ 
		\noalign{\global\arrayrulewidth=0.4mm} \arrayrulecolor{white}\hline \noalign{\global\arrayrulewidth=0.2mm} \arrayrulecolor{gray}
		\cline{1-1} \cline{3-3} \cline{5-5} \cline{7-11} \cline{13-16}
		ImageNet	&&	1,000	&&	1,281,167	&&	100\%	&	100\%	&	100\%	&	100\%	&	100\%	&&	100\%	&	100\%	&	100\%  & 100\% \rule{0pt}{2ex}	\\
		CIFAR100	&&	100	&&	50,000	&&	100\%	&	100\%	&	100\%	&	100\%	&	100\%	&&	100\%	&	99\%	&	92\% & 75\%	\\
		CIFAR10	&&	10	&&	50,000	&&	100\%	&	100\%	&	100\%	&	100\%	&	100\%	&&	100\%	&	100\%	&	100\% & 100\%	\\
		EuroSAT80	&&	10	&&	21,600	&&	100\%	&	100\%	&	100\%	&	100\%	&	100\%	&&	100\%	&	100\%	&	96\% & 88\%	\\
		EuroSAT40	&&	10	&&	10,800	&&	100\%	&	100\%	&	100\%	&	100\%	&	100\%	&&	100\%	&	90\%	&	86\% & 78\%	\\
		EuroSAT20	&&	10	&&	5,400	&&	100\%	&	100\%	&	100\%	&	100\%	&	100\%	&&	98\%	&	90\%	&	76\% & 66\%	\\
		EuroSAT10	&&	10	&&	2,700	&&	100\%	&	100\%	&	100\%	&	100\%	&	100\%	&&	92\%	&	82\%	&	64\% & 58\%	\\
		\noalign{\global\arrayrulewidth=0.4mm} \arrayrulecolor{black}\hline \noalign{\global\arrayrulewidth=0.2mm}
	\end{tabular}
\end{table*}

\section*{Class Recall after ZCore Selection}
We provide the class recall of ZCore selections across all datasets and data selection rates of the main paper in \cref{tab:recall}.
Notably, ZCore achieves 100\% recall across all datasets and original data selection rates (10-70\%).
In addition to the main paper's quantitative and qualitative results (\cref{sec:quant,sec:qual}), ZCore's reliable class recall further establishes the viability of using iterative subspace sampling (\cref{sec:method}) to achieve coreset data coverage without labels or training.

As an additional challenge, we also provide the class recall of ZCore selections on a few minimal data selection rates beyond those of the main paper (0.5-5\%).
Notably, ZCore maintains 100\% class recall all the way down to a 0.5\% data selection rate on CIFAR10 and even ImageNet, which has 1,000 classes.
On the other hand, class recall drops below 100\% for datasets with fewer initial images per class (e.g., CIFAR100 recall is 99\% at a 2\% data rate) or small datasets with much smaller corresponding coresets (EuroSAT10, EuroSAT20 at 5\% data rates). 
Predictably, the lowest recall of 58\% occurs on EuroSAT10 at a 0.5\% data rate, which is selecting for a coreset of only 14 images.

\setlength{\tabcolsep}{2.35pt}
\renewcommand{\arraystretch}{1}
\begin{table}
	\centering
	\caption{Comparison of coreset selection methods using downstream model validation on {\bf ImageNet}.
		Full dataset training on the ResNet32 model training achieves 73.54\% accuracy.
		A corresponding results plot is provided in \cref{fig:cifar} of the main paper.
	}
	\vspace{-0.875em}
	\label{tab:imagenet}
	\begin{tabular}{l  c c c c c c c c c c}
		\noalign{\global\arrayrulewidth=0.4mm} \hline \noalign{\global\arrayrulewidth=0.2mm} \arrayrulecolor{gray}
		\multicolumn{1}{c}{\% of Data Selected} &&   30\% & 20\% & 10\%  & & $\methodlat{\text{Mean}}{\text{\scriptsize ~/ Rel. Rand.}}$  \\ 
		\cline{1-1} \cline{3-5} \cline{7-7} 
		\multicolumn{7}{l}{\textcolor{gray}{ \small \bf Unlabeled Coreset Selection without Training}} \rule{0pt}{2.5ex} \\ 
		{\bf ZCore}$_{\text{\tiny CLIP}}$ &&  \bf 64.42	&	\bf 61.39	&	\bf 53.54	& & $\methodlat{\text{\bf 59.78}}{\text{~+0.59}}$\rule{0pt}{2ex} \\
		Random &&  64.19	&	60.76	&	52.63	& & $\methodlat{\text{59.19}}{\text{~+0.00}}$ \\
		\multicolumn{7}{l}{\textcolor{gray}{ \small \bf Unlabeled Coreset Selection with Self-Supervised Training}} \rule{0pt}{2.5ex} \\ 
		Prototypes$_{\text{\tiny SS}}$ \cite{Sorscher_NEURIPS2022} &&  60.42 &	53.73	&	38.06	& & $\methodlat{\text{50.74}}{\text{~-8.45}}$ \\
		\multicolumn{7}{l}{\textcolor{gray}{ \small \bf Labeled Coreset Selection with Training-based Pruning}} \rule{0pt}{2.5ex}  \\ 
		TDDS \cite{Zhang_2024_CVPR} &&  \bf 64.69	&	\bf 62.56	&	53.91	& & $\methodlat{\text{\bf 60.39}}{\text{~+1.19}}$\rule{0pt}{2ex} \\
		ZCore$_{\text{\tiny {\bf ResNet18}, CLIP}}$ &&  64.43	&	61.31	&	\bf 53.99	& & $\methodlat{\text{59.91}}{\text{~+0.72}}$\rule{0pt}{2ex} \\
		Forgetting \cite{toneva2018an} && 64.29	&	62.01	&	52.14	& & $\methodlat{\text{59.48}}{\text{~+0.29}}$ \\
		Moderate \cite{xia2023moderate} && 64.04	&	61.35	&	52.45	& & $\methodlat{\text{59.28}}{\text{~+0.09}}$ \\
		ZCore$_{\text{\tiny \bf ResNet18}}$ &&  63.37	&	60.57	&	52.97	& & $\methodlat{\text{58.97}}{\text{~-0.22}}$\rule{0pt}{2ex} \\
		Entropy \cite{Coleman2020Selection} && 62.34	&	56.80	&	43.39	& & $\methodlat{\text{54.18}}{\text{~-5.02}}$ \\
		Prototypes$_{\text{\tiny Sup.}}$ \cite{Sorscher_NEURIPS2022} && 53.59 &	42.25	& 22.41	& & $\methodlat{\text{~39.42}}{\text{~-19.77}}$ \\
		EL2N \cite{Paul_NEURIPS2021}  && 46.92	&	32.68	&	15.90	& & $\methodlat{\text{~31.83}}{\text{~-27.36}}$ \\
		AUM \cite{Pleiss_NEURIPS2020} && 39.34	&	23.64	&	11.70	& & $\methodlat{\text{~24.89}}{\text{~-34.30}}$ \\
		\noalign{\global\arrayrulewidth=0.4mm} \arrayrulecolor{black}\hline \noalign{\global\arrayrulewidth=0.2mm}
	\end{tabular}
\end{table}

\begin{figure}[t]
	\centering
	\begin{minipage}{0.5\textwidth}
		\input{imagenet_paper_zcore.tex}
	\end{minipage}%
	\vspace{-1em}
	\caption{Comparison of coreset selection using downstream model validation on {\bf ImageNet} for ZCore embedding ablations.
	}
	\label{fig:imagenet}
\end{figure}

\section*{ImageNet Embedding Ablations}
ZCore selects coreset data using embeddings from ``off-the-shelf" foundation models that were trained before this work and remain static.
Notably, a component of the default embeddings (ResNet18) is previously trained on ImageNet.
On the other hand, we made a methodological commitment to use constant ZCore settings across all datasets and selection rates in our experiments, which includes consistent foundation models for embeddings.

To understand the relative contribution of ResNet18 embeddings for coreset selection on ImageNet, we perform an additional ablative experiment with ResNet18- and CLIP-only embeddings (see ZCore$_{\text{\tiny ResNet18}}$ and ZCore$_{\text{\tiny CLIP}}$ in \cref{tab:imagenet} and \cref{fig:imagenet}).
Notably, ZCore selections using CLIP-only embeddings are outperformed by \textit{all} other embedding settings on CIFAR100 (see \cref{tab:ablate} in the main paper).
On the other hand, ZCore$_{\text{\tiny CLIP}}$ selections on ImageNet outperform ZCore$_{\text{\tiny ResNet18}}$ across all data selection rates, despite ResNet18 pre-training on ImageNet.
Furthermore, ZCore$_{\text{\tiny CLIP}}$ performs comparably to the concatenated ZCore$_{\text{\tiny ResNet18, CLIP}}$ setting on ImageNet, validating that ZCore is state-of-the-art for coreset selection without labels or training on candidate data across all experiment settings.

\setlength{\tabcolsep}{1.1pt}
\begin{table*}
	\centering
	\caption{Comparison of {\bf ZCore ablations} on CIFAR 100 with ResNet18. 
		Accuracy is the mean across 70\%, 50\%, 30\%, 20\%, and 10\% data rates over five repeat trials.
		``ResNet18, CLIP" is the default concatenated embedding space using both models off-the-shelf.
	}
	\vspace{-0.875em}
	\label{tab:fullablate}
	\footnotesize
	\begin{tabular}{l c  c c c c c c c c c c c c c c c c c c}
		\noalign{\global\arrayrulewidth=0.4mm} \hline \noalign{\global\arrayrulewidth=0.2mm}
		\noalign{\global\arrayrulewidth=0.4mm} \arrayrulecolor{white}\hline \noalign{\global\arrayrulewidth=0.2mm} \arrayrulecolor{gray}
		& &  Zero-shot  &&  Dimension && \multicolumn{3}{c}{Sampling} && Distance && \multicolumn{3}{c}{Redundancy} && Use Full && CIFAR100  \\
		\multicolumn{1}{c}{\small \bf Ablation} & &  Embedding Model && Reduction && Distribution && Dimensions && Metric &&  Neighbors && Exponent && Score && Accuracy \\
		\noalign{\global\arrayrulewidth=0.4mm} \arrayrulecolor{white}\hline \noalign{\global\arrayrulewidth=0.2mm} \arrayrulecolor{gray}
		\cline{1-1} \cline{3-3} \cline{5-5}  \cline{7-7} \cline{9-9}  \cline{11-11} \cline{13-13} \cline{15-15} \cline{17-17}
		Full Method ({\bf ZCore}) & &  \bf ResNet18, CLIP && \boldmath$\mZ \mD$ && \bf Triangular & & \bf 2 &&  \boldmath$L_1$ && \bf 1,000 && \bf 4 && \bf Yes && \bf 65.77 \rule{0pt}{2ex} \\ 
		\hline
		Embedding Space {\color{gray} ($\mZ$)} & & \bf ResNet18 \cite{HeEtAl16} && \color{gray} \boldmath$\mZ \mD$ & & \color{gray} Triangular & & \color{gray} 2 && \color{gray} $L_1$ && \color{gray}1,000&& \color{gray} 4 && \color{gray} Yes && 65.52 \rule{0pt}{2ex} \\
		\cline{3-3} \cline{19-19}
		& & \bf CLIP \cite{clip} && \color{gray} \boldmath$\mZ \mD$ & & \color{gray} Triangular & & \color{gray} 2 && \color{gray} $L_1$ && \color{gray}1,000 && \color{gray} 4 && \color{gray} Yes && 64.84 \rule{0pt}{2ex} \\
		\cline{3-3} \cline{19-19}
		& & \bf DINOv2 \cite{oquab2024dinov}&& \color{gray} \boldmath$\mZ \mD$  & & \color{gray} Triangular & & \color{gray} 2 && \color{gray} $L_1$ && \color{gray}1,000 && \color{gray} 4 && \color{gray} Yes &&  65.61 \rule{0pt}{2ex} \\
		\hline 
		Embedding Dimension Reduction {\color{gray} ($\hat{\mZ}$)} & & \color{gray} ResNet18, CLIP && \bf PCA & & \color{gray} Triangular & & \color{gray} 2 && \color{gray} $L_1$ && \color{gray}1,000&& \color{gray} 4 && \color{gray} Yes && 65.55 \rule{0pt}{2.5ex} \\
		\cline{5-5} \cline{19-19}
		& & \color{gray} ResNet18, CLIP && \bf t-SNE \cite{maaten2008visualizing} & & \color{gray} Triangular & & \color{gray} 2 && \color{gray} $L_1$ && \color{gray}1,000 && \color{gray} 4 && \color{gray} Yes && 65.62 \rule{0pt}{2ex} \\
		\cline{5-5} \cline{19-19}
		& & \color{gray} ResNet18, CLIP && \bf UMAP \cite{mcinnes2018umap-software} & & \color{gray} Triangular & & \color{gray} 2 && \color{gray} $L_1$ && \color{gray}1,000 && \color{gray} 4 && \color{gray} Yes && 62.43 \rule{0pt}{2ex} \\
		\hline
		Sampling Distribution {\color{gray} ($\rvz$)} & & \color{gray} ResNet18, CLIP && \color{gray} \boldmath$\mZ \mD$ & & \bf Gaussian & & \color{gray} 2 && \color{gray} $L_1$ && \color{gray}1,000&& \color{gray} 4 && \color{gray}Yes && 65.75 \rule{0pt}{2ex} \\
		\cline{7-7} \cline{19-19}
		& & \color{gray} ResNet18, CLIP && \color{gray} \boldmath$\mZ \mD$ & & \bf Uniform & & \color{gray} 2 && \color{gray} $L_1$ && \color{gray}1,000 && \color{gray} 4&& \color{gray}Yes && 64.84 \rule{0pt}{2ex} \\
		\hline
		Subspace Sampling Dimensions {\color{gray} ($m$)}  & &  \color{gray} ResNet18, CLIP && \color{gray} \boldmath$\mZ \mD$ & & \color{gray} Triangular & & \bf 1 && \color{gray} $L_1$ && \color{gray}1,000&& \color{gray} 4&& \color{gray}Yes && 64.59 \rule{0pt}{2ex} \\
		\cline{9-9} \cline{19-19}
		& & \color{gray} ResNet18, CLIP && \color{gray} \boldmath$\mZ \mD$ & &\color{gray} Triangular & & \bf  3 && \color{gray} $L_1$ && \color{gray}1,000&& \color{gray} 4 && \color{gray}Yes && 65.10 \rule{0pt}{2ex} \\
		\cline{9-9} \cline{19-19}
		\small  & & \color{gray} ResNet18, CLIP && \color{gray} \boldmath$\mZ \mD$ & & \color{gray} Triangular & & \bf 10 && \color{gray} $L_1$ && \color{gray}1,000&& \color{gray} 4 && \color{gray}Yes && 63.27 \rule{0pt}{2ex} \\
		\cline{9-9} \cline{19-19}
		\small  & & \color{gray} ResNet18, CLIP && \color{gray} \boldmath$\mZ \mD$ & & \color{gray} Triangular & & \bf 100 && \color{gray} $L_1$ && \color{gray}1,000&& \color{gray} 4 && \color{gray}Yes && 62.20 \rule{0pt}{2ex} \\ \hline
		Distance Metric {\color{gray}(\cref{eq:mces}, \cref{eq:vr})} && \color{gray} ResNet18, CLIP && \color{gray} \boldmath$\mZ \mD$ && \color{gray} Triangular && \color{gray} 2 && \boldmath$L_2$ && \color{gray}1,000 && \color{gray} 4 && \color{gray}Yes &&  65.60 \rule{0pt}{2ex} \\
		\cline{11-11} \cline{19-19}
		& & \color{gray} ResNet18, CLIP && \color{gray} \boldmath$\mZ \mD$ && \color{gray} Triangular && \color{gray} 2 && \boldmath$\cos$ && \color{gray}1,000 && \color{gray} 4 && \color{gray}Yes &&  64.48 \rule{0pt}{2ex} \\ \hline
		Number of Nearest  Neighbors {\color{gray}($\alpha$)} & & \color{gray} ResNet18, CLIP && \color{gray} \boldmath$\mZ \mD$ & & \color{gray} Triangular & & \color{gray} 2 && \color{gray} $L_1$ && \bf 100&& \color{gray} 4 && \color{gray}Yes &&  65.68 \rule{0pt}{2ex} \\
		\cline{13-13} \cline{19-19}
		& & \color{gray} ResNet18, CLIP && \color{gray} \boldmath$\mZ \mD$ & & \color{gray} Triangular & & \color{gray} 2 && \color{gray} $L_1$ && \bf 10,000&& \color{gray} 4 && \color{gray}Yes &&  65.72 \rule{0pt}{2ex} \\ \hline
		Distance Penalty Exponent {\color{gray}($\beta$)} & & \color{gray} ResNet18, CLIP && \color{gray} \boldmath$\mZ \mD$ & & \color{gray} Triangular & & \color{gray} 2 && \color{gray} $L_1$ && \color{gray}1,000&& \bf 3 && \color{gray}Yes &&  65.64 \rule{0pt}{2ex} \\
		\cline{15-15} \cline{19-19}
		& & \color{gray} ResNet18, CLIP && \color{gray} \boldmath$\mZ \mD$ & & \color{gray} Triangular & & \color{gray} 2 && \color{gray} $L_1$ && \color{gray}1,000&& \bf 5 && \color{gray}Yes &&  65.74 \rule{0pt}{2ex} \\ \hline
		Random Coverage Sample {\color{gray}($k$)} & & \bf NA & & \bf NA & & \bf NA & & \bf NA && \color{gray} $L_1$ && \color{gray}1,000&& \color{gray} 4 && \bf No && 64.74 \rule{0pt}{2ex} \\
		\cline{17-17} \cline{19-19}
		No Redundancy Score {\color{gray} ($\vs^\text{\tiny R}$)} & & \color{gray} ResNet18, CLIP && \color{gray} \boldmath$\mZ \mD$ & & \color{gray} Triangular & & \color{gray} 2 && \color{gray} $L_1$ && \bf NA && \bf NA&& \bf No && 61.71 \rule{0pt}{2ex} \\
		\cline{17-17} \cline{19-19}
		No Random Initialization {\color{gray} ($\rvs$)} & & \color{gray} ResNet18, CLIP && \color{gray} \boldmath$\mZ \mD$ & & \color{gray} Triangular & & \color{gray} 2 && \color{gray} $L_1$ && \color{gray}1,000&& \color{gray} 4 && \bf No && 65.66 \rule{0pt}{2ex} \\
		\cline{17-17} \cline{19-19}
		No Score Loss Weight {\color{gray} ($\vw$)} & & \color{gray} ResNet18, CLIP && \color{gray} \boldmath$\mZ \mD$ & & \color{gray} Triangular & & \color{gray} 2 && \color{gray} $L_1$ && \color{gray}1,000&& \color{gray} 4 && \bf No && 63.53 \rule{0pt}{2ex} \\
		\noalign{\global\arrayrulewidth=0.4mm} \arrayrulecolor{black}\hline \noalign{\global\arrayrulewidth=0.2mm}
	\end{tabular}
\end{table*}

\section*{Embedding Dimension Reduction Ablations}
For dimension reduction, ZCore iteratively slices random subsets of the full embedding space to generate numerous subspace distributions, which are then sampled to find valuable examples based on coverage and redundancy (see \cref{fig:front} and \cref{sec:method} of the main paper for details).
However, there are several alternative dimension reduction methods that are applicable to this work, including Principal Component Analysis (PCA), t-distributed Stochastic Neighbor Embedding (t-SNE) \cite{maaten2008visualizing}, and Uniform Manifold Approximation and Projection (UMAP) \cite{mcinnes2018umap-software}.
Notably, PCA maximizes variance to preserve pairwise global embedding distances while t-SNE and UMAP better preserve local embedding distances (see McInnes \etal for additional details and experiment comparisons \cite{mcinnes2018umap-software}).

\begin{figure}      
	\centering
	\includegraphics[width=0.475\textwidth]{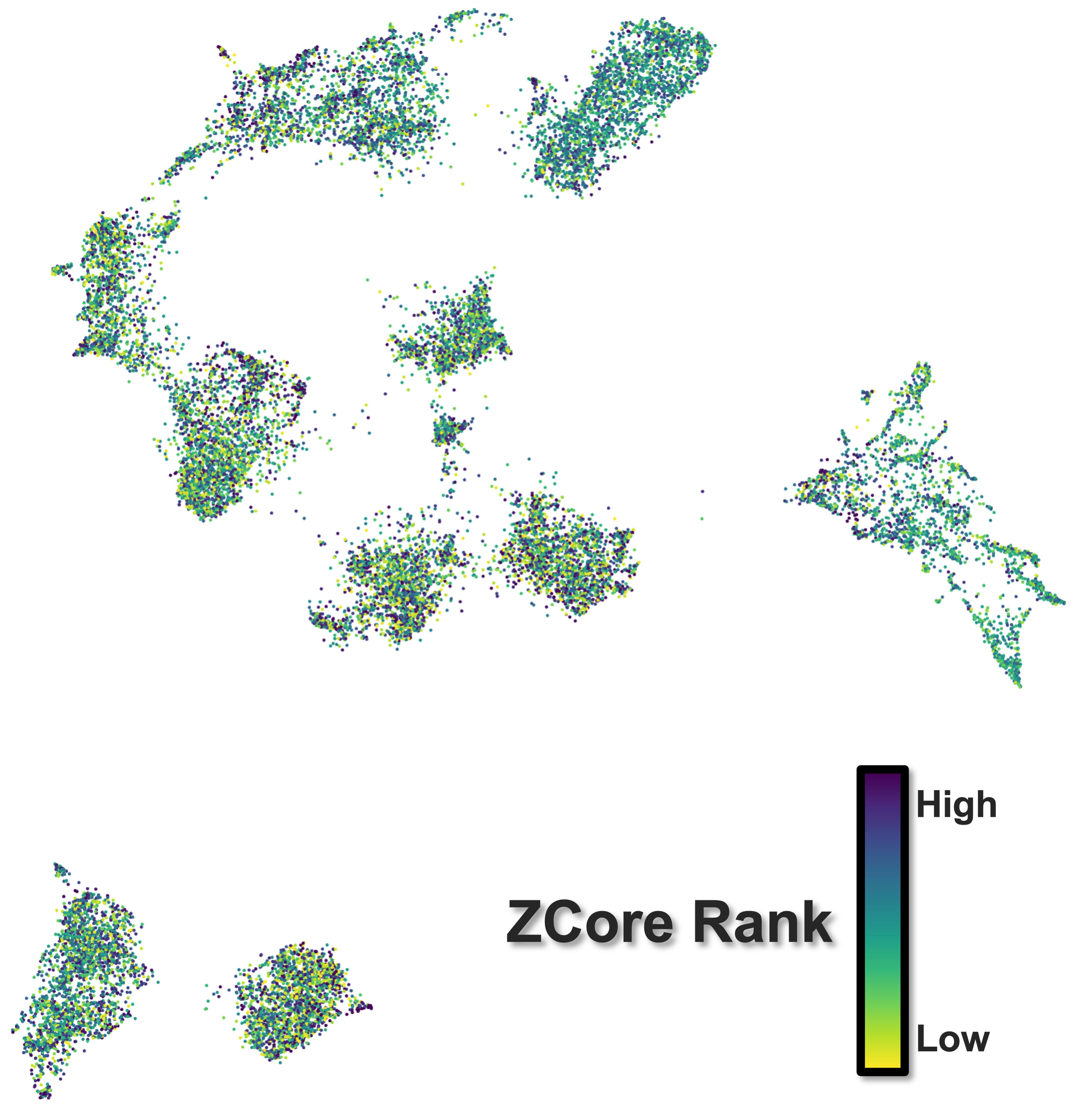}
	\vspace{-2em}
	\caption{\textbf{ZCore Coreset Rank Visualization} for EuroSAT80. 
		Model embeddings and 2D approximation of the full embedding space generated using the FiftyOne Library and UMAP \cite{moore2020fiftyone,mcinnes2018umap-software}.
	}
	\label{fig:euro100score}
\end{figure}

To validate the design decision to use ZCore's iterative embedding dimension reduction technique, we provide experiment ablations using PCA, t-SNE, and UMAP as alternative embedding reduction processes for subsequent sampling and ZCore score generation in \cref{tab:fullablate}.
Notably, the proposed ZCore dimension reduction technique ($\hat{\mZ}=\mZ \mD$ in \cref{eq:zhat}) outperforms all three dimension reduction alternatives.
Nonetheless, the t-SNE and PCA ablations exhibit good performance, indicating that the subsequent ZCore processes for coverage and redundancy scoring are relatively robust to changes in the underlying dimension reduction technique used.
From the results, we do not recommend UMAP as an alternative dimension reduction technique for ZCore, but we still find UMAP to be an excellent tool for embedding space visualization (see \cref{fig:euro100score}).

\section*{Supplementary Figures \& Tables}
To supplement the evaluation in \cref{sec:eval} of the main paper, we provide several additional figures and tables.

\begin{figure*}      [t!]
	\centering
	\includegraphics[width=0.995\textwidth]{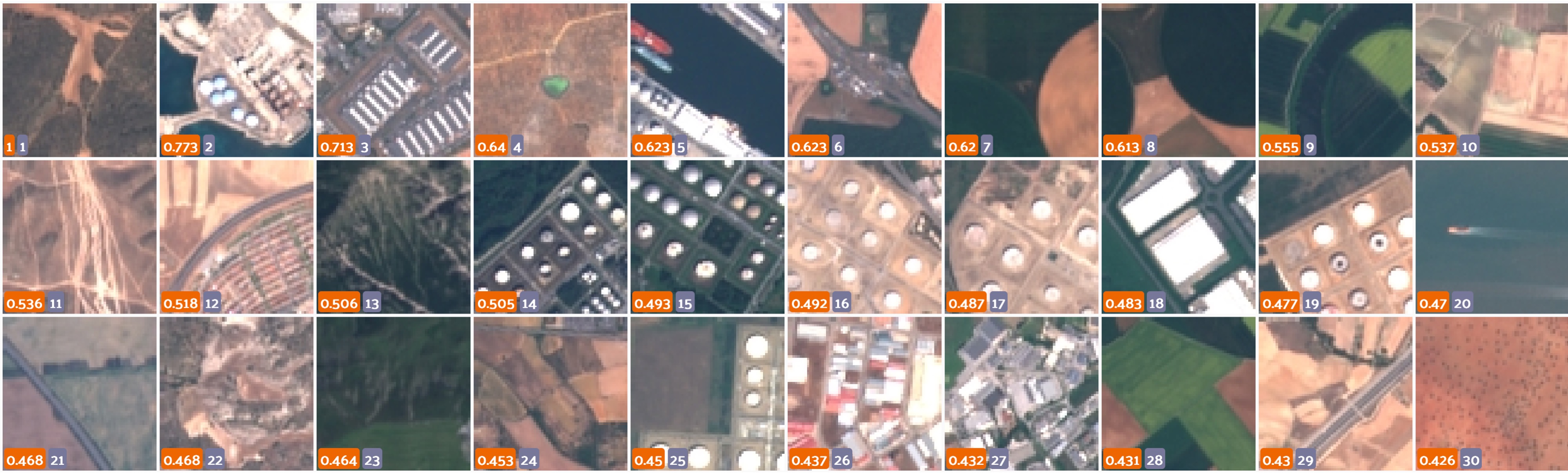}
	\vspace{-0.9em}
	\caption{{\bf Top-30 ZCore Selection Scores} (orange) for EuroSAT80.
	}
	\label{fig:euro30}
\end{figure*}

\begin{figure*} [t!]
	\centering
	\includegraphics[width=0.995\textwidth]{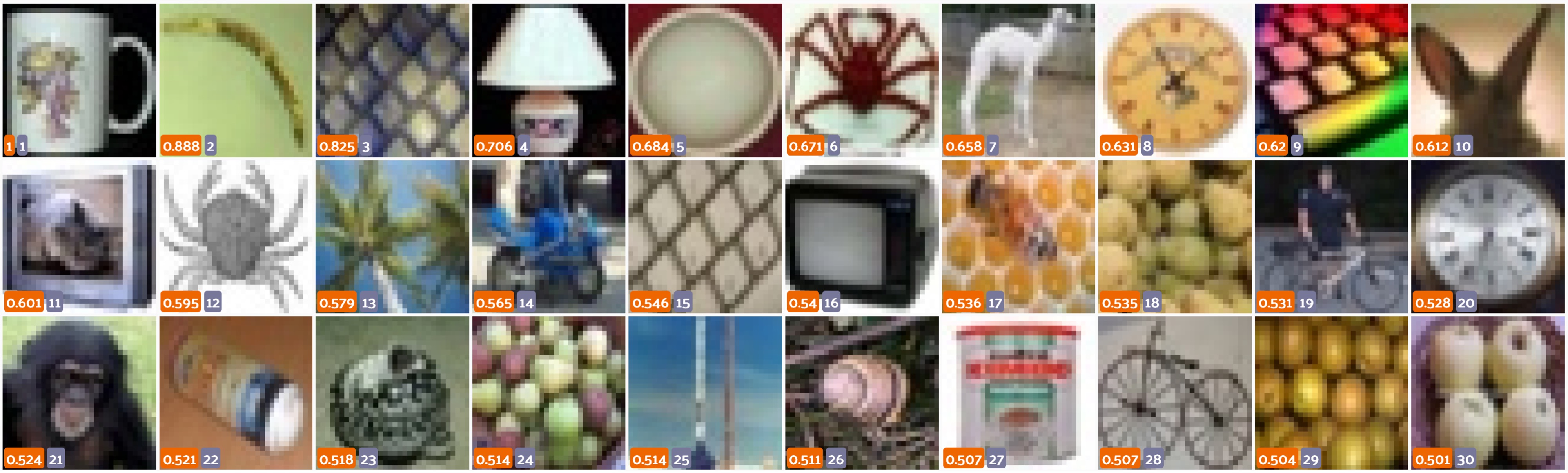}
	\vspace{-0.9em}
	\caption{{\bf Top-30 ZCore Selection Scores} (orange) for CIFAR100.
	}
	\label{fig:cifar30}
\end{figure*}

\begin{figure*}
	\centering
	\includegraphics[width=0.995\textwidth]{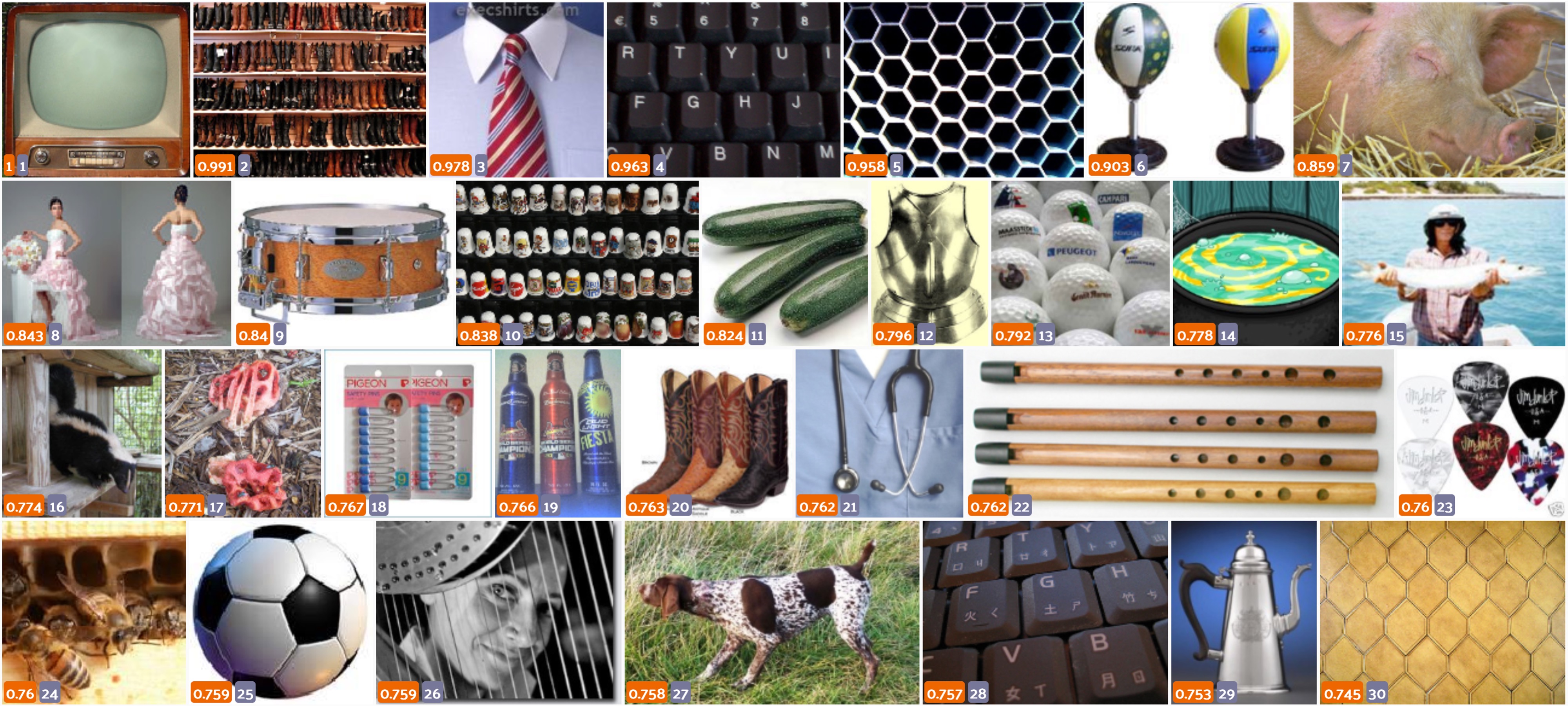}
	\vspace{-0.9em}
	\caption{{\bf Top-30 ZCore Selection Scores} (orange) for ImageNet.
	}
	\label{fig:imagenet30}
\end{figure*}

\begin{figure*}
	\centering
	\includegraphics[width=0.995\textwidth]{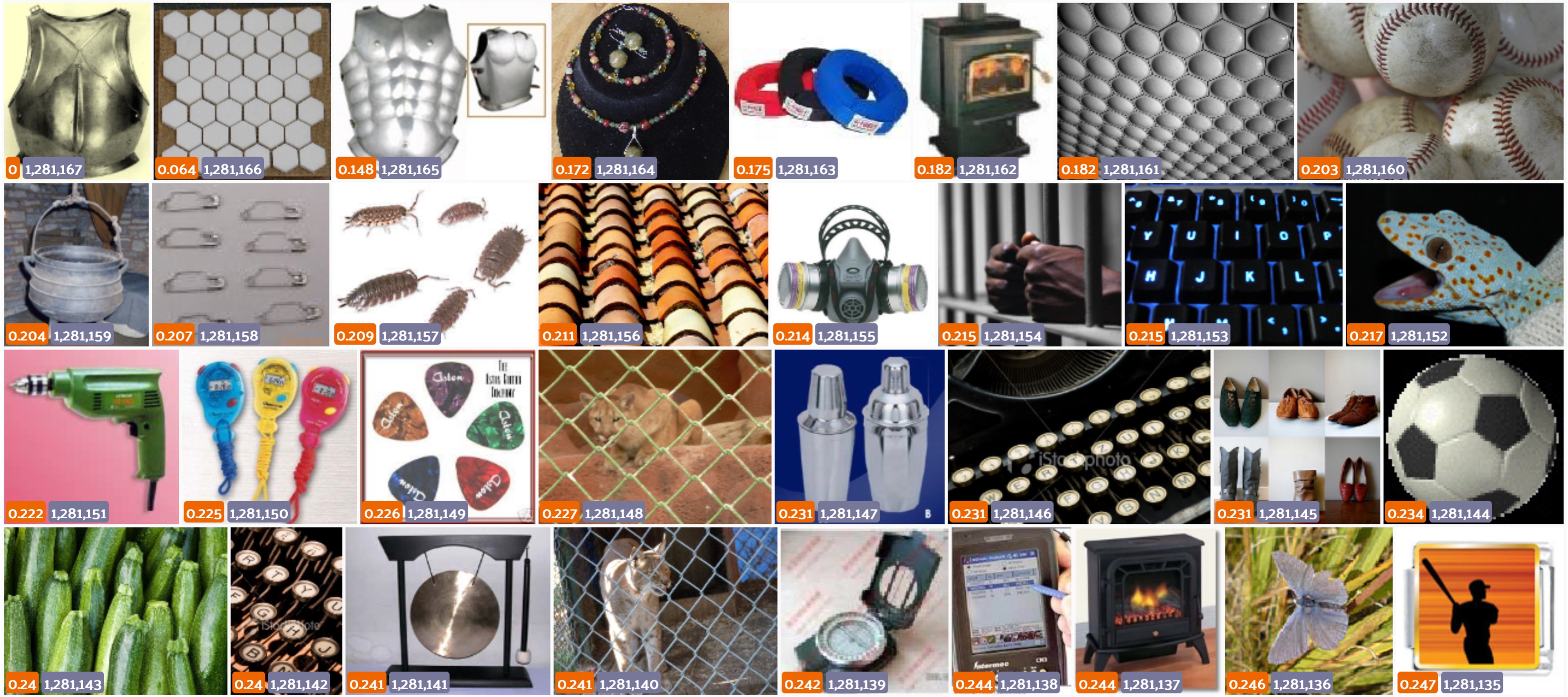}
	\vspace{-0.9em}
	\caption{{\bf Lowest-33 ZCore Selection Scores} (orange) for ImageNet.
	}
	\label{fig:imagenet30bad}
\end{figure*}

First, we show the entire ZCore rankings for EuroSAT80 within a 2D manifold approximation of the full embedding space in \cref{fig:euro100score}.
Second, we plot self-supervised DINOv2 model embeddings and corresponding sample distributions in \cref{fig:dino-dist}.
Third, we plot the runtime and accuracy performance of ZCore over a range of subspace sampling dimensions in \cref{fig:dimruntime}.
Fourth, we show the highest 30 ranked ZCore images for EuroSAT80 in \cref{fig:euro30}, CIFAR100 in \cref{fig:cifar30}, and ImageNet in \cref{fig:imagenet30}.
We also show the lowest 30 ranked ZCore images for ImageNet in \cref{fig:imagenet30bad}.
Finally, we tabulate coreset selection results for CIFAR10 and CIFAR100 in \cref{tab:cifar} (evaluating selection for two medium-sized datasets), ImageNet in \cref{tab:imagenet} (evaluating selection at a large scale), and all EuroSAT splits in \cref{tab:euro} (evaluating selection of satellite images at a decreasing scale).

\section*{Future Work}

For downstream model training experiments in \cref{sec:quant}, ZCore outperforms all prior coreset selection methods except for the state-of-the-art TDDS method.
However, TDDS performance relies on several requirements that, although useful in experiment settings, limit applicability and efficiency at scale.  
Specifically, TDDS assumes all candidate data are labeled prior to selection (which is cost prohibitive), performs fully-supervised model training on all candidate data prior to selection (which is runtime prohibitive), and uses a hyperparameter search on the validation set to determine the final configuration used on a per-dataset and per-prune rate basis (which limits generalization).
On the other hand, ZCore uses no labels or training on candidate data prior to selection (which is label \textit{and} runtime efficient) and uses constant algorithmic settings across all experiments (which is generalizable).
Despite these procedural differences, ZCore maintains competitive performance and even outperforms TDDS at a 10\% selection rate on EuroSAT40 and ImageNet.
Nonetheless, there remain several promising methodological avenues to improve ZCore and further advance the viability of coreset selection without labels, training, or parameter tuning.

First, although ZCore performance degrades relative to TDDS on the small EuroSAT20 and EuroSAT10 datasets at low selection rates, we find that ZCore performance improves with alternative settings (e.g., reducing the number of nearest neighbors for redundancy in \cref{eq:vr}). 
Thus, one area of future work is developing adaptive iterative subspace sampling so that ZCore automatically adjusts with dataset size and data selection rate to improve performance in challenge settings.

Second, in addition to the coverage and redundancy scores developed in this paper, we postulate that there are many more label- and training-free features and metrics that can be developed to quantify coreset value for individual candidate examples.
Expanding the set of metrics ZCore uses to consider each data example will likely make ZCore coreset selection more robust and performant across a wide range of datasets and data selection rates. 

Finally, there is no domain-specific limitation to our method, so we can apply ZCore in other domains like point cloud and natural language and other problems like object detection and segmentation.
However, each new domain and problem space will present new challenges \textit{and} opportunities to improve ZCore.
Taking object detection as one example, ZCore currently selects data using image-level embeddings but can also operate on object-level embeddings that are more representative of what downstream models will use for object detection.
Furthermore, guiding data selection and annotation at an object level can lead to massive improvements in terms of overall human annotation time and downstream model performance \cite{Lyu_2023_CVPR}.
Thus, we find that there are many exciting opportunities to expand and improve upon ZCore coreset selection in future work.

\begin{figure}
	\centering
	\begin{minipage}{0.5\textwidth}
		\input{dinov2_0.tex}
	\end{minipage}%
	\vspace{-1.1em}
	\caption{\textbf{Comparison of embeddings and sampling techniques}. 
		DINOv2 is the first dimension embeddings for 50,000 CIFAR100 train set examples, while each corresponding distribution type is sampled 50,000 times.
		Corresponding plots for ResNet18 and CLIP are provided in \cref{fig:dist} of the main paper.
	}
	\label{fig:dino-dist}
\end{figure}

\begin{figure}
	\centering
	\begin{minipage}{0.5\textwidth}
		\input{nembdim_accuracy_paper_vertical.tex}
	\end{minipage}%
	\\
	\vspace{-0.25em}
	\begin{minipage}{0.5\textwidth}
		\input{nembdim_runtime_paper_vertical.tex}
	\end{minipage}%
	\vspace{-1.1em}
	\caption{Comparison of number of {\bf embedding subspace sample dimensions} ($m$) vs. accuracy (top) and {\bf runtime} (bottom) on CIFAR100 with ResNet18.
		Accuracy is the mean across 70\%, 50\%, 30\%, 20\%, and 10\% data selection rates over five repeat trials.
		Runtime is selection time using a M3 Max-equipped laptop.
	}
	\label{fig:dimruntime}
\end{figure}



\setlength{\tabcolsep}{3.85pt}
\renewcommand{\arraystretch}{1}
\begin{table*}
	\centering
	\caption{Comparison of coreset selection methods using downstream model validation on {\bf CIFAR10} and {\bf CIFAR100}. 
		Full dataset training on the ResNet18 model achieves 95.23\% (CIFAR10) and 78.21\% (CIFAR100) accuracy.
		``Rel. Rand.'' is Mean accuracy across all data selection rates on both datasets relative to Random.
		A corresponding results plot is provided in \cref{fig:cifar} of the main paper.
	}
	\vspace{-0.875em}
	\label{tab:cifar}
	\begin{tabular}{l  c  c c c c c c c c c c c c c}
		\noalign{\global\arrayrulewidth=0.4mm} \hline \noalign{\global\arrayrulewidth=0.2mm}
		\multicolumn{1}{c}{} &  & \multicolumn{5}{c}{\bf CIFAR10} & & \multicolumn{5}{c}{\bf CIFAR100} \rule{0pt}{2ex} \\ 
		\arrayrulecolor{gray} 
		\multicolumn{1}{c}{\% Data Selected} && 70\% & 50\% & 30\% & 20\% & 10\%  & & 70\% & 50\% & 30\% & 20\% & 10\% & & $\meanrelrand{\text{Mean}}{\text{Rel. Rand.}}$   \rule{0pt}{0ex} \\ 
		\noalign{\global\arrayrulewidth=0.4mm} \arrayrulecolor{white}\hline \noalign{\global\arrayrulewidth=0.2mm} \arrayrulecolor{gray}
		\cline{1-1} \cline{3-7} \cline{9-13}  \cline{15-15}
		\multicolumn{13}{l}{\textcolor{gray}{ \small \bf Unlabeled Coreset Selection without Training}} \rule{0pt}{2.5ex} \\ 
		$\methodpub{\text{\bf	ZCore}}{}$ 	& & $\stddev{\text{	\bf 94.58}}{\text{0.09}}$ & $\stddev{\text{	\bf 93.46}}{\text{0.16}}$ & $\stddev{\text{	\bf 90.97}}{\text{0.17}}$ & $\stddev{\text{	\bf 89.06}}{\text{0.33}}$ & $\stddev{\text{	\bf 84.18}}{\text{0.21}}$ && $\stddev{\text{	\bf76.04 }}{\text{0.15}}$ & $\stddev{\text{	\bf 72.87}}{\text{0.18}}$ & $\stddev{\text{	\bf 65.92}}{\text{0.15}}$ & $\stddev{\text{	\bf61.92 }}{\text{0.39}}$ & $\stddev{\text{	\bf 52.11}}{\text{0.66}}$ & & $\meanrelrand{\text{\bf 78.11}}{\text{\bf +1.34}}$  \rule{0pt}{3ex}  \\
		$\methodpub{\text{Random}}{}$ & & $\stddev{\text{	\bf 94.58}}{\text{0.04}}$ & $\stddev{\text{	93.38}}{\text{0.17}}$ & $\stddev{\text{	90.61}}{\text{0.44}}$ & $\stddev{\text{	88.87}}{\text{0.47}}$ & $\stddev{\text{	83.77}}{\text{0.26}}$ && $\stddev{\text{	75.53}}{\text{0.04}}$ & $\stddev{\text{	71.95}}{\text{0.16}}$ & $\stddev{\text{	64.59}}{\text{0.32}}$ & $\stddev{\text{	57.79}}{\text{0.24}}$ & $\stddev{\text{	46.68}}{\text{1.07}}$ & & $\meanrelrand{\text{76.78}}{\text{+0.00}}$ \stddevrow \\
		\multicolumn{13}{l}{\textcolor{gray}{ \small \bf Unlabeled Coreset Selection with Self-Supervised Training}} \rule{0pt}{2.5ex} \\ 
		$\methodpub{\text{ELFS \cite{zheng2025elfs}}}{\text{ICLR 2025}}$ & & $\stddev{\text{94.04}}{\text{0.14}}$ & $\stddev{\text{	 92.61}}{\text{0.26}}$ & $\stddev{\text{90.37}}{\text{0.07}}$ & $\stddev{\text{88.04}}{\text{0.38}}$ & $\stddev{\text{82.39}}{\text{0.35}}$ && $\stddev{\text{74.10}}{\text{0.15}}$ & $\stddev{\text{	70.36}}{\text{0.18}}$ & $\stddev{\text{	63.83}}{\text{0.56}}$ & $\stddev{\text{58.51}}{\text{0.20}}$ & $\stddev{\text{47.42}}{\text{1.32}}$ & & $\meanrelrand{\text{76.17}}{\text{\text{-}0.61}}$ \rule{0pt}{3ex} \\
		\multicolumn{13}{l}{\textcolor{gray}{ \small \bf Labeled Coreset Selection with Training-based Pruning}} \rule{0pt}{3ex}  \\ 
		$\methodpub{\text{TDDS \cite{Zhang_2024_CVPR}}}{\text{CVPR 2024}}$ & & $\stddev{\text{	\bf 95.47}}{\text{0.06}}$ & $\stddev{\text{	\bf 95.21}}{\text{0.04}}$ & $\stddev{\text{	\bf 93.03}}{\text{0.25}}$ & $\stddev{\text{	\bf 91.30}}{\text{0.25}}$ & $\stddev{\text{	\bf 85.46}}{\text{0.21}}$ && $\stddev{\text{	\bf 77.56}}{\text{0.06}}$ & $\stddev{\text{	\bf 74.04}}{\text{0.34}}$ & $\stddev{\text{	\bf 67.78}}{\text{0.44}}$ & $\stddev{\text{	\bf 63.01}}{\text{0.12}}$ & $\stddev{\text{	\bf 54.51}}{\text{0.22}}$ & & $\meanrelrand{\text{\bf 79.74}}{\text{\bf +2.96}}$  \rule{0pt}{3ex} \\
		$\methodpub{\text{Moderate \cite{xia2023moderate}}}{\text{ICLR 2023}}$	& & $\stddev{\text{	93.96}}{\text{0.06}}$ & $\stddev{\text{	92.34}}{\text{0.09}}$ & $\stddev{\text{	89.71}}{\text{0.14}}$ & $\stddev{\text{	87.75}}{\text{0.27}}$ & $\stddev{\text{	83.61}}{\text{0.24}}$ && $\stddev{\text{	74.60}}{\text{0.10}}$ & $\stddev{\text{	70.29}}{\text{0.31}}$ & $\stddev{\text{	62.81}}{\text{0.08}}$ & $\stddev{\text{	56.52}}{\text{0.37}}$ & $\stddev{\text{	41.82}}{\text{1.12}}$ & & $\meanrelrand{\text{75.34}}{\text{\text{-}1.43}}$ \stddevrow \\
		$\methodpub{\text{Entropy \cite{Coleman2020Selection}}}{\text{ICLR 2020}}$	& & $\stddev{\text{	94.45}}{\text{0.07}}$ & $\stddev{\text{	91.90}}{\text{0.16}}$ & $\stddev{\text{	86.24}}{\text{0.26}}$ & $\stddev{\text{	83.49}}{\text{0.21}}$ & $\stddev{\text{	72.06}}{\text{0.81}}$ && $\stddev{\text{	72.39}}{\text{0.20}}$ & $\stddev{\text{	64.44}}{\text{0.36}}$ & $\stddev{\text{	50.73}}{\text{0.86}}$ & $\stddev{\text{	42.86}}{\text{0.25}}$ & $\stddev{\text{	29.56}}{\text{0.54}}$ & & $\meanrelrand{\text{68.81}}{\text{\text{-}7.96}}$ \stddevrow \\
		$\methodpub{\text{Forgetting \cite{toneva2018an}}}{\text{ICLR 2019}}$	& & $\stddev{\text{	95.45}}{\text{0.24}}$ & $\stddev{\text{	95.05}}{\text{0.05}}$ & $\stddev{\text{	89.14}}{\text{2.04}}$ & $\stddev{\text{	76.18}}{\text{3.18}}$ & $\stddev{\text{	45.87}}{\text{1.87}}$ && $\stddev{\text{	77.38}}{\text{0.09}}$ & $\stddev{\text{	70.76}}{\text{0.40}}$ & $\stddev{\text{	49.92}}{\text{0.28}}$ & $\stddev{\text{	38.42}}{\text{1.13}}$ & $\stddev{\text{	25.82}}{\text{0.52}}$ & & $\meanrelrand{\text{66.40}}{\text{\text{-}10.38}}$ \stddevrow \\
		$\methodpub{\text{Dyn-Unc  \cite{He_2024_CVPR}}}{\text{CVPR WS `24 }}$	& & $\stddev{\text{	95.08}}{\text{0.02}}$ & $\stddev{\text{	94.03}}{\text{0.14}}$ & $\stddev{\text{	89.40}}{\text{0.13}}$ & $\stddev{\text{	79.76}}{\text{1.09}}$ & $\stddev{\text{	37.12}}{\text{1.12}}$ && $\stddev{\text{	73.36}}{\text{0.10}}$ & $\stddev{\text{	65.90}}{\text{0.25}}$ & $\stddev{\text{	50.16}}{\text{0.47}}$ & $\stddev{\text{	39.19}}{\text{0.27}}$ & $\stddev{\text{	15.20}}{\text{0.41}}$ & & $\meanrelrand{\text{63.92}}{\text{\text{-}12.86}}$ \stddevrow \\
		$\methodpub{\text{AUM \cite{Pleiss_NEURIPS2020}}}{\text{NeurIPS 2020}}$	& & $\stddev{\text{	95.44}}{\text{0.09}}$ & $\stddev{\text{	95.19}}{\text{0.09}}$ & $\stddev{\text{	91.19}}{\text{0.63}}$ & $\stddev{\text{	69.60}}{\text{3.11}}$ & $\stddev{\text{	34.74}}{\text{0.11}}$ && $\stddev{\text{	77.35}}{\text{0.18}}$ & $\stddev{\text{	68.17}}{\text{0.52}}$ & $\stddev{\text{	31.69}}{\text{0.34}}$ & $\stddev{\text{	18.43}}{\text{0.47}}$ & $\stddev{\text{	9.29}}{\text{0.27}}$ & & $\meanrelrand{\text{59.11}}{\text{\text{-}17.67}}$ \stddevrow \\
		$\methodpub{\text{EL2N \cite{Paul_NEURIPS2021}}}{\text{NeurIPS 2021}}$	& & $\stddev{\text{	95.43}}{\text{0.10}}$ & $\stddev{\text{	95.06}}{\text{0.04}}$ & $\stddev{\text{	86.69}}{\text{1.71}}$ & $\stddev{\text{	68.64}}{\text{3.70}}$ & $\stddev{\text{	31.89}}{\text{1.51}}$ && $\stddev{\text{	76.89}}{\text{0.31}}$ & $\stddev{\text{	67.57}}{\text{0.15}}$ & $\stddev{\text{	36.45}}{\text{1.36}}$ & $\stddev{\text{	17.31}}{\text{0.33}}$ & $\stddev{\text{	9.10}}{\text{0.69}}$ & & $\meanrelrand{\text{58.50}}{\text{\text{-}18.27}}$ \stddevrow \\
		\noalign{\global\arrayrulewidth=0.4mm} \arrayrulecolor{white}\hline \noalign{\global\arrayrulewidth=0.2mm}
		\noalign{\global\arrayrulewidth=0.4mm} \arrayrulecolor{black}\hline \noalign{\global\arrayrulewidth=0.2mm}
	\end{tabular}
\end{table*}

\setlength{\tabcolsep}{7.75pt}
\renewcommand{\arraystretch}{1}
\begin{table*}
	\centering
	\caption{Comparison of coreset selection methods using downstream model validation on decreasing sized splits of {\bf EuroSAT}.
		Full dataset training on the ResNet18 model achieves 98.59\% (EuroSAT80), 98.20\% (EuroSAT40), 97.36\% (EuroSAT20), and 93.64\% (EuroSAT10) accuracy.  
		``Rel. Rand.'' is Mean accuracy across all data selection rates relative to Random. 
		``EuroSAT All'' is Mean accuracy for all EuroSAT splits.
		EuroSAT10 10\% selection has only 270 examples.
		A corresponding results plot is provided in \cref{fig:euro} of the main paper.
	}
	\vspace{-0.875em}
	\label{tab:euro}
	\begin{tabular}{l  c c c c c c c c c c c c c c c}
		\noalign{\global\arrayrulewidth=0.4mm} \hline \noalign{\global\arrayrulewidth=0.2mm} \arrayrulecolor{gray}
		\multicolumn{1}{c}{Selection} & & \multicolumn{3}{c}{}&  & \multicolumn{5}{c}{$\xlabel{\text{ ~Percent of Dataset Selected}}{\text{~/ Number of Examples}}$} &  \rule{0pt}{2ex} \\ 
		\multicolumn{1}{c}{Method} && \multicolumn{3}{c}{Coreset Selection Requirements}& &  70\% & 50\% & 30\% & 20\% & 10\%  & & $\meanrelrand{\text{Mean}}{\text{Rel. Rand.}}$   \rule{0pt}{0ex} \\ 
		\noalign{\global\arrayrulewidth=0.4mm} \arrayrulecolor{white}\hline \noalign{\global\arrayrulewidth=0.2mm} \arrayrulecolor{gray}
		\cline{1-1} \cline{3-5} \cline{7-11} \cline{13-13}
		\multicolumn{6}{l}{\textcolor{gray}{\bf EuroSAT All}} \rule{0pt}{3ex}\\ 
		$\methodpub{\text{ZCore}}{}$ && \multicolumn{3}{c}{$\methodpub{\text{Unlabeled Data}}{}$} && $\methodpub{\text{96.53}}{	}$ & $\methodpub{\text{95.74}}{	}$ & $\methodpub{\text{93.21}}{	}$ & $\methodpub{\text{91.74}}{	}$ & $\methodpub{\text{83.27}}{	}$ & & $\meanrelrand{\text{\bf 92.10}}{\text{\bf +1.14	}}$ \stddevrow \\
		$\methodpub{\text{Random}}{}$ && \multicolumn{3}{c}{$\methodpub{\text{Unlabeled Data}}{}$} && $\methodpub{\text{94.56}}{	}$ & $\methodpub{\text{92.91}}{	}$ & $\methodpub{\text{89.80}}{	}$ & $\methodpub{\text{87.88}}{	}$ & $\methodpub{\text{83.61}}{	}$ & & $\meanrelrand{\text{90.96	}}{\text{+0.00	}}$ \stddevrow \\
		$\methodpub{\text{TDDS}}{\text{CVPR 2024}}$ && \multicolumn{3}{c}{$\methodpub{\text{Full Training on Labeled Data}}{}$}  && $\methodpub{\text{96.93}}{	}$ & $\methodpub{\text{96.35}}{	}$ & $\methodpub{\text{94.55}}{	}$ & $\methodpub{\text{93.56}}{	}$ & $\methodpub{\text{87.80}}{	}$ & & $\meanrelrand{\text{\bf93.96	}}{\text{\bf +	3.01	}}$ \stddevrow \\
		\noalign{\global\arrayrulewidth=0.4mm} \arrayrulecolor{white}\hline \noalign{\global\arrayrulewidth=0.2mm} \arrayrulecolor{gray}
		\hline
		\multicolumn{6}{l}{\textcolor{gray}{\bf EuroSAT80 $ \scriptstyle (21,600)$}}  & $\color{gray} \scriptstyle 15,120$ & $\color{gray} \scriptstyle 10,800$ & $\color{gray} \scriptstyle 6,480$ & $\color{gray} \scriptstyle 4,320$ & $\color{gray} \scriptstyle 2,160$ \rule{0pt}{3ex} \\ 
		$\methodpub{\text{ZCore}}{}$  && \multicolumn{3}{c}{$\methodpub{\text{Unlabeled Data}}{}$} &&  $\stddev{\text{98.32}}{\text{0.08}}$ & $\stddev{\text{98.15}}{\text{0.13}}$ & $\stddev{\text{97.72}}{\text{0.13}}$ & $\stddev{\text{97.31}}{\text{0.17}}$ & $\stddev{\text{95.80}}{\text{0.18}}$ & & $\meanrelrand{\text{97.46}}{\text{\bf +0.56}}$ \stddevrow \\
		$\methodpub{\text{Random}}{}$  && \multicolumn{3}{c}{$\methodpub{\text{Unlabeled Data}}{}$} &&  $\stddev{\text{98.20}}{\text{0.11}}$ & $\stddev{\text{97.94}}{\text{0.10}}$ & $\stddev{\text{96.98}}{\text{0.17}}$ & $\stddev{\text{96.65}}{\text{0.29}}$ & $\stddev{\text{94.72}}{\text{0.49}}$ & & $\meanrelrand{\text{96.90}}{\text{+0.00}}$ \stddevrow \\
		$\methodpub{\text{TDDS}}{\text{CVPR 2024}}$&& \multicolumn{3}{c}{$\methodpub{\text{Full Training on Labeled Data}}{}$}  && $\stddev{\text{98.62}}{\text{0.05}}$ & $\stddev{\text{98.58}}{\text{0.11}}$ & $\stddev{\text{98.43}}{\text{0.03}}$ & $\stddev{\text{98.09}}{\text{0.10}}$ & $\stddev{\text{96.28}}{\text{0.11}}$ & & $\meanrelrand{\text{98.00}}{\text{\bf +1.10}}$ \stddevrow \\
		\noalign{\global\arrayrulewidth=0.4mm} \arrayrulecolor{white}\hline \noalign{\global\arrayrulewidth=0.2mm} \arrayrulecolor{gray}
		\hline
		\multicolumn{6}{l}{\textcolor{gray}{\bf EuroSAT40 $ \scriptstyle (10,800)$}}  & $\color{gray} \scriptstyle 7,560$ & $\color{gray} \scriptstyle 5,400$ & $\color{gray} \scriptstyle 3,240$ & $\color{gray} \scriptstyle 2,160$ & $\color{gray} \scriptstyle 1,080$ \rule{0pt}{3ex} \\ 
		$\methodpub{\text{ZCore}}{}$  && \multicolumn{3}{c}{$\methodpub{\text{Unlabeled Data}}{}$} && $\stddev{\text{97.59}}{\text{0.05}}$ & $\stddev{\text{97.53}}{\text{0.16}}$ & $\stddev{\text{96.45}}{\text{0.12}}$ & $\stddev{\text{96.06}}{\text{0.19}}$ & $\stddev{\text{92.94}}{\text{0.55}}$ & & $\meanrelrand{\text{96.11}}{\text{\bf +1.54}}$ \stddevrow \\
		$\methodpub{\text{Random}}{}$  && \multicolumn{3}{c}{$\methodpub{\text{Unlabeled Data}}{}$} &&  $\stddev{\text{97.04}}{\text{0.07}}$ & $\stddev{\text{96.43}}{\text{0.37}}$ & $\stddev{\text{95.00}}{\text{0.67}}$ & $\stddev{\text{93.73}}{\text{0.58}}$ & $\stddev{\text{90.69}}{\text{0.53}}$ & & $\meanrelrand{\text{94.58}}{\text{+0.00}}$ \stddevrow \\
		$\methodpub{\text{TDDS}}{\text{CVPR 2024}}$ && \multicolumn{3}{c}{$\methodpub{\text{Full Training on Labeled Data}}{}$}  &&  $\stddev{\text{97.97}}{\text{0.09}}$ & $\stddev{\text{98.06}}{\text{0.06}}$ & $\stddev{\text{97.55}}{\text{0.08}}$ & $\stddev{\text{96.79}}{\text{0.16}}$ & $\stddev{\text{92.78}}{\text{0.27}}$ & & $\meanrelrand{\text{96.63}}{\text{\bf +2.05}}$ \stddevrow \\
		\noalign{\global\arrayrulewidth=0.4mm} \arrayrulecolor{white}\hline \noalign{\global\arrayrulewidth=0.2mm} \arrayrulecolor{gray}
		\hline
		\multicolumn{6}{l}{\textcolor{gray}{\bf EuroSAT20 $\scriptstyle (5,400)$}}  & $\color{gray} \scriptstyle 3,780$ & $\color{gray} \scriptstyle 2,700$ & $\color{gray} \scriptstyle 1,620$ & $\color{gray} \scriptstyle 1,080$ & $\color{gray} \scriptstyle 540$ \rule{0pt}{3ex} \\ 
		$\methodpub{\text{ZCore}}{}$  && \multicolumn{3}{c}{$\methodpub{\text{Unlabeled Data}}{}$} &&  $\stddev{\text{96.49}}{\text{0.16}}$ & $\stddev{\text{95.45}}{\text{0.22}}$ & $\stddev{\text{92.60}}{\text{0.29}}$ & $\stddev{\text{91.80}}{\text{0.70}}$ & $\stddev{\text{80.39}}{\text{3.91}}$ & & $\meanrelrand{\text{91.35}}{\text{\bf +1.53}}$ \stddevrow \\
		$\methodpub{\text{Random}}{}$  && \multicolumn{3}{c}{$\methodpub{\text{Unlabeled Data}}{}$} && $\stddev{\text{95.14}}{\text{0.32}}$ & $\stddev{\text{93.25}}{\text{0.59}}$ & $\stddev{\text{89.36}}{\text{0.54}}$ & $\stddev{\text{88.01}}{\text{0.22}}$ & $\stddev{\text{83.30}}{\text{0.73}}$ & & $\meanrelrand{\text{89.81}}{\text{+0.00}}$ \stddevrow \\
		$\methodpub{\text{TDDS}}{\text{CVPR 2024}}$ && \multicolumn{3}{c}{$\methodpub{\text{Full Training on Labeled Data}}{}$}  && $\stddev{\text{96.94}}{\text{0.10}}$ & $\stddev{\text{96.45}}{\text{0.07}}$ & $\stddev{\text{93.86}}{\text{0.56}}$ & $\stddev{\text{94.70}}{\text{0.35}}$ & $\stddev{\text{86.94}}{\text{0.55}}$ & & $\meanrelrand{\text{93.78}}{\text{\bf +3.97}}$ \stddevrow \\
		\noalign{\global\arrayrulewidth=0.4mm} \arrayrulecolor{white}\hline \noalign{\global\arrayrulewidth=0.2mm} \arrayrulecolor{gray}
		\hline
		\multicolumn{6}{l}{\textcolor{gray}{\bf EuroSAT10 $\scriptstyle (2,700)$}}  & $\color{gray} \scriptstyle 1,890$ & $\color{gray} \scriptstyle 1,350$ & $\color{gray} \scriptstyle 810$ & $\color{gray} \scriptstyle 540$ & $\color{gray} \scriptstyle 270$ \rule{0pt}{3ex} \\ 
		$\methodpub{\text{ZCore}}{}$  && \multicolumn{3}{c}{$\methodpub{\text{Unlabeled Data}}{}$} &&  $\stddev{\text{93.71}}{\text{0.23}}$ & $\stddev{\text{91.82}}{\text{0.23}}$ & $\stddev{\text{86.08}}{\text{1.16}}$ & $\stddev{\text{81.77}}{\text{2.68}}$ & $\stddev{\text{63.96}}{\text{2.76}}$ & & $\meanrelrand{\text{83.47}}{\text{\bf +0.92}}$ \stddevrow \\
		$\methodpub{\text{Random}}{}$  && \multicolumn{3}{c}{$\methodpub{\text{Unlabeled Data}}{}$} &&  $\stddev{\text{90.35}}{\text{0.64}}$ & $\stddev{\text{87.55}}{\text{0.67}}$ & $\stddev{\text{83.06}}{\text{1.61}}$ & $\stddev{\text{78.97}}{\text{1.88}}$ & $\stddev{\text{72.81}}{\text{2.25}}$ & & $\meanrelrand{\text{82.55	}}{\text{+0.00}}$ \stddevrow \\
		$\methodpub{\text{TDDS}}{\text{CVPR 2024}}$ && \multicolumn{3}{c}{$\methodpub{\text{Full Training on Labeled Data}}{}$}  &&  $\stddev{\text{94.62}}{\text{0.09}}$ & $\stddev{\text{92.92}}{\text{0.33}}$ & $\stddev{\text{89.41}}{\text{0.52}}$ & $\stddev{\text{85.56}}{\text{0.67}}$ & $\stddev{\text{74.74}}{\text{2.02}}$ & & $\meanrelrand{\text{87.45}}{\text{\bf +4.90}}$ \stddevrow \\
		\noalign{\global\arrayrulewidth=0.4mm} \arrayrulecolor{white}\hline \noalign{\global\arrayrulewidth=0.2mm}
		\noalign{\global\arrayrulewidth=0.4mm} \arrayrulecolor{black}\hline \noalign{\global\arrayrulewidth=0.2mm}
	\end{tabular}
\end{table*}

\end{document}

%% file: math_commands.tex
\usepackage{amsmath,amsfonts,bm}

\def\eqref#1{equation~\ref{#1}}

\def\1{\bm{1}}

\def\rs{{\textnormal{s}}}

\def\rx{{\textnormal{x}}}

\def\rz{{\textnormal{z}}}

\def\rvs{{\mathbf{s}}}

\def\rvx{{\mathbf{x}}}

\def\rvz{{\mathbf{z}}}

\def\ervd{{\textnormal{d}}}

\def\vs{{\bm{s}}}

\def\vv{{\bm{v}}}
\def\vw{{\bm{w}}}

\def\vz{{\bm{z}}}

\def\evs{{s}}

\def\evv{{v}}
\def\evw{{w}}

\def\evz{{z}}

\def\mD{{\bm{D}}}

\def\mZ{{\bm{Z}}}

\DeclareMathAlphabet{\mathsfit}{\encodingdefault}{\sfdefault}{m}{sl}
\SetMathAlphabet{\mathsfit}{bold}{\encodingdefault}{\sfdefault}{bx}{n}

\def\sK{{\mathbb{K}}}

\def\sN{{\mathbb{N}}}

\def\sR{{\mathbb{R}}}
\def\sS{{\mathbb{S}}}

\newcommand{\E}{\mathbb{E}}

\newcommand{\R}{\mathbb{R}}

\DeclareMathOperator*{\argmin}{arg\,min}

%% file: resnet18_0_paper.tex
\begin{tikzpicture}

\definecolor{chocolate2267451}{RGB}{50,50,50}
\definecolor{dimgray85}{RGB}{85,85,85}
\definecolor{gainsboro229}{RGB}{239,239,239}
\definecolor{gainsboro247}{RGB}{253,253,253}
\definecolor{darkorange2551094}{RGB}{255,109,4}
\definecolor{gray119}{RGB}{255,109,4}
\definecolor{mediumpurple152142213}{RGB}{142,186,66}
\definecolor{yellowgreen14218666}{RGB}{142,186,66}
\definecolor{steelblue52138189}{RGB}{52,138,189}
\definecolor{whitesmoke238}{RGB}{238,238,238}

\definecolor{lightgray204}{RGB}{204,204,204}

\begin{axis}[
width=5.1cm,
height=4cm,
ylabel=\textcolor{dimgray85} {\footnotesize Frequency},
ymajorgrids,
y grid style={gainsboro229},
legend style={
	fill opacity=0,
	draw opacity=0,
  text opacity=1,
  at={(1.04,1.01)},
  anchor=north east,
  draw=lightgray204,
  fill=gainsboro247,
},
legend cell align={left},
axis background/.style={fill=gainsboro247},
scaled y ticks=false,
ylabel near ticks,
ytick style={color=white},
xtick pos=left,
ytick pos=left,
axis line style={white},
tick align=outside,
x grid style={white},
xmajorgrids,
xmin=-2.35, xmax=6.75,
xtick style={color=dimgray85},
ymajorgrids,
ymin=0, ymax=23000,
ticklabel style={font=\tiny},
tick align=inside,
axis line style={gainsboro229},
ytick={5000,10000,15000,20000},
yticklabels={5\textrm{K}, 10\textrm{K}, 15\textrm{K}, 20\textrm{K}},
y label style={at={(axis description cs:-0.1, 0.5)}},
yticklabel style = {xshift=0.5ex},
xticklabel style = {yshift=0.5ex},
]
\draw[draw=whitesmoke238,fill=chocolate2267451,fill opacity=0.5,very thin] (axis cs:0,0) rectangle (axis cs:0.263441753387451,22261);
\addlegendimage{ybar,ybar legend,draw=whitesmoke238,fill=chocolate2267451,fill opacity=0.5,very thin}
\addlegendentry{\tiny ResNet18}

\draw[draw=whitesmoke238,fill=chocolate2267451,fill opacity=0.5,very thin] (axis cs:0.263441753387451,0) rectangle (axis cs:0.526883506774902,9403);
\draw[draw=whitesmoke238,fill=chocolate2267451,fill opacity=0.5,very thin] (axis cs:0.526883506774902,0) rectangle (axis cs:0.790325260162353,5903);
\draw[draw=whitesmoke238,fill=chocolate2267451,fill opacity=0.5,very thin] (axis cs:0.790325260162354,0) rectangle (axis cs:1.0537670135498,4035);
\draw[draw=whitesmoke238,fill=chocolate2267451,fill opacity=0.5,very thin] (axis cs:1.0537670135498,0) rectangle (axis cs:1.31720876693726,2629);
\draw[draw=whitesmoke238,fill=chocolate2267451,fill opacity=0.5,very thin] (axis cs:1.31720876693726,0) rectangle (axis cs:1.58065052032471,1744);
\draw[draw=whitesmoke238,fill=chocolate2267451,fill opacity=0.5,very thin] (axis cs:1.58065052032471,0) rectangle (axis cs:1.84409227371216,1300);
\draw[draw=whitesmoke238,fill=chocolate2267451,fill opacity=0.5,very thin] (axis cs:1.84409227371216,0) rectangle (axis cs:2.10753402709961,859);
\draw[draw=whitesmoke238,fill=chocolate2267451,fill opacity=0.5,very thin] (axis cs:2.10753402709961,0) rectangle (axis cs:2.37097578048706,602);
\draw[draw=whitesmoke238,fill=chocolate2267451,fill opacity=0.5,very thin] (axis cs:2.37097578048706,0) rectangle (axis cs:2.63441753387451,412);
\draw[draw=whitesmoke238,fill=chocolate2267451,fill opacity=0.5,very thin] (axis cs:2.63441753387451,0) rectangle (axis cs:2.89785928726196,306);
\draw[draw=whitesmoke238,fill=chocolate2267451,fill opacity=0.5,very thin] (axis cs:2.89785928726196,0) rectangle (axis cs:3.16130104064941,200);
\draw[draw=whitesmoke238,fill=chocolate2267451,fill opacity=0.5,very thin] (axis cs:3.16130104064941,0) rectangle (axis cs:3.42474279403687,136);
\draw[draw=whitesmoke238,fill=chocolate2267451,fill opacity=0.5,very thin] (axis cs:3.42474279403687,0) rectangle (axis cs:3.68818454742432,86);
\draw[draw=whitesmoke238,fill=chocolate2267451,fill opacity=0.5,very thin] (axis cs:3.68818454742432,0) rectangle (axis cs:3.95162630081177,47);
\draw[draw=whitesmoke238,fill=chocolate2267451,fill opacity=0.5,very thin] (axis cs:3.95162630081177,0) rectangle (axis cs:4.21506805419922,32);
\draw[draw=whitesmoke238,fill=chocolate2267451,fill opacity=0.5,very thin] (axis cs:4.21506805419922,0) rectangle (axis cs:4.47850980758667,18);
\draw[draw=whitesmoke238,fill=chocolate2267451,fill opacity=0.5,very thin] (axis cs:4.47850980758667,0) rectangle (axis cs:4.74195156097412,8);
\draw[draw=whitesmoke238,fill=chocolate2267451,fill opacity=0.5,very thin] (axis cs:4.74195156097412,0) rectangle (axis cs:5.00539331436157,5);
\draw[draw=whitesmoke238,fill=chocolate2267451,fill opacity=0.5,very thin] (axis cs:5.00539331436157,0) rectangle (axis cs:5.26883506774902,5);
\draw[draw=whitesmoke238,fill=chocolate2267451,fill opacity=0.5,very thin] (axis cs:5.26883506774902,0) rectangle (axis cs:5.53227682113647,2);
\draw[draw=whitesmoke238,fill=chocolate2267451,fill opacity=0.5,very thin] (axis cs:5.53227682113648,0) rectangle (axis cs:5.79571857452393,4);
\draw[draw=whitesmoke238,fill=chocolate2267451,fill opacity=0.5,very thin] (axis cs:5.79571857452393,0) rectangle (axis cs:6.05916032791138,1);
\draw[draw=whitesmoke238,fill=chocolate2267451,fill opacity=0.5,very thin] (axis cs:6.05916032791138,0) rectangle (axis cs:6.32260208129883,1);
\draw[draw=whitesmoke238,fill=chocolate2267451,fill opacity=0.5,very thin] (axis cs:6.32260208129883,0) rectangle (axis cs:6.58604383468628,1);
\draw[draw=whitesmoke238,fill=steelblue52138189,fill opacity=0.25,very thin] (axis cs:0.000391461169069574,0) rectangle (axis cs:0.26381752033833,2047);
\addlegendimage{ybar,ybar legend,draw=whitesmoke238,fill=steelblue52138189,fill opacity=0.5,very thin}
\addlegendentry{\tiny Uniform}

\draw[draw=whitesmoke238,fill=steelblue52138189,fill opacity=0.25,very thin] (axis cs:0.26381752033833,0) rectangle (axis cs:0.527243579507591,1983);
\draw[draw=whitesmoke238,fill=steelblue52138189,fill opacity=0.25,very thin] (axis cs:0.527243579507591,0) rectangle (axis cs:0.790669638676852,1945);
\draw[draw=whitesmoke238,fill=steelblue52138189,fill opacity=0.25,very thin] (axis cs:0.790669638676852,0) rectangle (axis cs:1.05409569784611,1983);
\draw[draw=whitesmoke238,fill=steelblue52138189,fill opacity=0.25,very thin] (axis cs:1.05409569784611,0) rectangle (axis cs:1.31752175701537,2087);
\draw[draw=whitesmoke238,fill=steelblue52138189,fill opacity=0.25,very thin] (axis cs:1.31752175701537,0) rectangle (axis cs:1.58094781618463,1977);
\draw[draw=whitesmoke238,fill=steelblue52138189,fill opacity=0.25,very thin] (axis cs:1.58094781618463,0) rectangle (axis cs:1.84437387535389,1990);
\draw[draw=whitesmoke238,fill=steelblue52138189,fill opacity=0.25,very thin] (axis cs:1.84437387535389,0) rectangle (axis cs:2.10779993452315,2028);
\draw[draw=whitesmoke238,fill=steelblue52138189,fill opacity=0.25,very thin] (axis cs:2.10779993452315,0) rectangle (axis cs:2.37122599369242,2068);
\draw[draw=whitesmoke238,fill=steelblue52138189,fill opacity=0.25,very thin] (axis cs:2.37122599369242,0) rectangle (axis cs:2.63465205286168,2046);
\draw[draw=whitesmoke238,fill=steelblue52138189,fill opacity=0.25,very thin] (axis cs:2.63465205286168,0) rectangle (axis cs:2.89807811203094,2002);
\draw[draw=whitesmoke238,fill=steelblue52138189,fill opacity=0.25,very thin] (axis cs:2.89807811203094,0) rectangle (axis cs:3.1615041712002,1937);
\draw[draw=whitesmoke238,fill=steelblue52138189,fill opacity=0.25,very thin] (axis cs:3.1615041712002,0) rectangle (axis cs:3.42493023036946,1988);
\draw[draw=whitesmoke238,fill=steelblue52138189,fill opacity=0.25,very thin] (axis cs:3.42493023036946,0) rectangle (axis cs:3.68835628953872,2013);
\draw[draw=whitesmoke238,fill=steelblue52138189,fill opacity=0.25,very thin] (axis cs:3.68835628953872,0) rectangle (axis cs:3.95178234870798,1994);
\draw[draw=whitesmoke238,fill=steelblue52138189,fill opacity=0.25,very thin] (axis cs:3.95178234870798,0) rectangle (axis cs:4.21520840787724,2057);
\draw[draw=whitesmoke238,fill=steelblue52138189,fill opacity=0.25,very thin] (axis cs:4.21520840787724,0) rectangle (axis cs:4.4786344670465,1931);
\draw[draw=whitesmoke238,fill=steelblue52138189,fill opacity=0.25,very thin] (axis cs:4.4786344670465,0) rectangle (axis cs:4.74206052621576,2023);
\draw[draw=whitesmoke238,fill=steelblue52138189,fill opacity=0.25,very thin] (axis cs:4.74206052621576,0) rectangle (axis cs:5.00548658538502,2029);
\draw[draw=whitesmoke238,fill=steelblue52138189,fill opacity=0.25,very thin] (axis cs:5.00548658538502,0) rectangle (axis cs:5.26891264455428,1969);
\draw[draw=whitesmoke238,fill=steelblue52138189,fill opacity=0.25,very thin] (axis cs:5.26891264455428,0) rectangle (axis cs:5.53233870372354,2003);
\draw[draw=whitesmoke238,fill=steelblue52138189,fill opacity=0.25,very thin] (axis cs:5.53233870372354,0) rectangle (axis cs:5.7957647628928,2004);
\draw[draw=whitesmoke238,fill=steelblue52138189,fill opacity=0.25,very thin] (axis cs:5.7957647628928,0) rectangle (axis cs:6.05919082206206,1896);
\draw[draw=whitesmoke238,fill=steelblue52138189,fill opacity=0.25,very thin] (axis cs:6.05919082206206,0) rectangle (axis cs:6.32261688123133,2052);
\draw[draw=whitesmoke238,fill=steelblue52138189,fill opacity=0.25,very thin] (axis cs:6.32261688123133,0) rectangle (axis cs:6.58604294040059,1948);
\draw[draw=whitesmoke238,fill=mediumpurple152142213,fill opacity=0.25,very thin] (axis cs:-2.14737014874828,0) rectangle (axis cs:-1.88282208972805,5);
\addlegendimage{ybar,ybar legend,draw=whitesmoke238,fill=mediumpurple152142213,fill opacity=0.5,very thin}
\addlegendentry{\tiny Gaussian}

\draw[draw=whitesmoke238,fill=mediumpurple152142213,fill opacity=0.25,very thin] (axis cs:-1.88282208972805,0) rectangle (axis cs:-1.61827403070782,21);
\draw[draw=whitesmoke238,fill=mediumpurple152142213,fill opacity=0.25,very thin] (axis cs:-1.61827403070782,0) rectangle (axis cs:-1.35372597168758,55);
\draw[draw=whitesmoke238,fill=mediumpurple152142213,fill opacity=0.25,very thin] (axis cs:-1.35372597168758,0) rectangle (axis cs:-1.08917791266735,187);
\draw[draw=whitesmoke238,fill=mediumpurple152142213,fill opacity=0.25,very thin] (axis cs:-1.08917791266735,0) rectangle (axis cs:-0.824629853647118,540);
\draw[draw=whitesmoke238,fill=mediumpurple152142213,fill opacity=0.25,very thin] (axis cs:-0.824629853647118,0) rectangle (axis cs:-0.560081794626885,1342);
\draw[draw=whitesmoke238,fill=mediumpurple152142213,fill opacity=0.25,very thin] (axis cs:-0.560081794626885,0) rectangle (axis cs:-0.295533735606652,2676);
\draw[draw=whitesmoke238,fill=mediumpurple152142213,fill opacity=0.25,very thin] (axis cs:-0.295533735606652,0) rectangle (axis cs:-0.0309856765864187,4355);
\draw[draw=whitesmoke238,fill=mediumpurple152142213,fill opacity=0.25,very thin] (axis cs:-0.0309856765864187,0) rectangle (axis cs:0.233562382433814,6319);
\draw[draw=whitesmoke238,fill=mediumpurple152142213,fill opacity=0.25,very thin] (axis cs:0.233562382433814,0) rectangle (axis cs:0.498110441454047,7666);
\draw[draw=whitesmoke238,fill=mediumpurple152142213,fill opacity=0.25,very thin] (axis cs:0.498110441454047,0) rectangle (axis cs:0.76265850047428,8069);
\draw[draw=whitesmoke238,fill=mediumpurple152142213,fill opacity=0.25,very thin] (axis cs:0.76265850047428,0) rectangle (axis cs:1.02720655949451,6992);
\draw[draw=whitesmoke238,fill=mediumpurple152142213,fill opacity=0.25,very thin] (axis cs:1.02720655949451,0) rectangle (axis cs:1.29175461851475,5287);
\draw[draw=whitesmoke238,fill=mediumpurple152142213,fill opacity=0.25,very thin] (axis cs:1.29175461851475,0) rectangle (axis cs:1.55630267753498,3368);
\draw[draw=whitesmoke238,fill=mediumpurple152142213,fill opacity=0.25,very thin] (axis cs:1.55630267753498,0) rectangle (axis cs:1.82085073655521,1829);
\draw[draw=whitesmoke238,fill=mediumpurple152142213,fill opacity=0.25,very thin] (axis cs:1.82085073655521,0) rectangle (axis cs:2.08539879557545,824);
\draw[draw=whitesmoke238,fill=mediumpurple152142213,fill opacity=0.25,very thin] (axis cs:2.08539879557545,0) rectangle (axis cs:2.34994685459568,319);
\draw[draw=whitesmoke238,fill=mediumpurple152142213,fill opacity=0.25,very thin] (axis cs:2.34994685459568,0) rectangle (axis cs:2.61449491361591,97);
\draw[draw=whitesmoke238,fill=mediumpurple152142213,fill opacity=0.25,very thin] (axis cs:2.61449491361591,0) rectangle (axis cs:2.87904297263614,38);
\draw[draw=whitesmoke238,fill=mediumpurple152142213,fill opacity=0.25,very thin] (axis cs:2.87904297263614,0) rectangle (axis cs:3.14359103165638,7);
\draw[draw=whitesmoke238,fill=mediumpurple152142213,fill opacity=0.25,very thin] (axis cs:3.14359103165638,0) rectangle (axis cs:3.40813909067661,4);
\draw[draw=whitesmoke238,fill=gray119,fill opacity=0.5,very thin] (axis cs:0.00305815222802983,0) rectangle (axis cs:0.266005934820616,1608);
\addlegendimage{ybar,ybar legend,draw=whitesmoke238,fill=gray119,fill opacity=0.5,very thin}
\addlegendentry{\tiny Triangular}

\draw[draw=whitesmoke238,fill=gray119,fill opacity=0.5,very thin] (axis cs:0.266005934820616,0) rectangle (axis cs:0.528953717413201,3919);
\draw[draw=whitesmoke238,fill=gray119,fill opacity=0.5,very thin] (axis cs:0.528953717413201,0) rectangle (axis cs:0.791901500005787,3761);
\draw[draw=whitesmoke238,fill=gray119,fill opacity=0.5,very thin] (axis cs:0.791901500005787,0) rectangle (axis cs:1.05484928259837,3669);
\draw[draw=whitesmoke238,fill=gray119,fill opacity=0.5,very thin] (axis cs:1.05484928259837,0) rectangle (axis cs:1.31779706519096,3469);
\draw[draw=whitesmoke238,fill=gray119,fill opacity=0.5,very thin] (axis cs:1.31779706519096,0) rectangle (axis cs:1.58074484778354,3310);
\draw[draw=whitesmoke238,fill=gray119,fill opacity=0.5,very thin] (axis cs:1.58074484778354,0) rectangle (axis cs:1.84369263037613,3097);
\draw[draw=whitesmoke238,fill=gray119,fill opacity=0.5,very thin] (axis cs:1.84369263037613,0) rectangle (axis cs:2.10664041296872,2907);
\draw[draw=whitesmoke238,fill=gray119,fill opacity=0.5,very thin] (axis cs:2.10664041296872,0) rectangle (axis cs:2.3695881955613,2799);
\draw[draw=whitesmoke238,fill=gray119,fill opacity=0.5,very thin] (axis cs:2.3695881955613,0) rectangle (axis cs:2.63253597815389,2565);
\draw[draw=whitesmoke238,fill=gray119,fill opacity=0.5,very thin] (axis cs:2.63253597815389,0) rectangle (axis cs:2.89548376074647,2463);
\draw[draw=whitesmoke238,fill=gray119,fill opacity=0.5,very thin] (axis cs:2.89548376074647,0) rectangle (axis cs:3.15843154333906,2253);
\draw[draw=whitesmoke238,fill=gray119,fill opacity=0.5,very thin] (axis cs:3.15843154333906,0) rectangle (axis cs:3.42137932593164,2118);
\draw[draw=whitesmoke238,fill=gray119,fill opacity=0.5,very thin] (axis cs:3.42137932593164,0) rectangle (axis cs:3.68432710852423,1923);
\draw[draw=whitesmoke238,fill=gray119,fill opacity=0.5,very thin] (axis cs:3.68432710852423,0) rectangle (axis cs:3.94727489111682,1725);
\draw[draw=whitesmoke238,fill=gray119,fill opacity=0.5,very thin] (axis cs:3.94727489111682,0) rectangle (axis cs:4.2102226737094,1624);
\draw[draw=whitesmoke238,fill=gray119,fill opacity=0.5,very thin] (axis cs:4.2102226737094,0) rectangle (axis cs:4.47317045630199,1389);
\draw[draw=whitesmoke238,fill=gray119,fill opacity=0.5,very thin] (axis cs:4.47317045630199,0) rectangle (axis cs:4.73611823889457,1264);
\draw[draw=whitesmoke238,fill=gray119,fill opacity=0.5,very thin] (axis cs:4.73611823889457,0) rectangle (axis cs:4.99906602148716,1038);
\draw[draw=whitesmoke238,fill=gray119,fill opacity=0.5,very thin] (axis cs:4.99906602148716,0) rectangle (axis cs:5.26201380407974,926);
\draw[draw=whitesmoke238,fill=gray119,fill opacity=0.5,very thin] (axis cs:5.26201380407974,0) rectangle (axis cs:5.52496158667233,789);
\draw[draw=whitesmoke238,fill=gray119,fill opacity=0.5,very thin] (axis cs:5.52496158667233,0) rectangle (axis cs:5.78790936926492,613);
\draw[draw=whitesmoke238,fill=gray119,fill opacity=0.5,very thin] (axis cs:5.78790936926491,0) rectangle (axis cs:6.0508571518575,441);
\draw[draw=whitesmoke238,fill=gray119,fill opacity=0.5,very thin] (axis cs:6.0508571518575,0) rectangle (axis cs:6.31380493445009,241);
\draw[draw=whitesmoke238,fill=gray119,fill opacity=0.5,very thin] (axis cs:6.31380493445009,0) rectangle (axis cs:6.57675271704267,89);
\addplot [thick, opacity=0.75,  chocolate2267451, forget plot]
table {%
0 16863.5383193701
0.0330956976617401 19772.7344904388
0.0661913953234802 20719.5329738883
0.0992870929852203 20001.5175058999
0.13238279064696 18391.0497752776
0.1654784883087 16588.7633445746
0.198574185970441 14962.3389122879
0.231669883632181 13603.2466127928
0.264765581293921 12482.6475052407
0.297861278955661 11549.5591541643
0.330956976617401 10757.6583037846
0.364052674279141 10065.5028416393
0.397148371940881 9439.11304196234
0.430244069602621 8857.42312764919
0.463339767264361 8312.00150416733
0.496435464926101 7800.48105993453
0.529531162587841 7321.36270591385
0.562626860249582 6875.18020605309
0.595722557911322 6467.73472174316
0.628818255573062 6107.70593949627
0.661913953234802 5798.54624240604
0.695009650896542 5533.23284567173
0.728105348558282 5297.75783355321
0.761201046220022 5078.72822664284
0.794296743881762 4866.90077409675
0.827392441543502 4655.67761470492
0.860488139205242 4440.03689924034
0.893583836866983 4218.71649022317
0.926679534528723 3996.16506432717
0.959775232190463 3780.09810597951
0.992870929852203 3576.12318325056
1.02596662751394 3385.1308090258
1.05906232517568 3205.92232997749
1.09215802283742 3038.93423774491
1.12525372049916 2886.16859531843
1.1583494181609 2748.03127204142
1.19144511582264 2621.15891004711
1.22454081348438 2499.22272396863
1.25763651114612 2375.98488399695
1.29073220880786 2249.07679719066
1.3238279064696 2122.39207636274
1.35692360413134 2004.75417134812
1.39001930179308 1904.47819268703
1.42311499945482 1823.22006316517
1.45621069711656 1754.60016434692
1.4893063947783 1689.71536008353
1.52240209244004 1624.36271771862
1.55549779010178 1560.18068364053
1.58859348776352 1499.16682613007
1.62168918542526 1439.43664146851
1.654784883087 1377.6520566475
1.68788058074874 1313.9461111162
1.72097627841048 1252.27944339008
1.75407197607222 1195.86352524171
1.78716767373397 1143.53016808875
1.82026337139571 1091.41083323172
1.85335906905745 1037.87574379974
1.88645476671919 985.898915531568
1.91955046438093 939.915875252778
1.95264616204267 900.580289208266
1.98574185970441 863.472791461635
2.01883755736615 823.464417581347
2.05193325502789 779.828015019916
2.08502895268963 736.377960294491
2.11812465035137 697.108725941482
2.15122034801311 662.865128514317
2.18431604567485 632.160212641149
2.21741174333659 603.778550994853
2.25050744099833 577.776830580036
2.28360313866007 554.546990168128
2.31669883632181 533.638608911984
2.34979453398355 513.578965171814
2.38289023164529 492.74537431413
2.41598592930703 470.52594989551
2.44908162696877 447.856355791693
2.48217732463051 426.761687112736
2.51527302229225 409.199283052792
2.54836871995399 395.893424598288
2.58146441761573 385.89888501065
2.61456011527747 377.052672471134
2.64765581293921 366.740763790949
2.68075151060095 352.64219080304
2.71384720826269 333.777286147111
2.74694290592443 311.612592656678
2.78003860358617 289.828944264485
2.81313430124791 272.077276373089
2.84622999890965 259.434241254441
2.87932569657139 250.03838449716
2.91242139423313 241.07071731566
2.94551709189487 230.86735905359
2.97861278955661 219.52651962615
3.01170848721835 208.300217731789
3.04480418488009 198.556732180683
3.07789988254183 190.771207091969
3.11099558020357 184.176001653252
3.14409127786531 177.45744669052
3.17718697552705 169.773664022271
3.21028267318879 160.974858632224
3.24337837085053 151.025624268226
3.27647406851227 139.741849774579
3.30956976617401 127.421308420589
3.34266546383575 115.47910888822
3.37576116149749 105.978490149869
3.40885685915923 100.273874768369
3.44195255682097 97.9412762781614
3.47504825448271 96.8564935435429
3.50814395214445 94.4768663855247
3.54123964980619 89.5159226581589
3.57433534746793 82.6198435251336
3.60743104512967 75.4047165741471
3.64052674279141 68.9558505499756
3.67362244045315 63.3569590383214
3.70671813811489 58.3265937979475
3.73981383577663 53.8336590363574
3.77290953343837 50.1113389361867
3.80600523110011 47.4011091698941
3.83910092876185 45.7107250483166
3.87219662642359 44.6052088195472
3.90529232408533 43.2595026504011
3.93838802174707 40.9807891529548
3.97148371940881 37.8241802279239
4.00457941707055 34.6430215276599
4.03767511473229 32.4303601105314
4.07077081239403 31.5035946256536
4.10386651005577 31.2423727609638
4.13696220771751 30.6113174946666
4.17005790537925 29.005033491461
4.20315360304099 26.6722638300257
4.23624930070273 24.3812427252713
4.26934499836447 22.6784369031234
4.30244069602621 21.4333926243694
4.33553639368795 20.0516030614049
4.36863209134969 18.0675696910457
4.40172778901143 15.5130420240318
4.43482348667317 12.819043294493
4.46791918433491 10.5180915341233
4.50101488199665 9.03623320129887
4.53411057965839 8.56594088068688
4.56720627732013 8.94013795025702
4.60030197498187 9.61284350989138
4.63339767264361 9.88463848627181
4.66649337030535 9.28822779235241
4.69958906796709 7.86826690493516
4.73268476562883 6.15059224098693
4.76578046329057 4.82151612056396
4.79887616095231 4.34362533689228
4.83197185861405 4.75590280416615
4.86506755627579 5.73131795215832
4.89816325393753 6.79009852799275
4.93125895159927 7.53528667076807
4.96435464926101 7.80401449452618
4.99745034692275 7.65272310844353
5.03054604458449 7.20450475973064
5.06364174224623 6.52616025489012
5.09673743990797 5.65527066177323
5.12983313756971 4.69165787271821
5.16292883523145 3.79850498612521
5.19602453289319 3.11332646422392
5.22912023055494 2.68895668654017
5.26221592821667 2.50603359485812
5.29531162587841 2.49651299465482
5.32840732354016 2.55326812270635
5.3615030212019 2.56293388249292
5.39459871886364 2.47422660733107
5.42769441652538 2.36258716699046
5.46079011418712 2.4271961462151
5.49388581184886 2.86721203037377
5.5269815095106 3.68167783561157
5.56007720717234 4.57108530030129
5.59317290483408 5.08506646941831
5.62626860249582 4.92153729069378
5.65936430015756 4.11488223621936
5.6924599978193 2.9711523608932
5.72555569548104 1.85219732833398
5.75865139314278 0.99711658092773
5.79174709080452 0.475466956096371
5.82484278846626 0.24882301943217
5.857938486128 0.26428773850643
5.89103418378974 0.507729041087031
5.92412988145148 0.981048172754711
5.95722557911322 1.62391436844804
5.99032127677496 2.25746777950619
6.0234169744367 2.63271567000027
6.05651267209844 2.57699168477758
6.08960836976018 2.11683514186092
6.12270406742192 1.45782843251438
6.15579976508366 0.840324915526012
6.1888954627454 0.404560151364992
6.22199116040714 0.162299592615389
6.25508685806888 0.0541969486307598
6.28818255573062 0.015465886721038
6.32127825339236 0.00605569440324546
6.3543739510541 0.0121054390797297
6.38746964871584 0.0411550478598579
6.42056534637758 0.120899708984463
6.45366104403932 0.29250253433939
6.48675674170106 0.581566418424504
6.5198524393628 0.950152412908252
6.55294813702454 1.27558586030591
6.58604383468628 1.40717627589795
};
\addplot [thick, opacity=0.75,  steelblue52138189, forget plot]
table {%
0.000391461169069582 1016.05062462919
0.0334851871953586 1135.9535843218
0.0665789132216476 1252.61030386083
0.0996726392479367 1363.51327712385
0.132766365274226 1466.53340292237
0.165860091300515 1560.03635755262
0.198953817326804 1642.94656343731
0.232047543353093 1714.75636754034
0.265141269379382 1775.48560842675
0.298234995405671 1825.60273994966
0.33132872143196 1865.92218709386
0.364422447458249 1897.49331559223
0.397516173484538 1921.49457479492
0.430609899510827 1939.14275778168
0.463703625537116 1951.62288793966
0.496797351563405 1960.03995360706
0.529891077589694 1965.39032347086
0.562984803615983 1968.5485991295
0.596078529642272 1970.2649514507
0.629172255668561 1971.16840339775
0.66226598169485 1971.77264924598
0.695359707721139 1972.48238486981
0.728453433747428 1973.59938604885
0.761547159773717 1975.32847278659
0.794640885800006 1977.78395603766
0.827734611826295 1980.99722799185
0.860828337852584 1984.92595548247
0.893922063878873 1989.46501675213
0.927015789905162 1994.45900887399
0.960109515931451 1999.71592120948
0.99320324195774 2005.0214411533
1.02629696798403 2010.15331433833
1.05939069401032 2014.89518685921
1.09248442003661 2019.04937962074
1.1255781460629 2022.44806959851
1.15867187208919 2024.96238546225
1.19176559811547 2026.50898596784
1.22485932414176 2027.05380071551
1.25795305016805 2026.61278398742
1.29104677619434 2025.24975154894
1.32414050222063 2023.07160342301
1.35723422824692 2020.22143537462
1.39032795427321 2016.87016353013
1.4234216802995 2013.20730550833
1.45651540632579 2009.43148439495
1.48960913235208 2005.74108732537
1.52270285837836 2002.32537604403
1.55579658440465 1999.35626778229
1.58889031043094 1996.98101081868
1.62198403645723 1995.31605942172
1.65507776248352 1994.44255666705
1.68817148850981 1994.40388690812
1.7212652145361 1995.20569380585
1.75435894056239 1996.81854085902
1.78745266658868 1999.18303887179
1.82054639261497 2002.2168518515
1.85364011864125 2005.82262633646
1.88673384466754 2009.8956767777
1.91982757069383 2014.33027514837
1.95292129672012 2019.02365237755
1.98601502274641 2023.87727517366
2.0191087487727 2028.79551786879
2.05220247479899 2033.68238715701
2.08529620082528 2038.43736954211
2.11838992685157 2042.95168211429
2.15148365287786 2047.10618686076
2.18457737890415 2050.77198943473
2.21767110493043 2053.81432844819
2.25076483095672 2056.09983131696
2.28385855698301 2057.50663461347
2.3169522830093 2057.93631016732
2.35004600903559 2057.32607430076
2.38313973506188 2055.65945996173
2.41623346108817 2052.97357028263
2.44932718711446 2049.36126180812
2.48242091314075 2044.96714701872
2.51551463916704 2039.97712708674
2.54860836519332 2034.6021716277
2.58170209121961 2029.05809691561
2.6147958172459 2023.5439635008
2.64788954327219 2018.22222385919
2.68098326929848 2013.20375198017
2.71407699532477 2008.54031945367
2.74717072135106 2004.22600073087
2.78026444737735 2000.2075620337
2.81335817340364 1996.40236748725
2.84645189942993 1992.72100956333
2.87954562545621 1989.09099698294
2.9126393514825 1985.47758379379
2.94573307750879 1981.89824711688
2.97882680353508 1978.42833560228
3.01192052956137 1975.19682463928
3.04501425558766 1972.37267283483
3.07810798161395 1970.14371360245
3.11120170764024 1968.69111436091
3.14429543366653 1968.16304911441
3.17738915969282 1968.65130364419
3.21048288571911 1970.17409549032
3.24357661174539 1972.667536448
3.27667033777168 1975.98702565622
3.30976406379797 1979.9185857148
3.34285778982426 1984.19889296646
3.37595151585055 1988.54164760971
3.40904524187684 1992.66710496713
3.44213896790313 1996.33114682513
3.47523269392942 1999.35027654802
3.50832641995571 2001.61939078787
3.54142014598199 2003.12007276975
3.57451387200828 2003.91836441987
3.60760759803457 2004.15235015535
3.64070132406086 2004.01123267847
3.67379505008715 2003.7087039386
3.70688877611344 2003.45414378204
3.73998250213973 2003.42540456719
3.77307622816602 2003.74663205502
3.80616995419231 2004.47378798504
3.8392636802186 2005.58941181114
3.87235740624489 2007.00687386266
3.90545113227117 2008.5831316001
3.93854485829746 2010.13798428964
3.97163858432375 2011.47715461584
4.00473231035004 2012.41626152915
4.03782603637633 2012.80286806677
4.07091976240262 2012.53421736665
4.10401348842891 2011.56890887355
4.1371072144552 2009.93151499176
4.17020094048149 2007.70991620739
4.20329466650778 2005.04588427813
4.23638839253406 2002.12012780528
4.26948211856035 1999.13359102025
4.30257584458664 1996.28721087486
4.33566957061293 1993.76252310867
4.36876329663922 1991.70540198155
4.40185702266551 1990.21478824714
4.4349507486918 1989.33753050795
4.46804447471809 1989.06953286716
4.50113820074438 1989.36242726313
4.53423192677067 1990.13416475456
4.56732565279695 1991.2814229639
4.60041937882324 1992.69166715169
4.63351310484953 1994.25308817535
4.66660683087582 1995.86136995042
4.69970055690211 1997.42312384336
4.7327942829284 1998.85664331761
4.76588800895469 2000.09117982511
4.79898173498098 2001.06610023839
4.83207546100727 2001.73104646751
4.86516918703356 2002.04767615518
4.89826291305985 2001.99289064703
4.93135663908613 2001.56284538907
4.96445036511242 2000.77664714845
4.99754409113871 1999.67855583509
5.030637817165 1998.33772061045
5.06373154319129 1996.84490885779
5.09682526921758 1995.30620947804
5.12991899524387 1993.83418610378
5.16301272127016 1992.53733278665
5.19610644729645 1991.50890664433
5.22920017332274 1990.81628721605
5.26229389934902 1990.49197592415
5.29538762537531 1990.52723848735
5.3284813514016 1990.86922971919
5.36157507742789 1991.42222339732
5.39466880345418 1992.05328501835
5.42776252948047 1992.60235884562
5.46085625550676 1992.89629662716
5.49394998153305 1992.7658642453
5.52704370755934 1992.06428068029
5.56013743358563 1990.68544322695
5.59323115961191 1988.5797471072
5.6263248856382 1985.76537581182
5.65941861166449 1982.33315486944
5.69251233769078 1978.4435288842
5.72560606371707 1974.31490852776
5.75869978974336 1970.20347714696
5.79179351576965 1966.37545401462
5.82488724179594 1963.07367154592
5.85798096782223 1960.48101977884
5.89107469384852 1958.68373972722
5.92416841987481 1957.63764135415
5.95726214590109 1957.14007302921
5.99035587192738 1956.80993746032
6.02344959795367 1956.07735785241
6.05654332397996 1954.18390932704
6.08963705000625 1950.1938016135
6.12273077603254 1943.0161308008
6.15582450205883 1931.43831065561
6.18891822808512 1914.1709251229
6.22201195411141 1889.90428298466
6.25510568013769 1857.37662206648
6.28819940616398 1815.45296170134
6.32129313219027 1763.21193734858
6.35438685821656 1700.0356933736
6.38748058424285 1625.69543741451
6.42057431026914 1540.42316276445
6.45366803629543 1444.95900433363
6.48676176232172 1340.56430208074
6.51985548834801 1228.99301981506
6.5529492143743 1112.41860334564
6.58604294040059 993.31908978066
};
\addplot [thick, opacity=0.75,  mediumpurple152142213, forget plot]
table {%
-2.14737014874828 2.60847232010359
-2.11945301689188 3.35925550697892
-2.09153588503547 4.17044103976609
-2.06361875317906 4.95206098433153
-2.03570162132266 5.5402996561364
-2.00778448946625 5.78254178596176
-1.97986735760984 5.68022679419386
-1.95195022575344 5.4828599041202
-1.92403309389703 5.63768955663305
-1.89611596204062 6.59457562966829
-1.86819883018422 8.56788729763853
-1.84028169832781 11.3966073892236
-1.8123645664714 14.6039523367515
-1.784447434615 17.654370405937
-1.75653030275859 20.2753076895782
-1.72861317090219 22.6343581465835
-1.70069603904578 25.2272708350004
-1.67277890718937 28.5263466821342
-1.64486177533297 32.6221163363806
-1.61694464347656 37.118559164493
-1.58902751162015 41.3867530191996
-1.56111037976375 45.0358814305418
-1.53319324790734 48.2769138033385
-1.50527611605093 51.8874123073607
-1.47735898419453 56.7699420717336
-1.44944185233812 63.4308014949012
-1.42152472048171 71.7810490227265
-1.39360758862531 81.3889558417782
-1.3656904567689 91.9459547177795
-1.33777332491249 103.577085337945
-1.30985619305609 116.817222033978
-1.28193906119968 132.386047585198
-1.25402192934327 151.033077094361
-1.22610479748687 173.5501831204
-1.19818766563046 200.751497150351
-1.17027053377406 233.17934583597
-1.14235340191765 270.627201544143
-1.11443627006124 311.897024301534
-1.08651913820484 355.090491590284
-1.05860200634843 398.282612550818
-1.03068487449202 440.178742663113
-1.00276774263562 480.528515185702
-0.97485061077921 520.300476062002
-0.946933478922803 561.597747883869
-0.919016347066397 607.244891301482
-0.89109921520999 660.191272987734
-0.863182083353584 723.087491431839
-0.835264951497177 798.138503713983
-0.807347819640771 886.832521522732
-0.779430687784364 989.165962901824
-0.751513555927958 1102.75579893817
-0.723596424071551 1222.89395612353
-0.695679292215145 1344.16751456475
-0.667762160358738 1462.92798917949
-0.639845028502332 1578.97551491114
-0.611927896645925 1695.35232262791
-0.584010764789519 1816.6062328689
-0.556093632933112 1946.81571362301
-0.528176501076706 2088.35907722276
-0.500259369220299 2241.59264240353
-0.472342237363893 2405.2239213198
-0.444425105507486 2577.14984866047
-0.41650797365108 2755.29796078771
-0.388590841794673 2937.75215733487
-0.360673709938267 3121.96204514197
-0.33275657808186 3304.04596924702
-0.304839446225454 3479.69212722981
-0.276922314369047 3646.83038086264
-0.249005182512641 3808.06823022968
-0.221088050656234 3970.3445433863
-0.193170918799828 4141.51401451414
-0.165253786943421 4326.55183308099
-0.137336655087015 4526.38137955967
-0.109419523230608 4739.47023359045
-0.0815023913742015 4963.65232217586
-0.0535852595177952 5196.16695390929
-0.0256681276613886 5432.54060078503
0.00224900419501806 5666.18730308905
0.0301661360514243 5889.37114579361
0.0580832679078309 6094.69533878074
0.0860003997642371 6276.48230225198
0.113917531620644 6432.35811012534
0.14183466347705 6565.11278195454
0.169751795333457 6683.53515215179
0.197668927189863 6800.43731014268
0.22558605904627 6927.76121403856
0.253503190902676 7071.21466465664
0.281420322759083 7227.75129606874
0.309337454615489 7387.43939111798
0.337254586471896 7538.49231400213
0.365171718328302 7672.5939808359
0.393088850184709 7787.73836239958
0.421005982041115 7887.17272704934
0.448923113897522 7975.25592407484
0.476840245753928 8053.08414692731
0.504757377610335 8116.82996039781
0.532674509466741 8159.49964507351
0.560591641323148 8174.40889426659
0.588508773179554 8158.26297260423
0.616425905035961 8112.7257570255
0.644343036892367 8043.99429631486
0.672260168748774 7960.20917751084
0.70017730060518 7867.65937516628
0.728094432461587 7768.07777671798
0.756011564317993 7658.98981630808
0.7839286961744 7536.89346534449
0.811845828030807 7400.95105054391
0.839762959887213 7254.53111956882
0.867680091743619 7103.45404749658
0.895597223600026 6952.22705978598
0.923514355456432 6801.28863650812
0.951431487312839 6647.59086749826
0.979348619169246 6487.80027168665
1.00726575102565 6320.85635938337
1.03518288288206 6147.52841347054
1.06310001473846 5968.09869088248
1.09101714659487 5781.30226674554
1.11893427845128 5585.88916092312
1.14685141030768 5382.9847430385
1.17476854216409 5176.57987845915
1.2026856740205 4971.53418528141
1.2306028058769 4770.77821499436
1.25851993773331 4573.81811098577
1.28643706958972 4377.44984841513
1.31435420144612 4178.07240543072
1.34227133330253 3974.02802874456
1.37018846515894 3766.48920109597
1.39810559701534 3558.59287598861
1.42602272887175 3353.86422600365
1.45393986072816 3155.19870163385
1.48185699258456 2964.68179060538
1.50977412444097 2783.60796730246
1.53769125629738 2612.24803573893
1.56560838815378 2449.70819842221
1.59352552001019 2294.41970999656
1.62144265186659 2145.14547534308
1.649359783723 2001.77113762053
1.67727691557941 1865.28129394673
1.70519404743581 1736.9626905742
1.73311117929222 1617.36378284274
1.76102831114863 1505.60359239286
1.78894544300503 1399.42440693915
1.81686257486144 1296.06854288779
1.84477970671785 1193.62593407265
1.87269683857425 1092.08709794258
1.90061397043066 993.338474887626
1.92853110228707 899.989685130601
1.95644823414347 813.841626486498
1.98436536599988 735.158506162486
2.01228249785629 663.234742442461
2.04019962971269 597.595694387281
2.0681167615691 538.661628082008
2.09603389342551 487.320574900923
2.12395102528191 443.885596761632
2.15186815713832 407.320614889874
2.17978528899472 375.231719868258
2.20770242085113 344.582327148532
2.23561955270754 312.839604248445
2.26353668456394 279.104824561908
2.29145381642035 244.575230401225
2.31937094827676 211.825363026871
2.34728808013316 183.185968300001
2.37520521198957 159.359226465156
2.40312234384598 139.34499117215
2.43103947570238 121.63941381553
2.45895660755879 105.572553953411
2.4868737394152 91.6607646764978
2.5147908712716 80.8618247297197
2.54270800312801 73.5508209708654
2.57062513498442 69.0603018354481
2.59854226684082 65.9658725267144
2.62645939869723 62.7341630589843
2.65437653055363 58.2924076196441
2.68229366241004 52.3090911373528
2.71021079426645 45.1605251148105
2.73812792612285 37.6485687101676
2.76604505797926 30.6136173929131
2.79396218983567 24.6269146145212
2.82187932169207 19.8755465163051
2.84979645354848 16.2230585718807
2.87771358540489 13.3581862981504
2.90563071726129 10.9584951502246
2.9335478491177 8.82349665687729
2.96146498097411 6.93517552852493
2.98938211283051 5.41708907735661
3.01729924468692 4.41199688957909
3.04521637654333 3.95099163240655
3.07313350839973 3.89632815260667
3.10105064025614 3.99274364265245
3.12896777211255 3.99024422000379
3.15688490396895 3.75843077933601
3.18480203582536 3.32928810366021
3.21271916768177 2.85959542020707
3.24063629953817 2.54559280984264
3.26855343139458 2.52978203851669
3.29647056325098 2.8312030956639
3.32438769510739 3.32429493523445
3.3523048269638 3.78138332555745
3.3802219588202 3.96879070056969
3.40813909067661 3.75378605364944
};
\addplot [thick, opacity=0.75,  gray119, forget plot]
table {%
0.00305815222802982 826.909703411522
0.0360917932572491 1028.27484882227
0.0691254342864684 1251.90318420926
0.102159075315688 1493.67422444108
0.135192716344907 1748.17812715198
0.168226357374126 2009.03584503247
0.201259998403346 2269.31911383292
0.234293639432565 2522.0274859667
0.267327280461784 2760.57414419022
0.300360921491004 2979.23231311811
0.333394562520223 3173.49909019649
0.366428203549442 3340.34279823135
0.399461844578662 3478.31287161299
0.432495485607881 3587.50686045868
0.4655291266371 3669.40574650302
0.498562767666319 3726.60409907023
0.531596408695539 3762.47307228753
0.564630049724758 3780.79972428984
0.597663690753977 3785.44467025218
0.630697331783197 3780.0522648533
0.663730972812416 3767.83536908691
0.696764613841635 3751.44313914393
0.729798254870855 3732.90802679442
0.762831895900074 3713.65937696691
0.795865536929293 3694.58648922725
0.828899177958513 3676.13333994192
0.861932818987732 3658.40900148948
0.894966460016951 3641.30054281999
0.928000101046171 3624.57766277842
0.96103374207539 3607.98015222915
0.994067383104609 3591.28101110883
1.02710102413383 3574.32057400814
1.06013466516305 3557.01095400712
1.09316830619227 3539.31528372733
1.12620194722149 3521.21143154835
1.15923558825071 3502.6533778925
1.19226922927992 3483.5437743277
1.22530287030914 3463.72787817559
1.25833651133836 3443.0128363593
1.29137015236758 3421.20896080572
1.3244037933968 3398.18327408632
1.35743743442602 3373.91185105284
1.39047107545524 3348.5170969453
1.42350471648446 3322.27892869153
1.45653835751368 3295.61407209263
1.4895719985429 3269.02426542055
1.52260563957212 3243.02089893356
1.55563928060134 3218.0393300788
1.58867292163056 3194.35962228203
1.62170656265978 3172.05075457498
1.654740203689 3150.95190072318
1.68777384471821 3130.69744367936
1.72080748574743 3110.78320932134
1.75384112677665 3090.66202136669
1.78687476780587 3069.84946148087
1.81990840883509 3048.01765871642
1.85294204986431 3025.05702651603
1.88597569089353 3001.09277060684
1.91900933192275 2976.45308470502
1.95204297295197 2951.59676206808
1.98507661398119 2927.01682210387
2.01811025501041 2903.14154027673
2.05114389603963 2880.25390665555
2.08417753706885 2858.44529730063
2.11721117809807 2837.61055073776
2.15024481912728 2817.48204099783
2.1832784601565 2797.69220723727
2.21631210118572 2777.8492322823
2.24934574221494 2757.60994178149
2.28237938324416 2736.73704364374
2.31541302427338 2715.13308558197
2.3484466653026 2692.84913941381
2.38148030633182 2670.070687834
2.41451394736104 2647.08574522607
2.44754758839026 2624.24104331326
2.48058122941948 2601.89193304385
2.5136148704487 2580.35138172918
2.54664851147792 2559.84357215536
2.57968215250714 2540.46798917619
2.61271579353636 2522.1798302646
2.64574943456557 2504.79130142138
2.67878307559479 2487.99544333411
2.71181671662401 2471.40988398658
2.74485035765323 2454.63337176057
2.77788399868245 2437.30452777786
2.81091763971167 2419.15122015966
2.84395128074089 2400.0208809894
2.87698492177011 2379.88660229196
2.91001856279933 2358.82976422319
2.94305220382855 2337.00561280627
2.97608584485777 2314.60201560079
3.00911948588699 2291.80250395765
3.04215312691621 2268.76240926435
3.07518676794543 2245.60207999714
3.10822040897465 2222.41526006456
3.14125405000386 2199.28556607287
3.17428769103308 2176.30136034135
3.2073213320623 2153.56021755971
3.24035497309152 2131.15850461241
3.27338861412074 2109.16793621135
3.30642225514996 2087.60702801336
3.33945589617918 2066.41874162568
3.3724895372084 2045.46477287924
3.40552317823762 2024.54193173554
3.43855681926684 2003.41863434845
3.47159046029606 1981.88245844116
3.50462410132528 1959.78571018765
3.5376577423545 1937.07651741513
3.57069138338372 1913.80777640675
3.60372502441293 1890.12335888388
3.63675866544215 1866.22758006537
3.66979230647137 1842.34769339015
3.70282594750059 1818.69911338857
3.73585958852981 1795.4598135665
3.76889322955903 1772.75573709474
3.80192687058825 1750.65521148456
3.83496051161747 1729.16872738027
3.86799415264669 1708.25126270472
3.90102779367591 1687.80666722101
3.93406143470513 1667.69590475159
3.96709507573435 1647.75176001194
4.00012871676357 1627.80131452397
4.03316235779279 1607.69447911602
4.066195998822 1587.33338232792
4.09922963985122 1566.69501568614
4.13226328088044 1545.83946883622
4.16529692190966 1524.89876593338
4.19833056293888 1504.04614256926
4.2313642039681 1483.4511589452
4.26439784499732 1463.23057663327
4.29743148602654 1443.40695708211
4.33046512705576 1423.88577312518
4.36349876808498 1404.45775562274
4.3965324091142 1384.82737562704
4.42956605014342 1364.66237808058
4.46259969117264 1343.65465137375
4.49563333220186 1321.58045368495
4.52866697323107 1298.34843065351
4.56170061426029 1274.02658786076
4.59473425528951 1248.84360786995
4.62776789631873 1223.16463386727
4.66080153734795 1197.44596564598
4.69383517837717 1172.17634864354
4.72686881940639 1147.81429266356
4.75990246043561 1124.73101406798
4.79293610146483 1103.16725436574
4.82596974249405 1083.20966586822
4.85900338352327 1064.78909684419
4.89203702455249 1047.69950354284
4.92507066558171 1031.63298471447
4.95810430661093 1016.22416696442
4.99113794764015 1001.09629220614
5.02417158866936 985.902003180416
5.05720522969858 970.353762273555
5.0902388707278 954.241526784349
5.12327251175702 937.438021629339
5.15630615278624 919.894017638727
5.18933979381546 901.627005883581
5.22237343484468 882.706502106431
5.2554070758739 863.238253925547
5.28844071690312 843.348443136682
5.32147435793234 823.168174094661
5.35450799896156 802.818467562808
5.38754163999078 782.396577295289
5.42057528102 761.965264603807
5.45360892204922 741.54708549491
5.48664256307844 721.125303165053
5.51967620410765 700.651697718067
5.55270984513687 680.059750682535
5.58574348616609 659.280165416229
5.61877712719531 638.255100371339
5.65181076822453 616.9480857045
5.68484440925375 595.348081174959
5.71787805028297 573.467862120028
5.75091169131219 551.338216092848
5.78394533234141 528.999934757027
5.81697897337063 506.495383786479
5.85001261439985 483.860940678085
5.88304625542907 461.121240956205
5.91607989645829 438.286138095489
5.94911353748751 415.351373359111
5.98214717851672 392.303769038682
6.01518081954594 369.130987857926
6.04821446057516 345.83457479044
6.08124810160438 322.443557300724
6.1142817426336 299.024994694315
6.14731538366282 275.688095344445
6.18034902469204 252.579999676781
6.21338266572126 229.873655232623
6.24641630675048 207.750585406292
6.2794499477797 186.382902058028
6.31248358880892 165.919054135327
6.34551722983814 146.476487584695
6.37855087086736 128.142101077558
6.41158451189658 110.978933929969
6.44461815292579 95.0357455370404
6.47765179395501 80.3555815604938
6.51068543498423 66.9801597558271
6.54371907601345 54.9485950671284
6.57675271704267 44.2909898644455
};
\end{axis}

\end{tikzpicture}

%% file: open_0_paper.tex
\begin{tikzpicture}

\definecolor{chocolate2267451}{RGB}{50,50,50}
\definecolor{dimgray85}{RGB}{85,85,85}
\definecolor{gainsboro229}{RGB}{239,239,239}
\definecolor{gainsboro247}{RGB}{253,253,253}
\definecolor{darkorange2551094}{RGB}{255,109,4}
\definecolor{gray119}{RGB}{255,109,4}
\definecolor{mediumpurple152142213}{RGB}{142,186,66}
\definecolor{yellowgreen14218666}{RGB}{142,186,66}
\definecolor{steelblue52138189}{RGB}{52,138,189}
\definecolor{whitesmoke238}{RGB}{238,238,238}


\definecolor{lightgray204}{RGB}{204,204,204}

\begin{axis}[
width=5.1cm,
height=4cm,
ymajorgrids,
y grid style={gainsboro229},
yticklabels={$\text{\scriptsize 0}$, $\text{\scriptsize 2,500}$, $\text{\scriptsize 5,000}$},
legend cell align={left},
legend style={
	fill opacity=0,
	draw opacity=0,
  text opacity=1,
  anchor=north east,
  draw=lightgray204,
  fill=gainsboro247,
  at={(1.04,1.01)},
},
axis background/.style={fill=gainsboro247},
scaled y ticks=false,
ylabel near ticks,
ytick style={color=white},
xtick pos=left,
axis line style={white},
tick align=outside,
x grid style={white},
xmajorgrids,
xmin=-1.55, xmax=1.56,
xtick style={color=dimgray85},
ymajorgrids,
ymin=0, ymax=6500,
ticklabel style={font=\tiny},
tick align=inside,
axis line style={gainsboro229},
ytick={2000,4000,6000},
yticklabels={2\textrm{K}, 4\textrm{K}, 6\textrm{K}},
ylabel=\textcolor{white}{.},
y label style={at={(axis description cs:0.2, 0.5)}},
yticklabel style = {xshift=0.5ex},
xticklabel style = {yshift=0.5ex},
]
\draw[draw=whitesmoke238,fill=chocolate2267451,fill opacity=0.5,very thin] (axis cs:-1.46022880077362,0) rectangle (axis cs:-1.34216928482056,11);
\addlegendimage{ybar,ybar legend,draw=whitesmoke238,fill=chocolate2267451,fill opacity=0.5,very thin}
\addlegendentry{\tiny CLIP}

\draw[draw=whitesmoke238,fill=chocolate2267451,fill opacity=0.5,very thin] (axis cs:-1.34216928482056,0) rectangle (axis cs:-1.22410976886749,28);
\draw[draw=whitesmoke238,fill=chocolate2267451,fill opacity=0.5,very thin] (axis cs:-1.22410976886749,0) rectangle (axis cs:-1.10605025291443,81);
\draw[draw=whitesmoke238,fill=chocolate2267451,fill opacity=0.5,very thin] (axis cs:-1.10605025291443,0) rectangle (axis cs:-0.987990736961365,193);
\draw[draw=whitesmoke238,fill=chocolate2267451,fill opacity=0.5,very thin] (axis cs:-0.987990736961365,0) rectangle (axis cs:-0.869931221008301,488);
\draw[draw=whitesmoke238,fill=chocolate2267451,fill opacity=0.5,very thin] (axis cs:-0.869931221008301,0) rectangle (axis cs:-0.751871705055237,1044);
\draw[draw=whitesmoke238,fill=chocolate2267451,fill opacity=0.5,very thin] (axis cs:-0.751871705055237,0) rectangle (axis cs:-0.633812189102173,1849);
\draw[draw=whitesmoke238,fill=chocolate2267451,fill opacity=0.5,very thin] (axis cs:-0.633812189102173,0) rectangle (axis cs:-0.515752673149109,2934);
\draw[draw=whitesmoke238,fill=chocolate2267451,fill opacity=0.5,very thin] (axis cs:-0.515752673149109,0) rectangle (axis cs:-0.397693157196045,4142);
\draw[draw=whitesmoke238,fill=chocolate2267451,fill opacity=0.5,very thin] (axis cs:-0.397693157196045,0) rectangle (axis cs:-0.279633641242981,5287);
\draw[draw=whitesmoke238,fill=chocolate2267451,fill opacity=0.5,very thin] (axis cs:-0.279633641242981,0) rectangle (axis cs:-0.161574125289917,6123);
\draw[draw=whitesmoke238,fill=chocolate2267451,fill opacity=0.5,very thin] (axis cs:-0.161574125289917,0) rectangle (axis cs:-0.043514609336853,6379);
\draw[draw=whitesmoke238,fill=chocolate2267451,fill opacity=0.5,very thin] (axis cs:-0.043514609336853,0) rectangle (axis cs:0.0745449066162109,5945);
\draw[draw=whitesmoke238,fill=chocolate2267451,fill opacity=0.5,very thin] (axis cs:0.0745449066162109,0) rectangle (axis cs:0.192604422569275,5106);
\draw[draw=whitesmoke238,fill=chocolate2267451,fill opacity=0.5,very thin] (axis cs:0.192604422569275,0) rectangle (axis cs:0.310663938522339,3869);
\draw[draw=whitesmoke238,fill=chocolate2267451,fill opacity=0.5,very thin] (axis cs:0.310663938522339,0) rectangle (axis cs:0.428723454475403,2753);
\draw[draw=whitesmoke238,fill=chocolate2267451,fill opacity=0.5,very thin] (axis cs:0.428723454475403,0) rectangle (axis cs:0.546782970428467,1744);
\draw[draw=whitesmoke238,fill=chocolate2267451,fill opacity=0.5,very thin] (axis cs:0.546782970428467,0) rectangle (axis cs:0.664842486381531,964);
\draw[draw=whitesmoke238,fill=chocolate2267451,fill opacity=0.5,very thin] (axis cs:0.664842486381531,0) rectangle (axis cs:0.782902002334595,544);
\draw[draw=whitesmoke238,fill=chocolate2267451,fill opacity=0.5,very thin] (axis cs:0.782902002334595,0) rectangle (axis cs:0.900961518287659,288);
\draw[draw=whitesmoke238,fill=chocolate2267451,fill opacity=0.5,very thin] (axis cs:0.900961518287659,0) rectangle (axis cs:1.01902103424072,135);
\draw[draw=whitesmoke238,fill=chocolate2267451,fill opacity=0.5,very thin] (axis cs:1.01902103424072,0) rectangle (axis cs:1.13708055019379,60);
\draw[draw=whitesmoke238,fill=chocolate2267451,fill opacity=0.5,very thin] (axis cs:1.13708055019379,0) rectangle (axis cs:1.25514006614685,24);
\draw[draw=whitesmoke238,fill=chocolate2267451,fill opacity=0.5,very thin] (axis cs:1.25514006614685,0) rectangle (axis cs:1.37319958209991,6);
\draw[draw=whitesmoke238,fill=chocolate2267451,fill opacity=0.5,very thin] (axis cs:1.37319958209991,0) rectangle (axis cs:1.49125909805298,3);
\draw[draw=whitesmoke238,fill=steelblue52138189,fill opacity=0.25,very thin] (axis cs:-1.46021542604527,0) rectangle (axis cs:-1.34216479814424,1991);
\addlegendimage{ybar,ybar legend,draw=whitesmoke238,fill=steelblue52138189,fill opacity=0.5,very thin}
\addlegendentry{\tiny Uniform}

\draw[draw=whitesmoke238,fill=steelblue52138189,fill opacity=0.25,very thin] (axis cs:-1.34216479814424,0) rectangle (axis cs:-1.22411417024321,1982);
\draw[draw=whitesmoke238,fill=steelblue52138189,fill opacity=0.25,very thin] (axis cs:-1.22411417024321,0) rectangle (axis cs:-1.10606354234218,1982);
\draw[draw=whitesmoke238,fill=steelblue52138189,fill opacity=0.25,very thin] (axis cs:-1.10606354234218,0) rectangle (axis cs:-0.988012914441148,1983);
\draw[draw=whitesmoke238,fill=steelblue52138189,fill opacity=0.25,very thin] (axis cs:-0.988012914441148,0) rectangle (axis cs:-0.869962286540118,1965);
\draw[draw=whitesmoke238,fill=steelblue52138189,fill opacity=0.25,very thin] (axis cs:-0.869962286540119,0) rectangle (axis cs:-0.751911658639089,1975);
\draw[draw=whitesmoke238,fill=steelblue52138189,fill opacity=0.25,very thin] (axis cs:-0.751911658639089,0) rectangle (axis cs:-0.633861030738059,1934);
\draw[draw=whitesmoke238,fill=steelblue52138189,fill opacity=0.25,very thin] (axis cs:-0.633861030738059,0) rectangle (axis cs:-0.51581040283703,2098);
\draw[draw=whitesmoke238,fill=steelblue52138189,fill opacity=0.25,very thin] (axis cs:-0.51581040283703,0) rectangle (axis cs:-0.397759774936,1971);
\draw[draw=whitesmoke238,fill=steelblue52138189,fill opacity=0.25,very thin] (axis cs:-0.397759774936,0) rectangle (axis cs:-0.27970914703497,2035);
\draw[draw=whitesmoke238,fill=steelblue52138189,fill opacity=0.25,very thin] (axis cs:-0.27970914703497,0) rectangle (axis cs:-0.161658519133941,2025);
\draw[draw=whitesmoke238,fill=steelblue52138189,fill opacity=0.25,very thin] (axis cs:-0.161658519133941,0) rectangle (axis cs:-0.0436078912329112,2000);
\draw[draw=whitesmoke238,fill=steelblue52138189,fill opacity=0.25,very thin] (axis cs:-0.0436078912329112,0) rectangle (axis cs:0.0744427366681184,1926);
\draw[draw=whitesmoke238,fill=steelblue52138189,fill opacity=0.25,very thin] (axis cs:0.0744427366681184,0) rectangle (axis cs:0.192493364569148,2002);
\draw[draw=whitesmoke238,fill=steelblue52138189,fill opacity=0.25,very thin] (axis cs:0.192493364569148,0) rectangle (axis cs:0.310543992470178,1969);
\draw[draw=whitesmoke238,fill=steelblue52138189,fill opacity=0.25,very thin] (axis cs:0.310543992470178,0) rectangle (axis cs:0.428594620371207,2000);
\draw[draw=whitesmoke238,fill=steelblue52138189,fill opacity=0.25,very thin] (axis cs:0.428594620371207,0) rectangle (axis cs:0.546645248272237,1999);
\draw[draw=whitesmoke238,fill=steelblue52138189,fill opacity=0.25,very thin] (axis cs:0.546645248272237,0) rectangle (axis cs:0.664695876173267,2054);
\draw[draw=whitesmoke238,fill=steelblue52138189,fill opacity=0.25,very thin] (axis cs:0.664695876173267,0) rectangle (axis cs:0.782746504074296,2017);
\draw[draw=whitesmoke238,fill=steelblue52138189,fill opacity=0.25,very thin] (axis cs:0.782746504074296,0) rectangle (axis cs:0.900797131975326,2030);
\draw[draw=whitesmoke238,fill=steelblue52138189,fill opacity=0.25,very thin] (axis cs:0.900797131975326,0) rectangle (axis cs:1.01884775987636,1956);
\draw[draw=whitesmoke238,fill=steelblue52138189,fill opacity=0.25,very thin] (axis cs:1.01884775987636,0) rectangle (axis cs:1.13689838777739,2024);
\draw[draw=whitesmoke238,fill=steelblue52138189,fill opacity=0.25,very thin] (axis cs:1.13689838777739,0) rectangle (axis cs:1.25494901567841,2031);
\draw[draw=whitesmoke238,fill=steelblue52138189,fill opacity=0.25,very thin] (axis cs:1.25494901567841,0) rectangle (axis cs:1.37299964357944,2001);
\draw[draw=whitesmoke238,fill=steelblue52138189,fill opacity=0.25,very thin] (axis cs:1.37299964357944,0) rectangle (axis cs:1.49105027148047,2050);
\draw[draw=whitesmoke238,fill=mediumpurple152142213,fill opacity=0.25,very thin] (axis cs:-1.5059016547968,0) rectangle (axis cs:-1.38758267619478,11);
\addlegendimage{ybar,ybar legend,draw=whitesmoke238,fill=mediumpurple152142213,fill opacity=0.5,very thin}
\addlegendentry{\tiny Gaussian}

\draw[draw=whitesmoke238,fill=mediumpurple152142213,fill opacity=0.25,very thin] (axis cs:-1.38758267619478,0) rectangle (axis cs:-1.26926369759276,26);
\draw[draw=whitesmoke238,fill=mediumpurple152142213,fill opacity=0.25,very thin] (axis cs:-1.26926369759276,0) rectangle (axis cs:-1.15094471899074,58);
\draw[draw=whitesmoke238,fill=mediumpurple152142213,fill opacity=0.25,very thin] (axis cs:-1.15094471899074,0) rectangle (axis cs:-1.03262574038871,186);
\draw[draw=whitesmoke238,fill=mediumpurple152142213,fill opacity=0.25,very thin] (axis cs:-1.03262574038871,0) rectangle (axis cs:-0.914306761786694,393);
\draw[draw=whitesmoke238,fill=mediumpurple152142213,fill opacity=0.25,very thin] (axis cs:-0.914306761786694,0) rectangle (axis cs:-0.795987783184673,829);
\draw[draw=whitesmoke238,fill=mediumpurple152142213,fill opacity=0.25,very thin] (axis cs:-0.795987783184673,0) rectangle (axis cs:-0.677668804582652,1469);
\draw[draw=whitesmoke238,fill=mediumpurple152142213,fill opacity=0.25,very thin] (axis cs:-0.677668804582652,0) rectangle (axis cs:-0.559349825980631,2434);
\draw[draw=whitesmoke238,fill=mediumpurple152142213,fill opacity=0.25,very thin] (axis cs:-0.559349825980631,0) rectangle (axis cs:-0.44103084737861,3625);
\draw[draw=whitesmoke238,fill=mediumpurple152142213,fill opacity=0.25,very thin] (axis cs:-0.44103084737861,0) rectangle (axis cs:-0.322711868776589,4759);
\draw[draw=whitesmoke238,fill=mediumpurple152142213,fill opacity=0.25,very thin] (axis cs:-0.322711868776589,0) rectangle (axis cs:-0.204392890174568,5798);
\draw[draw=whitesmoke238,fill=mediumpurple152142213,fill opacity=0.25,very thin] (axis cs:-0.204392890174568,0) rectangle (axis cs:-0.0860739115725466,6279);
\draw[draw=whitesmoke238,fill=mediumpurple152142213,fill opacity=0.25,very thin] (axis cs:-0.0860739115725466,0) rectangle (axis cs:0.0322450670294745,6179);
\draw[draw=whitesmoke238,fill=mediumpurple152142213,fill opacity=0.25,very thin] (axis cs:0.0322450670294745,0) rectangle (axis cs:0.150564045631496,5570);
\draw[draw=whitesmoke238,fill=mediumpurple152142213,fill opacity=0.25,very thin] (axis cs:0.150564045631496,0) rectangle (axis cs:0.268883024233517,4476);
\draw[draw=whitesmoke238,fill=mediumpurple152142213,fill opacity=0.25,very thin] (axis cs:0.268883024233517,0) rectangle (axis cs:0.387202002835538,3230);
\draw[draw=whitesmoke238,fill=mediumpurple152142213,fill opacity=0.25,very thin] (axis cs:0.387202002835538,0) rectangle (axis cs:0.505520981437559,2146);
\draw[draw=whitesmoke238,fill=mediumpurple152142213,fill opacity=0.25,very thin] (axis cs:0.505520981437559,0) rectangle (axis cs:0.62383996003958,1285);
\draw[draw=whitesmoke238,fill=mediumpurple152142213,fill opacity=0.25,very thin] (axis cs:0.62383996003958,0) rectangle (axis cs:0.742158938641601,690);
\draw[draw=whitesmoke238,fill=mediumpurple152142213,fill opacity=0.25,very thin] (axis cs:0.742158938641601,0) rectangle (axis cs:0.860477917243622,330);
\draw[draw=whitesmoke238,fill=mediumpurple152142213,fill opacity=0.25,very thin] (axis cs:0.860477917243622,0) rectangle (axis cs:0.978796895845643,146);
\draw[draw=whitesmoke238,fill=mediumpurple152142213,fill opacity=0.25,very thin] (axis cs:0.978796895845643,0) rectangle (axis cs:1.09711587444766,52);
\draw[draw=whitesmoke238,fill=mediumpurple152142213,fill opacity=0.25,very thin] (axis cs:1.09711587444766,0) rectangle (axis cs:1.21543485304968,20);
\draw[draw=whitesmoke238,fill=mediumpurple152142213,fill opacity=0.25,very thin] (axis cs:1.21543485304968,0) rectangle (axis cs:1.33375383165171,7);
\draw[draw=whitesmoke238,fill=mediumpurple152142213,fill opacity=0.25,very thin] (axis cs:1.33375383165171,0) rectangle (axis cs:1.45207281025373,2);
\draw[draw=whitesmoke238,fill=gray119,fill opacity=0.5,very thin] (axis cs:-1.458784483609,0) rectangle (axis cs:-1.34101106828416,186);
\addlegendimage{ybar,ybar legend,draw=whitesmoke238,fill=gray119,fill opacity=0.5,very thin}
\addlegendentry{\tiny Triangular}

\draw[draw=whitesmoke238,fill=gray119,fill opacity=0.5,very thin] (axis cs:-1.34101106828416,0) rectangle (axis cs:-1.22323765295932,497);
\draw[draw=whitesmoke238,fill=gray119,fill opacity=0.5,very thin] (axis cs:-1.22323765295932,0) rectangle (axis cs:-1.10546423763448,846);
\draw[draw=whitesmoke238,fill=gray119,fill opacity=0.5,very thin] (axis cs:-1.10546423763448,0) rectangle (axis cs:-0.987690822309644,1258);
\draw[draw=whitesmoke238,fill=gray119,fill opacity=0.5,very thin] (axis cs:-0.987690822309644,0) rectangle (axis cs:-0.869917406984805,1591);
\draw[draw=whitesmoke238,fill=gray119,fill opacity=0.5,very thin] (axis cs:-0.869917406984805,0) rectangle (axis cs:-0.752143991659965,1889);
\draw[draw=whitesmoke238,fill=gray119,fill opacity=0.5,very thin] (axis cs:-0.752143991659965,0) rectangle (axis cs:-0.634370576335126,2226);
\draw[draw=whitesmoke238,fill=gray119,fill opacity=0.5,very thin] (axis cs:-0.634370576335126,0) rectangle (axis cs:-0.516597161010287,2576);
\draw[draw=whitesmoke238,fill=gray119,fill opacity=0.5,very thin] (axis cs:-0.516597161010287,0) rectangle (axis cs:-0.398823745685448,2929);
\draw[draw=whitesmoke238,fill=gray119,fill opacity=0.5,very thin] (axis cs:-0.398823745685448,0) rectangle (axis cs:-0.281050330360608,3243);
\draw[draw=whitesmoke238,fill=gray119,fill opacity=0.5,very thin] (axis cs:-0.281050330360608,0) rectangle (axis cs:-0.163276915035769,3652);
\draw[draw=whitesmoke238,fill=gray119,fill opacity=0.5,very thin] (axis cs:-0.163276915035769,0) rectangle (axis cs:-0.0455034997109298,3917);
\draw[draw=whitesmoke238,fill=gray119,fill opacity=0.5,very thin] (axis cs:-0.0455034997109298,0) rectangle (axis cs:0.0722699156139095,3793);
\draw[draw=whitesmoke238,fill=gray119,fill opacity=0.5,very thin] (axis cs:0.0722699156139095,0) rectangle (axis cs:0.190043330938749,3395);
\draw[draw=whitesmoke238,fill=gray119,fill opacity=0.5,very thin] (axis cs:0.190043330938749,0) rectangle (axis cs:0.307816746263588,3074);
\draw[draw=whitesmoke238,fill=gray119,fill opacity=0.5,very thin] (axis cs:0.307816746263588,0) rectangle (axis cs:0.425590161588427,2894);
\draw[draw=whitesmoke238,fill=gray119,fill opacity=0.5,very thin] (axis cs:0.425590161588427,0) rectangle (axis cs:0.543363576913267,2505);
\draw[draw=whitesmoke238,fill=gray119,fill opacity=0.5,very thin] (axis cs:0.543363576913267,0) rectangle (axis cs:0.661136992238106,2144);
\draw[draw=whitesmoke238,fill=gray119,fill opacity=0.5,very thin] (axis cs:0.661136992238106,0) rectangle (axis cs:0.778910407562945,1863);
\draw[draw=whitesmoke238,fill=gray119,fill opacity=0.5,very thin] (axis cs:0.778910407562945,0) rectangle (axis cs:0.896683822887784,1594);
\draw[draw=whitesmoke238,fill=gray119,fill opacity=0.5,very thin] (axis cs:0.896683822887784,0) rectangle (axis cs:1.01445723821262,1459);
\draw[draw=whitesmoke238,fill=gray119,fill opacity=0.5,very thin] (axis cs:1.01445723821262,0) rectangle (axis cs:1.13223065353746,1094);
\draw[draw=whitesmoke238,fill=gray119,fill opacity=0.5,very thin] (axis cs:1.13223065353746,0) rectangle (axis cs:1.2500040688623,753);
\draw[draw=whitesmoke238,fill=gray119,fill opacity=0.5,very thin] (axis cs:1.2500040688623,0) rectangle (axis cs:1.36777748418714,458);
\draw[draw=whitesmoke238,fill=gray119,fill opacity=0.5,very thin] (axis cs:1.36777748418714,0) rectangle (axis cs:1.48555089951198,164);
\addplot [thick, opacity=0.75, chocolate2267451, forget plot]
table {%
-1.46022880077362 4.73265418476401
-1.44539720329208 6.17447637552127
-1.43056560581054 7.61799980336397
-1.415734008329 8.98766097180421
-1.40090241084746 10.2688886730048
-1.38607081336592 11.5135967467377
-1.37123921588438 12.8184524396764
-1.35640761840284 14.296338734906
-1.3415760209213 16.0678883209865
-1.32674442343975 18.2764583346499
-1.31191282595821 21.0963356658176
-1.29708122847667 24.7019719253193
-1.28224963099513 29.2072644481325
-1.26741803351359 34.6219634154988
-1.25258643603205 40.8537665257498
-1.23775483855051 47.7291650400199
-1.22292324106897 54.9959219409026
-1.20809164358743 62.334257727697
-1.19326004610589 69.4534744613457
-1.17842844862435 76.2866906933957
-1.16359685114281 83.1623692354613
-1.14876525366127 90.7867676492078
-1.13393365617973 99.9958507433577
-1.11910205869818 111.420643460369
-1.10427046121664 125.281773820154
-1.0894388637351 141.432456342342
-1.07460726625356 159.601430078422
-1.05977566877202 179.6772096713
-1.04494407129048 201.876605011849
-1.03011247380894 226.724432039765
-1.0152808763274 254.870754815214
-1.00044927884586 286.835404817798
-0.985617681364318 322.79410819993
-0.970786083882778 362.525414949678
-0.955954486401237 405.59934732331
-0.941122888919696 451.757240900517
-0.926291291438155 501.252398147412
-0.911459693956615 554.8715860684
-0.896628096475074 613.564445371815
-0.881796498993533 677.929669546941
-0.866964901511993 747.921556209045
-0.852133304030452 822.913810170216
-0.837301706548911 901.941897402873
-0.822470109067371 983.902008089589
-0.80763851158583 1067.72570177043
-0.792806914104289 1152.7027998014
-0.777975316622748 1238.93221497689
-0.763143719141208 1327.58365565238
-0.748312121659667 1420.69037138456
-0.733480524178126 1520.56540044838
-0.718648926696586 1629.17359959171
-0.703817329215045 1747.63931025544
-0.688985731733504 1875.85132904864
-0.674154134251963 2012.22116651182
-0.659322536770423 2153.86832970308
-0.644490939288882 2297.39256333426
-0.629659341807341 2439.96752214717
-0.614827744325801 2580.22931796566
-0.59999614684426 2718.56781213263
-0.585164549362719 2856.72330148627
-0.570332951881179 2996.83623625859
-0.555501354399638 3140.33975248146
-0.540669756918097 3287.2685758871
-0.525838159436557 3436.39998595763
-0.511006561955016 3586.08958648887
-0.496174964473475 3735.15501755757
-0.481343366991934 3883.19293696544
-0.466511769510394 4030.28493346245
-0.451680172028853 4176.62037622909
-0.436848574547312 4322.56027839775
-0.422016977065772 4469.04659058544
-0.407185379584231 4617.66842479902
-0.39235378210269 4769.85946732968
-0.37752218462115 4925.58054737159
-0.362690587139609 5082.53648411648
-0.347858989658068 5236.66685987127
-0.333027392176527 5383.56384312292
-0.318195794694987 5519.70772212486
-0.303364197213446 5642.7566446221
-0.288532599731905 5751.16969384699
-0.273701002250365 5844.03296101008
-0.258869404768824 5921.51028244378
-0.244037807287283 5985.4595726827
-0.229206209805743 6039.4561952222
-0.214374612324202 6087.9704843072
-0.199543014842661 6135.08334224924
-0.18471141736112 6183.29776473781
-0.16987981987958 6232.81751792895
-0.155048222398039 6281.46499544802
-0.140216624916498 6325.24061468846
-0.125385027434958 6359.32046386492
-0.110553429953417 6379.19007351028
-0.0957218324718763 6381.6881502999
-0.0808902349903355 6365.73833133396
-0.0660586375087948 6332.39961836222
-0.0512270400272541 6284.00378027655
-0.0363954425457134 6222.8275380525
-0.0215638450641726 6150.32008199436
-0.00673224758263191 6067.51410874677
0.00809934989890859 5976.12410293408
0.0229309473804493 5879.17191227235
0.03776254486199 5780.43533926119
0.0525941423435308 5682.99909707798
0.0674257398250715 5587.77644093711
0.0822573373066122 5492.82751837482
0.0970889347881529 5393.91053337453
0.111920532269694 5286.14822127537
0.126752129751234 5166.07667184572
0.141583727232775 5033.01129213441
0.156415324714316 4888.99345833131
0.171246922195856 4737.45073397582
0.186078519677397 4581.48834813565
0.200910117158938 4422.87857205479
0.215741714640479 4262.28035731187
0.230573312122019 4100.37078375917
0.24540490960356 3938.90807263297
0.2602365070851 3780.7326874885
0.275068104566641 3628.50312916305
0.289899702048182 3483.07907090126
0.304731299529723 3342.93910584148
0.319562897011263 3205.26651946708
0.334394494492804 3067.91939766832
0.349226091974345 2930.73125630129
0.364057689455886 2795.193471075
0.378889286937426 2662.94543132518
0.393720884418967 2534.37307814054
0.408552481900508 2408.32955856515
0.423384079382048 2282.99069511925
0.438215676863589 2157.07687863474
0.45304727434513 2030.62114392198
0.46787887182667 1904.95210289629
0.482710469308211 1782.10441735878
0.497542066789752 1664.12149576803
0.512373664271292 1552.58892870172
0.527205261752833 1448.40144017593
0.542036859234374 1351.57645055893
0.556868456715915 1261.11299402005
0.571700054197455 1175.21410338607
0.586531651678996 1092.12427906709
0.601363249160537 1011.25848749701
0.616194846642077 933.781488645685
0.631026444123618 862.004701346043
0.645858041605159 797.863912629572
0.6606896390867 741.536602938186
0.67552123656824 691.205611559274
0.690352834049781 644.112299161464
0.705184431531321 598.115472674706
0.720016029012862 552.710723258735
0.734847626494403 508.96309080602
0.749679223975944 468.56792843065
0.764510821457484 432.718097027018
0.779342418939025 401.420105892992
0.794174016420566 373.535188741435
0.809005613902106 347.404155810323
0.823837211383647 321.646341726929
0.838668808865188 295.690195150466
0.853500406346729 269.79491035068
0.868332003828269 244.655547751896
0.88316360130981 220.940295912857
0.897995198791351 199.088519742951
0.912826796272892 179.406958273843
0.927658393754432 162.20673147391
0.942489991235973 147.722783454185
0.957321588717513 135.847577530427
0.972153186199054 125.966685356174
0.986984783680595 117.118901808301
1.00181638116214 108.398319615632
1.01664797864368 99.3067968237276
1.03147957612522 89.8452127960497
1.04631117360676 80.3644630266224
1.0611427710883 71.3330476862182
1.07597436856984 63.1462641767183
1.09080596605138 56.0179278955238
1.10563756353292 49.9558454345044
1.12046916101446 44.8072209524567
1.135300758496 40.3342233751928
1.15013235597754 36.273620759631
1.16496395345908 32.374946412475
1.17979555094062 28.4500164297633
1.19462714842216 24.4434468412873
1.20945874590371 20.4766055670857
1.22429034338525 16.8067950415739
1.23912194086679 13.7023920108911
1.25395353834833 11.3041719020451
1.26878513582987 9.5589662620056
1.28361673331141 8.2647934423939
1.29844833079295 7.19299506113003
1.31327992827449 6.20282738947739
1.32811152575603 5.27474828993779
1.34294312323757 4.455708386235
1.35777472071911 3.77841748059815
1.37260631820065 3.2265040415434
1.38743791568219 2.76381143244008
1.40226951316373 2.38426936047183
1.41710111064528 2.12580496202463
1.43193270812682 2.031893192945
1.44676430560836 2.09457852471433
1.4615959030899 2.22944436410194
1.47642750057144 2.3067716400507
1.49125909805298 2.21874152864454
};
\addplot [thick, opacity=0.75, steelblue52138189, forget plot]
table {%
-1.46021542604527 984.613751740473
-1.44538494515318 1104.6413116735
-1.43055446426109 1222.31582265608
-1.415723983369 1335.10263313506
-1.40089350247691 1440.78811604994
-1.38606302158482 1537.60536255217
-1.37123254069273 1624.31246468241
-1.35640205980064 1700.21869381183
-1.34157157890855 1765.1611083286
-1.32674109801646 1819.44027754156
-1.31191061712438 1863.72792593831
-1.29708013623229 1898.96091227822
-1.2822496553402 1926.23513279317
-1.26741917444811 1946.71020946729
-1.25258869355602 1961.53201018019
-1.23775821266393 1971.77603202247
-1.22292773177184 1978.41121456309
-1.20809725087975 1982.28131983195
-1.19326676998766 1984.09978496806
-1.17843628909557 1984.45380059483
-1.16360580820348 1983.81397822734
-1.14877532731139 1982.54695285617
-1.13394484641931 1980.92926517256
-1.11911436552722 1979.16163380803
-1.10428388463513 1977.38315418261
-1.08945340374304 1975.68506985673
-1.07462292285095 1974.12366725941
-1.05979244195886 1972.73169450736
-1.04496196106677 1971.52763512676
-1.03013148017468 1970.52226325739
-1.01530099928259 1969.72218733738
-1.0004705183905 1969.13050990661
-0.985640037498414 1968.74520237886
-0.970809556606325 1968.55620717664
-0.955979075714236 1968.54253647502
-0.941148594822146 1968.67066988357
-0.926318113930057 1968.89534086643
-0.911487633037968 1969.16336926602
-0.896657152145879 1969.42061068264
-0.88182667125379 1969.62144437381
-0.866996190361701 1969.73961127355
-0.852165709469612 1969.77873944399
-0.837335228577522 1969.78063696395
-0.822504747685433 1969.82945036747
-0.807674266793344 1970.05011049429
-0.792843785901255 1970.6001133118
-0.778013305009166 1971.6545667513
-0.763182824117077 1973.38548564568
-0.748352343224987 1975.93739870537
-0.733521862332898 1979.40227025395
-0.718691381440809 1983.7973466717
-0.70386090054872 1989.04964474025
-0.689030419656631 1994.99029911433
-0.674199938764542 2001.36087035641
-0.659369457872453 2007.83209540902
-0.644538976980363 2014.03366877867
-0.629708496088274 2019.59179041694
-0.614878015196185 2024.16974900293
-0.600047534304096 2027.5060280005
-0.585217053412007 2029.44451632452
-0.570386572519918 2029.9524061568
-0.555556091627829 2029.12312330962
-0.540725610735739 2027.16386516284
-0.52589512984365 2024.36962809499
-0.511064648951561 2021.08758505804
-0.496234168059472 2017.67698435349
-0.481403687167383 2014.47017383801
-0.466573206275294 2011.73986981605
-0.451742725383204 2009.67651494808
-0.436912244491115 2008.37776912861
-0.422081763599026 2007.85019474304
-0.407251282706937 2008.02138593804
-0.392420801814848 2008.75944691905
-0.377590320922759 2009.89603564901
-0.36275984003067 2011.24921066223
-0.347929359138581 2012.64296907913
-0.333098878246491 2013.9214529595
-0.318268397354402 2014.95707423452
-0.303437916462313 2015.65300193758
-0.288607435570224 2015.94135059324
-0.273776954678135 2015.77887316675
-0.258946473786046 2015.14197041563
-0.244115992893956 2014.02245698782
-0.229285512001867 2012.42492255032
-0.214455031109778 2010.36587217799
-0.199624550217689 2007.87428449148
-0.1847940693256 2004.99289270817
-0.169963588433511 2001.77940233983
-0.155133107541422 1998.30696966167
-0.140302626649332 1994.66349274033
-0.125472145757243 1990.94951805809
-0.110641664865154 1987.27477077514
-0.0958111839730649 1983.75344888341
-0.0809807030809757 1980.49849783549
-0.0661502221888866 1977.61514551252
-0.0513197412967974 1975.19406933075
-0.0364892604047082 1973.30470332598
-0.0216587795126193 1971.98934957663
-0.00682829862053014 1971.258877419
0.00800218227155902 1971.09080403972
0.0228326631636482 1971.43039475156
0.0376631440557373 1972.19508496282
0.0524936249478265 1973.28204681864
0.0673241058399157 1974.57818915989
0.0821545867320048 1975.97140521474
0.096985067624094 1977.36158097174
0.111815548516183 1978.66982794089
0.126646029408272 1979.84463212894
0.141476510300361 1980.86408111557
0.156306991192451 1981.73395638577
0.17113747208454 1982.48214147516
0.185967952976629 1983.15037681794
0.200798433868718 1983.78478884596
0.215628914760807 1984.42677175153
0.230459395652896 1985.10569002503
0.245289876544986 1985.83452820629
0.260120357437075 1986.60910659149
0.274950838329164 1987.41089423095
0.289781319221253 1988.21287476448
0.304611800113342 1988.98743859903
0.319442281005431 1989.71494954051
0.33427276189752 1990.39150386234
0.349103242789609 1991.03447862624
0.363933723681699 1991.6847446744
0.378764204573788 1992.40486859282
0.393594685465877 1993.27319913738
0.408425166357966 1994.37436187883
0.423255647250055 1995.78729086319
0.438086128142144 1997.57241824179
0.452916609034234 1999.75993385744
0.467747089926323 2002.34104551202
0.482577570818412 2005.26388051772
0.497408051710501 2008.43508275619
0.51223853260259 2011.72734495611
0.527069013494679 2014.9921914488
0.541899494386769 2018.07644291354
0.556729975278858 2020.84010798721
0.571560456170947 2023.17308929002
0.586390937063036 2025.00814473954
0.601221417955125 2026.3280217726
0.616051898847214 2027.16552125111
0.630882379739304 2027.59632262651
0.645712860631393 2027.72553982197
0.660543341523482 2027.66998857407
0.675373822415571 2027.53885675423
0.69020430330766 2027.41575309066
0.705034784199749 2027.344912548
0.719865265091838 2027.32368860488
0.734695745983928 2027.30247453173
0.749526226876017 2027.19203851151
0.764356707768105 2026.87712926691
0.779187188660195 2026.23429648922
0.794017669552284 2025.15131362801
0.808848150444373 2023.54545837319
0.823678631336462 2021.37818869476
0.838509112228551 2018.66437038028
0.85333959312064 2015.47503917093
0.86817007401273 2011.93357187896
0.883000554904819 2008.20596093603
0.897831035796908 2004.48653167336
0.912661516688997 2000.98085062502
0.927491997581086 1997.88773073147
0.942322478473175 1995.3821681825
0.957152959365265 1993.60079427948
0.971983440257354 1992.63105410645
0.986813921149443 1992.5048906582
1.00164440204153 1993.19726694923
1.01647488293362 1994.62943430196
1.03130536382571 1996.67647440254
1.0461358447178 1999.17831850289
1.06096632560989 2001.95318675298
1.07579680650198 2004.81219858598
1.09062728739407 2007.57378324213
1.10545776828616 2010.07646638691
1.12028824917825 2012.18861965353
1.13511873007033 2013.81382991724
1.14994921096242 2014.89067507493
1.16477969185451 2015.38589677986
1.1796101727466 2015.28026890798
1.19444065363869 2014.54691924733
1.20927113453078 2013.12251968114
1.22410161542287 2010.87264760367
1.23893209631496 2007.55372503341
1.25376257720705 2002.77518040962
1.26859305809914 1995.96668861427
1.28342353899123 1986.35628841045
1.29825401988332 1972.96556638103
1.3130845007754 1954.62764902848
1.32791498166749 1930.03224778476
1.34274546255958 1897.79938757885
1.35757594345167 1856.57985655508
1.37240642434376 1805.17621791711
1.38723690523585 1742.67402151441
1.40206738612794 1668.56938867884
1.41689786702003 1582.87718789043
1.43172834791212 1486.20420073036
1.44655882880421 1379.77432225555
1.4613893096963 1265.39784143237
1.47621979058838 1145.38362227007
1.49105027148047 1022.40055430185
};
\addplot [thick, opacity=0.75, mediumpurple152142213, forget plot]
table {%
-1.5059016547968 4.13471113874536
-1.49103746150509 5.39267452145833
-1.47617326821338 6.80202921109001
-1.46130907492167 8.37660576760996
-1.44644488162995 10.0856736953394
-1.43158068833824 11.8237970893544
-1.41671649504653 13.4437102941745
-1.40185230175482 14.8505718293438
-1.38698810846311 16.1000956752628
-1.3721239151714 17.4230072832435
-1.35725972187969 19.1366957085055
-1.34239552858798 21.4820931739214
-1.32753133529626 24.4857111165989
-1.31266714200455 27.9430685017641
-1.29780294871284 31.5487074948171
-1.28293875542113 35.1099458459206
-1.26807456212942 38.7316460259081
-1.25321036883771 42.8669610010318
-1.238346175546 48.1858119366013
-1.22348198225429 55.3029088542891
-1.20861778896258 64.4969308587104
-1.19375359567086 75.5829106143763
-1.17888940237915 88.0291609710518
-1.16402520908744 101.268841225293
-1.14916101579573 115.034471501801
-1.13429682250402 129.519772443437
-1.11943262921231 145.267393763809
-1.1045684359206 162.852868460855
-1.08970424262889 182.585271990096
-1.07484004933717 204.446831471329
-1.05997585604546 228.3025984584
-1.04511166275375 254.166477852275
-1.03024746946204 282.254359336075
-1.01538327617033 312.780986089453
-1.00051908287862 345.74979022124
-0.985654889586908 381.010162267356
-0.970790696295196 418.550444614263
-0.955926503003485 458.693815517371
-0.941062309711774 501.943376114689
-0.926198116420063 548.624160333864
-0.911333923128352 598.725744065048
-0.89646972983664 652.134807429109
-0.881605536544929 708.997206053957
-0.866741343253218 769.793655760934
-0.851877149961507 835.034981225441
-0.837012956669796 904.900525664958
-0.822148763378085 979.186676765062
-0.807284570086373 1057.59216496565
-0.792420376794662 1140.04255866528
-0.777556183502951 1226.75859667643
-0.76269199021124 1318.04206329101
-0.747827796919529 1414.01679661345
-0.732963603627817 1514.59539802279
-0.718099410336106 1619.7035211407
-0.703235217044395 1729.49209976259
-0.688371023752684 1844.2482083472
-0.673506830460973 1964.07054663031
-0.658642637169261 2088.71488061423
-0.64377844387755 2217.86991889295
-0.628914250585839 2351.58469238156
-0.614050057294128 2490.25370003829
-0.599185864002417 2633.91070196135
-0.584321670710706 2781.32832590375
-0.569457477418994 2929.85073547447
-0.554593284127283 3076.4769274498
-0.539729090835572 3219.6527319965
-0.524864897543861 3360.40201750569
-0.51000070425215 3501.69761643031
-0.495136510960438 3646.29843156575
-0.480272317668727 3794.59815820629
-0.465408124377016 3944.1816384116
-0.450543931085305 4091.56194215226
-0.435679737793594 4234.91220322333
-0.420815544501882 4375.815746009
-0.405951351210171 4518.74278146312
-0.39108715791846 4668.53330455618
-0.376222964626749 4827.3887812955
-0.361358771335038 4992.98691408534
-0.346494578043327 5158.57246761821
-0.331630384751615 5314.94248649875
-0.316766191459904 5453.53809861024
-0.301901998168193 5569.43528980583
-0.287037804876482 5662.99732970766
-0.272173611584771 5739.46974746487
-0.257309418293059 5806.72200956492
-0.242445225001348 5872.17094781923
-0.227581031709637 5940.1958365762
-0.212716838417926 6010.99304519259
-0.197852645126215 6081.11705930284
-0.182988451834504 6145.31528995564
-0.168124258542792 6198.88294855083
-0.153260065251081 6239.59436798619
-0.13839587195937 6268.31685489257
-0.123531678667659 6287.90087919431
-0.108667485375948 6300.89443379624
-0.0938032920842364 6307.47273103106
-0.0789390987925251 6304.93308835688
-0.0640749055008139 6289.10021457599
-0.0492107122091028 6256.79326490526
-0.0343465189173915 6207.94036993835
-0.0194823256256804 6146.17775628566
-0.00461813233396913 6077.55426139308
0.0102460609577419 6007.91375126199
0.0251102542494532 5940.24900174138
0.0399744475411643 5873.39408962854
0.0548386408328756 5802.7285447059
0.0697028341245867 5722.48989210599
0.084567027416298 5628.44768914957
0.099431220708009 5519.52042976223
0.11429541399972 5397.53500955589
0.129159607291431 5265.48861844721
0.144023800583143 5125.67236222253
0.158887993874854 4979.00646733724
0.173752187166565 4825.7903495138
0.188616380458276 4666.82023435461
0.203480573749987 4503.72651416321
0.218344767041699 4338.4515598924
0.23320896033341 4172.74352087553
0.248073153625121 4008.33402998893
0.262937346916832 3847.47091887006
0.277801540208543 3692.90437852963
0.292665733500254 3546.89074706831
0.307529926791966 3409.75425056999
0.322394120083677 3279.08416651845
0.337258313375388 3150.31180354205
0.352122506667099 3018.53144362927
0.366986699958811 2880.661730451
0.381850893250522 2736.86435788574
0.396715086542233 2590.57355574214
0.411579279833944 2447.19219664167
0.426443473125655 2312.05822969686
0.441307666417367 2188.48094512018
0.456171859709078 2076.53912573732
0.471036053000789 1973.04324001867
0.4859002462925 1872.68220307376
0.500764439584211 1769.98750017653
0.515628632875922 1661.4398808301
0.530492826167634 1546.8918525615
0.545357019459345 1429.60672425699
0.560221212751056 1314.76847068602
0.575085406042767 1207.18053478713
0.589949599334478 1109.46583178293
0.60481379262619 1021.77786033977
0.6196779859179 942.928587582285
0.634542179209612 871.847260120275
0.649406372501323 808.245732784207
0.664270565793034 752.192795153424
0.679134759084746 703.178657727174
0.693998952376456 659.518748083971
0.708863145668168 618.571475071494
0.723727338959879 577.621484996702
0.73859153225159 534.848834699548
0.753455725543301 489.820945866108
0.768319918835012 443.341749767974
0.783184112126724 396.915705445184
0.798048305418435 352.203730019203
0.812912498710146 310.671200065973
0.827776692001857 273.418371377585
0.842640885293569 241.116441000468
0.85750507858528 213.999272780202
0.872369271876991 191.878165755471
0.887233465168702 174.151689984028
0.902097658460413 159.821722488006
0.916961851752125 147.585949998236
0.931826045043836 136.077797755229
0.946690238335547 124.224614129731
0.961554431627258 111.565302522739
0.976418624918969 98.34221998696
0.991282818210681 85.3068934139488
1.00614701150239 73.3517830252375
1.0210112047941 63.1564297720036
1.03587539808581 54.9813544567674
1.05073959137753 48.6416328387081
1.06560378466924 43.628472564847
1.08046797796095 39.3237677236207
1.09533217125266 35.2309869785738
1.11019636454437 31.1267624976684
1.12506055783608 27.0642242218253
1.13992475112779 23.2439549454073
1.1547889444195 19.8522706534588
1.16965313771121 16.972993008695
1.18451733100293 14.6009512807416
1.19938152429464 12.698331326521
1.21424571758635 11.2192871697605
1.22910991087806 10.0883950356184
1.24397410416977 9.18190216051355
1.25883829746148 8.35728318142134
1.27370249075319 7.51556312435866
1.2885666840449 6.63773770260174
1.30343087733662 5.76215580678014
1.31829507062833 4.9305613035544
1.33315926392004 4.15574615657338
1.34802345721175 3.43234595311109
1.36288765050346 2.76753424753162
1.37775184379517 2.1958293081289
1.39261603708688 1.76328080463371
1.40748023037859 1.49381854847189
1.4223444236703 1.36371412445382
1.43720861696202 1.30432208023015
1.45207281025373 1.23374458014562
};
\addplot [thick, opacity=0.75, gray119, forget plot]
table {%
-1.458784483609 87.2173120548539
-1.44398882841744 111.235374288835
-1.42919317322588 138.494265502427
-1.41439751803431 168.693982033945
-1.39960186284275 201.462710087665
-1.38480620765119 236.402679500602
-1.37001055245962 273.126565155403
-1.35521489726806 311.279357851595
-1.3404192420765 350.546794625122
-1.32562358688494 390.655795458514
-1.31082793169337 431.373261450163
-1.29603227650181 472.507409154472
-1.28123662131025 513.91247020869
-1.26644096611869 555.495176475845
-1.25164531092712 597.220882686114
-1.23684965573556 639.117750203054
-1.222054000544 681.277647575914
-1.20725834535243 723.851465154606
-1.19246269016087 767.035064426269
-1.17766703496931 811.041942201495
-1.16287137977775 856.061540072735
-1.14807572458618 902.207973577655
-1.13328006939462 949.470663576382
-1.11848441420306 997.682541109761
-1.1036887590115 1046.5202737324
-1.08889310381993 1095.5435348802
-1.07409744862837 1144.26863641587
-1.05930179343681 1192.25987182234
-1.04450613824524 1239.21420113289
-1.02971048305368 1285.01472877417
-1.01491482786212 1329.73650457105
-1.00011917267056 1373.60228340576
-0.985323517478994 1416.90147451495
-0.970527862287431 1459.89726917916
-0.955732207095869 1502.75062247935
-0.940936551904306 1545.48384958556
-0.926140896712743 1587.9929607058
-0.91134524152118 1630.10127028951
-0.896549586329618 1671.63327323415
-0.881753931138055 1712.48226215846
-0.866958275946492 1752.64962040184
-0.85216262075493 1792.24617806157
-0.837366965563367 1831.46120460863
-0.822571310371804 1870.51614326811
-0.807775655180241 1909.62334263961
-0.792979999988679 1948.96392885157
-0.778184344797116 1988.68701246393
-0.763388689605553 2028.92075384876
-0.74859303441399 2069.78023073208
-0.733797379222428 2111.36017846856
-0.719001724030865 2153.71063903581
-0.704206068839302 2196.80501450786
-0.68941041364774 2240.51694542134
-0.674614758456177 2284.62120744873
-0.659819103264614 2328.82500828888
-0.645023448073051 2372.82395911014
-0.630227792881489 2416.36722465863
-0.615432137689926 2459.31297419611
-0.600636482498363 2501.6588844408
-0.585840827306801 2543.54047561647
-0.571045172115238 2585.19860527357
-0.556249516923675 2626.92370787703
-0.541453861732112 2668.98788734676
-0.52665820654055 2711.57798423215
-0.511862551348987 2754.74430545705
-0.497066896157424 2798.3800914641
-0.482271240965861 2842.24355421723
-0.467475585774299 2886.02521933426
-0.452679930582736 2929.44887537908
-0.437884275391173 2972.37917975199
-0.423088620199611 3014.90001673902
-0.408292965008048 3057.33122765042
-0.393497309816485 3100.16873685565
-0.378701654624922 3143.95924016949
-0.36390599943336 3189.14542553389
-0.349110344241797 3235.93054451632
-0.334314689050234 3284.20591910608
-0.319519033858672 3333.56310084339
-0.304723378667109 3383.38249439872
-0.289927723475546 3432.96430071068
-0.275132068283983 3481.65572436424
-0.260336413092421 3528.93442340608
-0.245540757900858 3574.42844678891
-0.230745102709295 3617.8781148368
-0.215949447517733 3659.06496738567
-0.20115379232617 3697.73998151545
-0.186358137134607 3733.57686376908
-0.171562481943044 3766.1613363985
-0.156766826751482 3795.01183591574
-0.141971171559919 3819.61765468551
-0.127175516368356 3839.47985316077
-0.112379861176793 3854.1460874638
-0.0975842059852308 3863.23787066983
-0.0827885507936681 3866.47312444586
-0.0679928956021052 3863.68649594111
-0.0531972404105425 3854.84654346981
-0.0384015852189799 3840.06603523539
-0.0236059300274172 3819.60191299057
-0.0088102748358545 3793.84489975538
0.00598538035570839 3763.30264442227
0.0207810355472711 3728.58147135842
0.0355766907388337 3690.36892562288
0.0503723459303964 3649.41420289358
0.0651680011219593 3606.50017587171
0.079963656313522 3562.40242191359
0.0947593115050847 3517.83757360041
0.109554966696647 3473.41184305048
0.12435062188821 3429.58535790981
0.139146277079773 3386.66527983582
0.153941932271336 3344.83097553676
0.168737587462898 3304.18223766067
0.183533242654461 3264.79261339472
0.198328897846024 3226.748302363
0.213124553037586 3190.15901226697
0.227920208229149 3155.13744846991
0.242715863420712 3121.7544654641
0.257511518612275 3089.98410904507
0.272307173803837 3059.6555677016
0.2871028289954 3030.42757552228
0.301898484186963 3001.79566005938
0.316694139378525 2973.13455047991
0.331489794570088 2943.76847534582
0.346285449761651 2913.05360740489
0.361081104953214 2880.45288715799
0.375876760144776 2845.58624806323
0.390672415336339 2808.24843112571
0.405468070527902 2768.39832954481
0.420263725719465 2726.13261170819
0.435059380911027 2681.65799615444
0.44985503610259 2635.27068361954
0.464650691294153 2587.34209558735
0.479446346485715 2538.30324032171
0.494242001677278 2488.62005913254
0.509037656868841 2438.75889888281
0.523833312060404 2389.15026520727
0.538628967251966 2340.1640309843
0.553424622443529 2292.10632450401
0.568220277635092 2245.23820985785
0.583015932826654 2199.80429222359
0.597811588018217 2156.05234471535
0.61260724320978 2114.2274328943
0.627402898401342 2074.53539538362
0.642198553592906 2037.08575309719
0.656994208784468 2001.83586559129
0.671789863976031 1968.56067855722
0.686585519167594 1936.86433412735
0.701381174359156 1906.23485391539
0.716176829550719 1876.12780704342
0.730972484742282 1846.05590219503
0.745768139933844 1815.66200126889
0.760563795125407 1784.76161410022
0.77535945031697 1753.35234215796
0.790155105508533 1721.59620206393
0.804950760700095 1689.78306048843
0.819746415891658 1658.28050394142
0.834542071083221 1627.47153771886
0.849337726274783 1597.68110195613
0.864133381466346 1569.09735450072
0.878929036657909 1541.70177032335
0.893724691849471 1515.22816800494
0.908520347041035 1489.169633077
0.923316002232597 1462.84192061751
0.93811165742416 1435.49489276825
0.952907312615723 1406.44630011536
0.967702967807285 1375.20218299279
0.982498622998848 1341.53037669874
0.997294278190411 1305.46820875613
1.01208993338197 1267.26738985678
1.02688558857354 1227.29994441724
1.0416812437651 1185.96061039858
1.05647689895666 1143.59895776064
1.07127255414822 1100.4996671868
1.08606820933979 1056.90825944658
1.10086386453135 1013.08079095307
1.11565951972291 969.327098870617
1.13045517491448 926.021270164438
1.14525083010604 883.567902402457
1.1600464852976 842.33195272514
1.17484214048916 802.555641550375
1.18963779568073 764.291923470927
1.20443345087229 727.378576659713
1.21922910606385 691.462820929157
1.23402476125541 656.069622219608
1.24882041644698 620.694017574427
1.26361607163854 584.893123066979
1.2784117268301 548.357361203938
1.29320738202166 510.949819968298
1.30800303721323 472.712975905939
1.32279869240479 433.849513786683
1.33759434759635 394.687134039046
1.35239000278792 355.636779455221
1.36718565797948 317.151470825499
1.38198131317104 279.690511062567
1.3967769683626 243.691817004074
1.41157262355417 209.55336921993
1.42636827874573 177.622909598537
1.44116393393729 148.193205046234
1.45595958912886 121.499089786203
1.47075524432042 97.7128042810283
1.48555089951198 76.9360255900918
};
\end{axis}

\end{tikzpicture}

%% file: CIFAR10_paper_bar.tex
\begin{tikzpicture}

\definecolor{chocolate2267451}{RGB}{226,74,51}
\definecolor{darkgray155}{RGB}{155,155,155}
\definecolor{darkorange2551094}{RGB}{255,109,4}
\definecolor{dimgray85}{RGB}{85,85,85}
\definecolor{gainsboro229}{RGB}{239,239,239}
\definecolor{gainsboro247}{RGB}{253,253,253}
\definecolor{gray119}{RGB}{119,119,119}
\definecolor{lightgray204}{RGB}{204,204,204}
\definecolor{mediumpurple152142213}{RGB}{152,142,213}
\definecolor{sandybrown25119394}{RGB}{251,193,94}
\definecolor{steelblue52138189}{RGB}{52,138,189}
\definecolor{yellowgreen14218666}{RGB}{142,186,66}
\definecolor{lightpink255181184}{RGB}{255,181,184}
\definecolor{elfsrgb}{RGB}{130,51,51}

\begin{axis}[
title style={at={(0.5,0.075)},anchor=north},
x label style={at={(axis description cs:1.525, 0.02)}},
y label style={at={(axis description cs:0.135, 0.5)}},
width=6.75cm,
height=6cm,
axis line style={gainsboro229},
legend cell align={left},
legend columns = 6,
legend style={
  fill opacity=0.8,
  draw opacity=1,
  text opacity=1,
  at={(-0.001,1.02)},
  anchor=south west,
  draw=lightgray204,
  fill=gainsboro247,
  column sep=1.24pt,
},
log basis x={10},
tick align=outside,
tick pos=left,
title={\bf  \textcolor{gray}{CIFAR10}},
x dir=reverse,
y grid style={gainsboro229},
x grid style={gainsboro229},
axis background/.style={fill=gainsboro247},
xlabel=\textcolor{dimgray85}{$\xlabel{\text{\footnotesize Percent of Dataset Selected for Coreset}}{\text{~/ Number of Training Examples in Coreset ($n$)}}$},
xmajorgrids,
xmin=4535.48663347337, xmax=38576.896844696,
xmode=log,
xtick style={color=dimgray85},
ylabel=\textcolor{dimgray85}{\footnotesize Accuracy (\%)},
ymajorgrids,
ymin=30, ymax=97,
ytick style={color=dimgray85},
ytick={90, 80, 70, 60, 50, 40, 30},
xtick={5000,10000,15000,25000,35000},
xticklabels={$\meanrelrand{\scriptstyle 10\%}{\scriptscriptstyle 5\textrm{K}}$,$\meanrelrand{\scriptstyle 20\%}{\scriptscriptstyle 10\textrm{K}}$,$\meanrelrand{\scriptstyle 30\%}{\scriptscriptstyle 15\textrm{K}}$,$\meanrelrand{\scriptstyle 50\%}{\scriptscriptstyle 25\textrm{K}}$,$\meanrelrand{\scriptstyle 70\%}{\scriptscriptstyle 35\textrm{K}}$},
ticklabel style={font=\tiny},
xtick style={draw=none},
ytick style={draw=none},
tick align=inside,
yticklabel style = {xshift=0.5ex},
xticklabel style = {yshift=0.5ex},
]
\path [fill=chocolate2267451, fill opacity=0.2, very thin]
(axis cs:4999,33.4)
--(axis cs:4999,30.38)
--(axis cs:9999,64.94)
--(axis cs:15000,84.98)
--(axis cs:25000,95.02)
--(axis cs:35000,95.33)
--(axis cs:35000,95.53)
--(axis cs:35000,95.53)
--(axis cs:25000,95.1)
--(axis cs:15000,88.4)
--(axis cs:9999,72.34)
--(axis cs:4999,33.4)
--cycle;

\path [fill=steelblue52138189, fill opacity=0.2, very thin]
(axis cs:4999,34.85)
--(axis cs:4999,34.63)
--(axis cs:9999,66.49)
--(axis cs:15000,90.56)
--(axis cs:25000,95.1)
--(axis cs:35000,95.35)
--(axis cs:35000,95.53)
--(axis cs:35000,95.53)
--(axis cs:25000,95.28)
--(axis cs:15000,91.82)
--(axis cs:9999,72.71)
--(axis cs:4999,34.85)
--cycle;

\path [fill=mediumpurple152142213, fill opacity=0.2, very thin]
(axis cs:4999,38.24)
--(axis cs:4999,36)
--(axis cs:9999,78.67)
--(axis cs:15000,89.27)
--(axis cs:25000,93.89)
--(axis cs:35000,95.06)
--(axis cs:35000,95.1)
--(axis cs:35000,95.1)
--(axis cs:25000,94.17)
--(axis cs:15000,89.53)
--(axis cs:9999,80.85)
--(axis cs:4999,38.24)
--cycle;

\path [fill=gray119, fill opacity=0.2, very thin]
(axis cs:4999,47.74)
--(axis cs:4999,44)
--(axis cs:9999,73)
--(axis cs:15000,87.1)
--(axis cs:25000,95)
--(axis cs:35000,95.21)
--(axis cs:35000,95.69)
--(axis cs:35000,95.69)
--(axis cs:25000,95.1)
--(axis cs:15000,91.18)
--(axis cs:9999,79.36)
--(axis cs:4999,47.74)
--cycle;

\path [fill=sandybrown25119394, fill opacity=0.2, very thin]
(axis cs:4999,72.87)
--(axis cs:4999,71.25)
--(axis cs:9999,83.28)
--(axis cs:15000,85.98)
--(axis cs:25000,91.74)
--(axis cs:35000,94.38)
--(axis cs:35000,94.52)
--(axis cs:35000,94.52)
--(axis cs:25000,92.06)
--(axis cs:15000,86.5)
--(axis cs:9999,83.7)
--(axis cs:4999,72.87)
--cycle;

\path [fill=yellowgreen14218666, fill opacity=0.2, very thin]
(axis cs:4999,83.85)
--(axis cs:4999,83.37)
--(axis cs:9999,87.48)
--(axis cs:15000,89.57)
--(axis cs:25000,92.25)
--(axis cs:35000,93.9)
--(axis cs:35000,94.02)
--(axis cs:35000,94.02)
--(axis cs:25000,92.43)
--(axis cs:15000,89.85)
--(axis cs:9999,88.02)
--(axis cs:4999,83.85)
--cycle;

\path [fill=elfsrgb, fill opacity=0.2, very thin]
(axis cs:4999,82.739 )
--(axis cs:4999,82.041)
--(axis cs:9999,87.663  )
--(axis cs:15000,90.301 )
--(axis cs:25000,92.349   )
--(axis cs:35000,93.899)
--(axis cs:35000,94.181)
--(axis cs:35000,94.181)
--(axis cs:25000,92.871)
--(axis cs:15000,90.439)
--(axis cs:9999,88.417 )
--(axis cs:4999,82.739)
--cycle;

\path [fill=darkgray155, fill opacity=0.2, very thin]
(axis cs:4999,84.03)
--(axis cs:4999,83.51)
--(axis cs:9999,88.4)
--(axis cs:15000,90.17)
--(axis cs:25000,93.21)
--(axis cs:35000,94.54)
--(axis cs:35000,94.62)
--(axis cs:35000,94.62)
--(axis cs:25000,93.55)
--(axis cs:15000,91.05)
--(axis cs:9999,89.34)
--(axis cs:4999,84.03)
--cycle;

\path [fill=darkorange2551094, fill opacity=0.2, very thin]
(axis cs:4999,84.39)
--(axis cs:4999,83.97)
--(axis cs:9999,88.73)
--(axis cs:15000,90.8)
--(axis cs:25000,93.3)
--(axis cs:35000,94.49)
--(axis cs:35000,94.67)
--(axis cs:35000,94.67)
--(axis cs:25000,93.62)
--(axis cs:15000,91.14)
--(axis cs:9999,89.39)
--(axis cs:4999,84.39)
--cycle;

\path [fill=black, fill opacity=0.2, very thin]
(axis cs:4999,85.67)
--(axis cs:4999,85.25)
--(axis cs:9999,91.05)
--(axis cs:15000,92.78)
--(axis cs:25000,95.17)
--(axis cs:35000,95.41)
--(axis cs:35000,95.53)
--(axis cs:35000,95.53)
--(axis cs:25000,95.25)
--(axis cs:15000,93.28)
--(axis cs:9999,91.55)
--(axis cs:4999,85.67)
--cycle;

\addplot [thick, chocolate2267451, mark=triangle*, mark size=1.25, mark options={solid}, opacity=0.4, fill opacity=0.8]
table {%
5000 31.89
10000 68.64
15000 86.69
25000 95.06
35000 95.43
};
\addlegendentry{\tiny EL2N \text{~\color{gray}NeurIPS `21~\cite{Paul_NEURIPS2021}}}
\addplot [thick, steelblue52138189, mark=square*, mark size=1.25, mark options={solid}, opacity=0.4, fill opacity=0.8]
table {%
5000 34.74
10000 69.6
15000 91.19
25000 95.19
35000 95.44
};
\addlegendentry{\tiny  AUM \text{~\color{gray}NeurIPS `20~\cite{Pleiss_NEURIPS2020}}}
\addplot [thick, mediumpurple152142213, mark=diamond*, mark size=1.25, mark options={solid}, opacity=0.4, fill opacity=0.8]
table {%
5000 37.12
10000 79.76
15000 89.4
25000 94.03
35000 95.08
};
\addlegendentry{\tiny  Dyn-Unc \text{~\color{gray}CVPR WS `24~\cite{He_2024_CVPR}}}
\addplot [thick, gray119, mark=pentagon*, mark size=1.25, mark options={solid}, opacity=0.4, fill opacity=0.8]
table {%
5000 45.87
10000 76.18
15000 89.14
25000 95.05
35000 95.45
};
\addlegendentry{\tiny  Forgetting \text{~\color{gray}ICLR `19~\cite{toneva2018an}}}
\addplot [thick, sandybrown25119394, mark=halfsquare*, mark size=1.25, mark options={solid}, opacity=0.4, fill opacity=0.8]
table {%
5000 72.06
10000 83.49
15000 86.24
25000 91.9
35000 94.45
};
\addlegendentry{\tiny  Entropy \text{~\color{gray}ICLR `20~\cite{Coleman2020Selection}}}
\addplot [thick, yellowgreen14218666, mark=halfdiamond*, mark size=1.25, mark options={solid}, opacity=0.4, fill opacity=0.8]
table {%
5000 83.61
10000 87.75
15000 89.71
25000 92.34
35000 93.96
};
\addlegendentry{\tiny  Moderate\text{~\color{gray}ICLR `23~\cite{xia2023moderate}}}
\addplot [thick, black, mark=halfsquare right*, mark size=1.25, mark options={solid}, opacity=0.4, fill opacity=0.8]
table {%
5000 85.46
10000 91.3
15000 93.03
25000 95.21
35000 95.47
};
\addlegendentry{\tiny  TDDS\text{~\color{gray}CVPR `24~\cite{Zhang_2024_CVPR} }}
\addplot [thick, lightpink255181184, mark=halfsquare left*, mark size=1.25, mark options={solid}, opacity=0.6, fill opacity=0.8]
table {%
  384350 53.59
  256233 42.25
  128116 22.41
};
\addlegendentry{\tiny Prototypes$_\text{sup.}$\text{~\color{gray}NeurIPS~\cite{Sorscher_NEURIPS2022}}}
\addplot [thick, lightpink255181184, dotted, mark=halfsquare left*, mark size=1.25, mark options={solid}, opacity=0.6, fill opacity=0.8]
table {%
  384350 60.42
  256233 53.73
  128116 38.06
};
\addlegendentry{\tiny Prototypes$_\text{\tiny ss}$\text{~\color{gray}NeurIPS `22~\cite{Sorscher_NEURIPS2022}}}
\addplot [thick, elfsrgb, dotted, mark=triangle*, mark size=1.25, mark options={solid, rotate=180}, opacity=0.4, fill opacity=0.8]
table{%
5000 82.39
10000 88.04  
15000 90.37
25000 92.61
35000 94.04
};
\addlegendentry{\tiny ELFS\text{~\color{gray}ICLR `25~\cite{zheng2025elfs}}}
\addplot [thick, darkgray155, dashed, mark=halfcircle*, mark size=1.25, mark options={solid}, opacity=0.4, fill opacity=0.8]
table {%
5000 83.77
10000 88.87
15000 90.61
25000 93.38
35000 94.58
};
\addlegendentry{\tiny  Random}
\addplot [thick, darkorange2551094, dashed, mark=*, mark size=1.25, mark options={solid}, opacity=0.4, fill opacity=0.8]
table {%
5000 84.18
10000 89.06
15000 90.97
25000 93.46
35000 94.58
};
\addlegendentry{\tiny  ZCore \text{ \color{gray} (ours)}}
\end{axis}

\end{tikzpicture}

%% file: CIFAR100_paper_bar.tex
\begin{tikzpicture}

\definecolor{chocolate2267451}{RGB}{226,74,51}
\definecolor{darkgray155}{RGB}{155,155,155}
\definecolor{darkorange2551094}{RGB}{255,109,4}
\definecolor{dimgray85}{RGB}{85,85,85}
\definecolor{gainsboro229}{RGB}{239,239,239}
\definecolor{gainsboro247}{RGB}{253,253,253}
\definecolor{gray119}{RGB}{119,119,119}
\definecolor{lightgray204}{RGB}{204,204,204}
\definecolor{mediumpurple152142213}{RGB}{152,142,213}
\definecolor{sandybrown25119394}{RGB}{251,193,94}
\definecolor{steelblue52138189}{RGB}{52,138,189}
\definecolor{yellowgreen14218666}{RGB}{142,186,66}
\definecolor{elfsrgb}{RGB}{130,51,51}

\begin{axis}[
title style={at={(0.5,0.075)},anchor=north},
ytick={70, 60, 50, 40, 30, 20, 10},
width=6.75cm,
height=6cm,
axis background/.style={fill=gainsboro247},
axis line style={white},
legend cell align={left},
legend cell align={left},
legend style={
	fill opacity=0,
	draw opacity=0,
	text opacity=1,
	at={(0,1.02)},
	anchor=south west,
	draw=lightgray204,
	fill=gainsboro247,
  text opacity=0,
},
log basis x={10},
tick align=outside,
tick pos=left,
title={\bf  \textcolor{gray}{CIFAR100}},
x dir=reverse,
x grid style={gainsboro229},
y grid style={gainsboro229},
xmajorgrids,
xmin=4535.48663347337, xmax=38576.896844696,
xmode=log,
xtick style={color=dimgray85},
ymajorgrids,
ymin=7.5, ymax=79,
ytick style={color=dimgray85},
xtick={5000,10000,15000,25000,35000},
xticklabels={$\meanrelrand{\scriptstyle 10\%}{\scriptscriptstyle 5\textrm{K}}$,$\meanrelrand{\scriptstyle 20\%}{\scriptscriptstyle 10\textrm{K}}$,$\meanrelrand{\scriptstyle 30\%}{\scriptscriptstyle 15\textrm{K}}$,$\meanrelrand{\scriptstyle 50\%}{\scriptscriptstyle 25\textrm{K}}$,$\meanrelrand{\scriptstyle 70\%}{\scriptscriptstyle 35\textrm{K}}$},
y label style={at={(axis description cs:0.1525, 0.5)}},
ylabel=\textcolor{white}{A},
xlabel=\textcolor{white}{.},
x label style={at={(axis description cs:0.45,1.32)}},
ticklabel style={font=\tiny},
xtick style={draw=none},
ytick style={draw=none},
tick align=inside,
axis line style={gainsboro229},
yticklabel style = {xshift=0.5ex},
xticklabel style = {yshift=0.5ex},
]

\path [fill=chocolate2267451, fill opacity=0.2, very thin]
(axis cs:4999,9.79)
--(axis cs:4999,8.41)
--(axis cs:9999,16.98)
--(axis cs:15000,35.09)
--(axis cs:25000,67.42)
--(axis cs:35000,76.58)
--(axis cs:35000,77.2)
--(axis cs:35000,77.2)
--(axis cs:25000,67.72)
--(axis cs:15000,37.81)
--(axis cs:9999,17.64)
--(axis cs:4999,9.79)
--cycle;

\path [fill=steelblue52138189, fill opacity=0.2, very thin]
(axis cs:4999,9.56)
--(axis cs:4999,9.02)
--(axis cs:9999,17.96)
--(axis cs:15000,31.35)
--(axis cs:25000,67.65)
--(axis cs:35000,77.17)
--(axis cs:35000,77.53)
--(axis cs:35000,77.53)
--(axis cs:25000,68.69)
--(axis cs:15000,32.03)
--(axis cs:9999,18.9)
--(axis cs:4999,9.56)
--cycle;

\path [fill=mediumpurple152142213, fill opacity=0.2, very thin]
(axis cs:4999,15.61)
--(axis cs:4999,14.79)
--(axis cs:9999,38.92)
--(axis cs:15000,49.69)
--(axis cs:25000,65.65)
--(axis cs:35000,73.26)
--(axis cs:35000,73.46)
--(axis cs:35000,73.46)
--(axis cs:25000,66.15)
--(axis cs:15000,50.63)
--(axis cs:9999,39.46)
--(axis cs:4999,15.61)
--cycle;

\path [fill=gray119, fill opacity=0.2, very thin]
(axis cs:4999,26.34)
--(axis cs:4999,25.3)
--(axis cs:9999,37.29)
--(axis cs:15000,49.64)
--(axis cs:25000,70.36)
--(axis cs:35000,77.29)
--(axis cs:35000,77.47)
--(axis cs:35000,77.47)
--(axis cs:25000,71.16)
--(axis cs:15000,50.2)
--(axis cs:9999,39.55)
--(axis cs:4999,26.34)
--cycle;

\path [fill=sandybrown25119394, fill opacity=0.2, very thin]
(axis cs:4999,30.1)
--(axis cs:4999,29.02)
--(axis cs:9999,42.61)
--(axis cs:15000,49.87)
--(axis cs:25000,64.08)
--(axis cs:35000,72.19)
--(axis cs:35000,72.59)
--(axis cs:35000,72.59)
--(axis cs:25000,64.8)
--(axis cs:15000,51.59)
--(axis cs:9999,43.11)
--(axis cs:4999,30.1)
--cycle;

\path [fill=yellowgreen14218666, fill opacity=0.2, very thin]
(axis cs:4999,42.94)
--(axis cs:4999,40.7)
--(axis cs:9999,56.15)
--(axis cs:15000,62.73)
--(axis cs:25000,69.98)
--(axis cs:35000,74.5)
--(axis cs:35000,74.7)
--(axis cs:35000,74.7)
--(axis cs:25000,70.6)
--(axis cs:15000,62.89)
--(axis cs:9999,56.89)
--(axis cs:4999,42.94)
--cycle;

\path [fill=elfsrgb, fill opacity=0.2, very thin]
(axis cs:4999,48.741)
--(axis cs:4999,46.099)
--(axis cs:9999,58.306  )
--(axis cs:15000,63.269 )
--(axis cs:25000,70.18   )
--(axis cs:35000,73.949 )
--(axis cs:35000,74.251)
--(axis cs:35000,74.251)
--(axis cs:25000,70.54)
--(axis cs:15000,64.391)
--(axis cs:9999,58.714 )
--(axis cs:4999,48.741)
--cycle;
 
\path [fill=darkgray155, fill opacity=0.2, very thin]
(axis cs:4999,47.75)
--(axis cs:4999,45.61)
--(axis cs:9999,57.55)
--(axis cs:15000,64.27)
--(axis cs:25000,71.79)
--(axis cs:35000,75.49)
--(axis cs:35000,75.57)
--(axis cs:35000,75.57)
--(axis cs:25000,72.11)
--(axis cs:15000,64.91)
--(axis cs:9999,58.03)
--(axis cs:4999,47.75)
--cycle;

\path [fill=darkorange2551094, fill opacity=0.2, very thin]
(axis cs:4999,52.77)
--(axis cs:4999,51.45)
--(axis cs:9999,61.53)
--(axis cs:15000,65.77)
--(axis cs:25000,72.69)
--(axis cs:35000,75.89)
--(axis cs:35000,76.19)
--(axis cs:35000,76.19)
--(axis cs:25000,73.05)
--(axis cs:15000,66.07)
--(axis cs:9999,62.31)
--(axis cs:4999,52.77)
--cycle;

\path [fill=black, fill opacity=0.2, very thin]
(axis cs:4999,54.73)
--(axis cs:4999,54.29)
--(axis cs:9999,62.89)
--(axis cs:15000,67.34)
--(axis cs:25000,73.7)
--(axis cs:35000,77.5)
--(axis cs:35000,77.62)
--(axis cs:35000,77.62)
--(axis cs:25000,74.38)
--(axis cs:15000,68.22)
--(axis cs:9999,63.13)
--(axis cs:4999,54.73)
--cycle;

\addplot [thick, chocolate2267451, mark=triangle*, mark size=1.25, mark options={solid}, opacity=0.4, fill opacity=0.8, forget plot]
table {%
5000 9.1
10000 17.31
15000 36.45
25000 67.57
35000 76.89
};
\addlegendentry{\scriptsize EL2N}
\addplot [thick, steelblue52138189, mark=square*, mark size=1.25, mark options={solid}, opacity=0.4, fill opacity=0.8, forget plot]
table {%
5000 9.29
10000 18.43
15000 31.69
25000 68.17
35000 77.35
};
\addlegendentry{\scriptsize AUM}
\addplot [thick, mediumpurple152142213, mark=diamond*, mark size=1.25, mark options={solid}, opacity=0.4, fill opacity=0.8, forget plot]
table {%
5000 15.2
10000 39.19
15000 50.16
25000 65.9
35000 73.36
};
\addlegendentry{\scriptsize Dyn-Unc}
\addplot [thick, gray119, mark=pentagon*, mark size=1.25, mark options={solid}, opacity=0.4, fill opacity=0.8, forget plot]
table {%
5000 25.82
10000 38.42
15000 49.92
25000 70.76
35000 77.38
};
\addlegendentry{\scriptsize Forgetting}
\addplot [thick, sandybrown25119394, mark=halfsquare*, mark size=1.25, mark options={solid}, opacity=0.4, fill opacity=0.8, forget plot]
table {%
5000 29.56
10000 42.86
15000 50.73
25000 64.44
35000 72.39
};
\addlegendentry{\scriptsize Entropy}
\addplot [thick, yellowgreen14218666, mark=halfdiamond*, mark size=1.25, mark options={solid}, opacity=0.4, fill opacity=0.8, forget plot]
table {%
5000 41.82
10000 56.52
15000 62.81
25000 70.29
35000 74.6
};
\addlegendentry{\scriptsize Moderate}
\addplot [thick, black, mark=halfsquare right*, mark size=1.25, mark options={solid}, opacity=0.4, fill opacity=0.8, forget plot]
table {%
5000 54.51
10000 63.01
15000 67.78
25000 74.04
35000 77.56
};
\addlegendentry{\scriptsize TDDS}
\addplot [thick, elfsrgb, dotted, mark=triangle*, mark size=1.25, mark options={solid, rotate=180}, opacity=0.4, fill opacity=0.8]
table {%
	5000 47.42
	10000 58.51
	15000 63.83
	25000 70.36
	35000 74.10
};
\addlegendentry{\scriptsize ELFS}
\addplot [thick, darkgray155, dashed, mark=halfcircle*, mark size=1.25, mark options={solid}, opacity=0.4, fill opacity=0.8, forget plot]
table {%
5000 46.68
10000 57.79
15000 64.59
25000 71.95
35000 75.53
};
\addlegendentry{\scriptsize Random}
\addplot [thick, darkorange2551094, dashed, mark=*, mark size=1.25, mark options={solid}, opacity=0.4, fill opacity=0.8, forget plot]
table {%
	5000 52.11
	10000 61.92
	15000 65.92
	25000 72.87
	35000 76.04
};
\addlegendentry{\scriptsize ZCore}
\legend{},
\end{axis}

\end{tikzpicture}

%% file: imagenet_paper.tex
\begin{tikzpicture}

\definecolor{chocolate2267451}{RGB}{226,74,51}
\definecolor{darkgray155}{RGB}{155,155,155}
\definecolor{darkorange2551094}{RGB}{255,109,4}
\definecolor{dimgray85}{RGB}{85,85,85}
\definecolor{gainsboro229}{RGB}{239,239,239}
\definecolor{gainsboro247}{RGB}{253,253,253}
\definecolor{gray119}{RGB}{119,119,119}
\definecolor{lightgray204}{RGB}{204,204,204}
\definecolor{mediumpurple152142213}{RGB}{152,142,213}
\definecolor{sandybrown25119394}{RGB}{251,193,94}
\definecolor{steelblue52138189}{RGB}{52,138,189}
\definecolor{yellowgreen14218666}{RGB}{142,186,66}
\definecolor{neonpink}{RGB}{128,8,120}
\definecolor{lightpink255181184}{RGB}{255,181,184}
\definecolor{darkorange-clip}{RGB}{255,54,2}
\definecolor{darkorange-clip}{RGB}{200,175,4}
\definecolor{darkorange-res}{RGB}{200,0,0}


\begin{axis}[
title style={at={(0.5,0.075)},anchor=north},
x label style={at={(axis description cs:0.5, -0.05)}},
width=6.75cm,
height=6cm,
axis line style={white},
legend cell align={left},
legend style={
  fill opacity=0.4,
  draw opacity=0.6,
  text opacity=1,
  at={(0.03,0.03)},
  anchor=south west,
  draw=lightgray204,
  fill=gainsboro247
},
log basis x={10},
tick align=inside,
title={\bf  \textcolor{gray}{ImageNet}},
x dir=reverse,
x grid style={gainsboro229},
y grid style={gainsboro229},
xmajorgrids,
xmin=121268.273641759, xmax=406053.315688034,
xmode=log,
ymajorgrids,
ymin=10, ymax=66,
xtick={128116,256233,384350},
xticklabels={$\meanrelrand{\scriptscriptstyle 10\%}{\scriptscriptstyle 128.1\textrm{K}}$,$\meanrelrand{\scriptscriptstyle 20\%}{\scriptscriptstyle 256.2\textrm{K}}$,$\meanrelrand{\scriptscriptstyle ~30\%}{\scriptscriptstyle ~384.4\textrm{K}}$},
ticklabel style={font=\tiny},
xtick style={draw=none},
ytick style={draw=none},
y label style={at={(axis description cs:0.19, 0.5)}},
ylabel=\textcolor{white}{A},
axis background/.style={fill=gainsboro247},
axis line style={gainsboro229},
xlabel=\textcolor{white}{.},
x label style={at={(axis description cs:0.45,1.32)}},
ytick={60, 50, 40, 30, 20, 10},
yticklabel style = {xshift=0.5ex},
xticklabel style = {yshift=0.5ex},
]
\addplot [very thick, lightpink255181184, dotted, mark=halfsquare left*, mark size=1.25, mark options={solid}, opacity=0.6, fill opacity=0.8, forget plot]
table {%
	384350 60.42
	256233 53.73
	128116 38.06
};
\addplot[very thick, chocolate2267451, mark=triangle*, mark size=1.25, mark options={solid}, opacity=0.4, fill opacity=0.8, forget plot]
table {%
	384350 46.92
	256233 32.68
	128116 15.9
};
\addplot [very thick, steelblue52138189, mark=square*, mark size=1.25, mark options={solid}, opacity=0.4, fill opacity=0.8, forget plot]
table {%
	384350 39.34
	256233 23.64
	128116 11.7
};
\addplot [very thick, lightpink255181184, mark=halfsquare left*, mark size=1.25, mark options={solid}, opacity=0.6, fill opacity=0.8, forget plot]
table {%
	384350 53.59
	256233 42.25
	128116 22.41
};
\addplot [very thick, gray119, mark=pentagon*, mark size=1.25, mark options={solid}, opacity=0.4, fill opacity=0.8, forget plot]
table {%
	384350 64.29
	256233 62.01
	128116 52.14
};
\addplot [very thick, sandybrown25119394, mark=halfsquare*, mark size=1.25, mark options={solid}, opacity=0.4, fill opacity=0.8, forget plot]
table {%
	384350 62.34
	256233 56.8
	128116 43.39
};
\addplot [very thick, yellowgreen14218666, mark=halfdiamond*, mark size=1.25, mark options={solid}, opacity=0.4, fill opacity=0.8, forget plot]
table {%
	384350 64.04
	256233 61.35
	128116 52.45
};
\addplot [very thick, gray119, mark=pentagon*, mark size=1.25, mark options={solid}, opacity=0.4, fill opacity=0.8, forget plot]
table {%
384350 64.29
256233 62.01
128116 52.14
};
\addplot [very thick, black, mark=halfsquare right*, mark size=1.25, mark options={solid}, opacity=0.4, fill opacity=0.8, forget plot]
table {%
384350 64.69
256233 62.56
128116 53.91
};
\addplot [very thick, darkgray155, dashed, mark=halfcircle*, mark size=1.25, mark options={solid}, opacity=0.4, fill opacity=0.8, forget plot]
table {%
384350 64.19
256233 60.76
128116 52.63
};
\addplot [very thick, darkorange2551094, dashed, mark=*, mark size=1.25, mark options={solid}, opacity=0.4, fill opacity=0.8, forget plot]
table {%
384350 64.43
256233 61.31
128116 53.99
};
\end{axis}

\end{tikzpicture}

%% file: EuroSAT80.tex
\begin{tikzpicture}

\definecolor{darkgray155}{RGB}{155,155,155}
\definecolor{darkorange2551094}{RGB}{255,109,4}
\definecolor{dimgray85}{RGB}{85,85,85}
\definecolor{gainsboro229}{RGB}{239,239,239}
\definecolor{gainsboro247}{RGB}{253,253,253}
\definecolor{lightgray204}{RGB}{204,204,204}
\definecolor{steelblue52138189}{RGB}{52,138,189}

\begin{axis}[
title style={at={(0.5,0.06)},anchor=north},
title={\bf \textcolor{gray}{\scriptsize EuroSAT80}},
width=5.33cm,
height=4.0cm,
xtick={2160,4320,6480,10800,15120},
axis background/.style={fill=gainsboro247},
axis line style={white},
legend cell align={left},
legend style={
  fill opacity=0,
  draw opacity=0,
  text opacity=1,
  at={(-0.02,0.06)},
  anchor=south west,
  draw=lightgray204,
  fill=gainsboro247,
  legend columns = 1,
},
log basis x={10},
tick align=outside,
tick pos=left,
x dir=reverse,
x grid style={gainsboro229},
xmajorgrids,
xmin=1958.79560585509, xmax=16664.2813075694,
xmode=log,
xtick style={color=dimgray85},
y grid style={gainsboro229},
ymajorgrids,
ymin=94, ymax=99,
ytick style={color=dimgray85},
ticklabel style={font=\tiny},
xtick style={draw=none},
ytick style={draw=none},
tick align=inside,
ylabel=\textcolor{dimgray85}{\footnotesize Accuracy (\%)},
y label style={at={(axis description cs:0.175, 0.5)}},
axis line style={gainsboro229},
xlabel=\textcolor{dimgray85}{$\xlabel{\text{\footnotesize Percent of Dataset Selected for Coreset}}{\text{~/ Number of Training Examples in Coreset ($n$)}}$},
x label style={at={(axis description cs:2.15, 0.06)}},
xticklabels={$\meanrelrand{\scriptstyle 10\%}{\scriptscriptstyle 2,160}$,$\meanrelrand{\scriptstyle 20\%}{\scriptscriptstyle 4,320}$,$\meanrelrand{\scriptstyle 30\%}{\scriptscriptstyle6,480~}$,$\meanrelrand{\scriptstyle 50\%}{\scriptscriptstyle ~~10.8\textrm{K}}$,$\meanrelrand{\scriptstyle ~~70\%}{\scriptscriptstyle 15.1\textrm{K}~}$},
yticklabel style = {xshift=0.5ex},
xticklabel style = {yshift=0.5ex},
]
\path [fill=darkgray155, fill opacity=0.2, very thin]
(axis cs:2159,95.21)
--(axis cs:2159,94.23)
--(axis cs:4319,96.36)
--(axis cs:6480,96.81)
--(axis cs:10800,97.84)
--(axis cs:15119,98.09)
--(axis cs:15119,98.31)
--(axis cs:15119,98.31)
--(axis cs:10800,98.04)
--(axis cs:6480,97.15)
--(axis cs:4319,96.94)
--(axis cs:2159,95.21)
--cycle;

\path [fill=darkorange2551094, fill opacity=0.2, very thin]
(axis cs:2159,95.98)
--(axis cs:2159,95.62)
--(axis cs:4319,97.14)
--(axis cs:6480,97.59)
--(axis cs:10800,98.02)
--(axis cs:15119,98.24)
--(axis cs:15119,98.4)
--(axis cs:15119,98.4)
--(axis cs:10800,98.28)
--(axis cs:6480,97.85)
--(axis cs:4319,97.48)
--(axis cs:2159,95.98)
--cycle;

\path [fill=black, fill opacity=0.2, very thin]
(axis cs:2159,96.39)
--(axis cs:2159,96.17)
--(axis cs:4319,97.99)
--(axis cs:6480,98.4)
--(axis cs:10800,98.47)
--(axis cs:15119,98.57)
--(axis cs:15119,98.67)
--(axis cs:15119,98.67)
--(axis cs:10800,98.69)
--(axis cs:6480,98.46)
--(axis cs:4319,98.19)
--(axis cs:2159,96.39)
--cycle;
\addplot [thick, black, mark=halfsquare right*, mark size=1.25, mark options={solid}, opacity=0.4, fill opacity=0.8]
table {%
2160 96.28
4320 98.09
6480 98.43
10800 98.58
15120 98.62
};
\addlegendentry{\tiny TDDS}
\addplot [thick, darkgray155, dashed, mark=halfcircle*, mark size=1.25, mark options={solid}, opacity=0.4, fill opacity=0.8]
table {%
2160 94.72
4320 96.65
6480 96.98
10800 97.94
15120 98.2
};
\addlegendentry{\tiny Random}
\addplot [thick, darkorange2551094, dashed, mark=*, mark size=1.25, mark options={solid}, opacity=0.4, fill opacity=0.8]
table {%
2160 95.80
4320 97.31
6480 97.72
10800 98.15
15120 98.32
};
\addlegendentry{\tiny ZCore}
\end{axis}

\end{tikzpicture}

%% file: EuroSAT40.tex
\begin{tikzpicture}

\definecolor{darkgray155}{RGB}{155,155,155}
\definecolor{darkorange2551094}{RGB}{255,109,4}
\definecolor{dimgray85}{RGB}{85,85,85}
\definecolor{gainsboro229}{RGB}{239,239,239}
\definecolor{gainsboro247}{RGB}{253,253,253}
\definecolor{lightgray204}{RGB}{204,204,204}
\definecolor{steelblue52138189}{RGB}{52,138,189}

\begin{axis}[
width=5.33cm,
height=4.0cm,
xtick={1080,2160,3240,5400,7560},
axis background/.style={fill=gainsboro247},
axis line style={white},
legend cell align={left},
legend style={
  fill opacity=0.8,
  draw opacity=1,
  text opacity=1,
  at={(0.03,0.03)},
  anchor=south west,
  draw=lightgray204,
  fill=gainsboro229
},
log basis x={10},
tick align=outside,
tick pos=left,
x dir=reverse,
x grid style={gainsboro229},
xmajorgrids,
xmin=978.924729273433, xmax=8331.75499208563,
xmode=log,
xtick style={color=dimgray85},
y grid style={gainsboro229},
ymajorgrids,
ymin=90, ymax=99,
ytick style={color=dimgray85},
ticklabel style={font=\tiny},
xtick style={draw=none},
ytick style={draw=none},
tick align=inside,
y label style={at={(axis description cs:0.1725, 0.5)}},
ylabel=\textcolor{white}{.},
axis line style={gainsboro229},
x label style={at={(axis description cs:0.5,-0.03)}},
xlabel=\textcolor{white}{.},
title style={at={(0.5,0.06)},anchor=north},
title={\bf \textcolor{gray}{\scriptsize EuroSAT40}},
xticklabels={$\meanrelrand{\scriptstyle 10\%}{\scriptscriptstyle 1,080}$,$\meanrelrand{\scriptstyle 20\%}{\scriptscriptstyle 2,160}$,$\meanrelrand{\scriptstyle 30\%}{\scriptscriptstyle 3,240}$,$\meanrelrand{\scriptstyle 50\%}{\scriptscriptstyle ~~5,400}$,$\meanrelrand{\scriptstyle ~~70\%}{\scriptscriptstyle 7,560~}$},
yticklabel style = {xshift=0.5ex},
xticklabel style = {yshift=0.5ex},
]
\path [fill=darkgray155, fill opacity=0.2, very thin]
(axis cs:7559,97.11)
--(axis cs:7559,96.97)
--(axis cs:5400,96.06)
--(axis cs:3240,94.33)
--(axis cs:2159,93.15)
--(axis cs:1079,90.16)
--(axis cs:1079,91.22)
--(axis cs:1079,91.22)
--(axis cs:2159,94.31)
--(axis cs:3240,95.67)
--(axis cs:5400,96.8)
--(axis cs:7559,97.11)
--cycle;

\path [fill=black, fill opacity=0.2, very thin]
(axis cs:7559,98.06)
--(axis cs:7559,97.88)
--(axis cs:5400,98)
--(axis cs:3240,97.47)
--(axis cs:2159,96.63)
--(axis cs:1079,92.51)
--(axis cs:1079,93.05)
--(axis cs:1079,93.05)
--(axis cs:2159,96.95)
--(axis cs:3240,97.63)
--(axis cs:5400,98.12)
--(axis cs:7559,98.06)
--cycle;

\path [fill=darkorange2551094, fill opacity=0.2, very thin]
(axis cs:7559,97.64)
--(axis cs:7559,97.54)
--(axis cs:5400,97.37)
--(axis cs:3240,96.33)
--(axis cs:2159,95.87)
--(axis cs:1079,92.39)
--(axis cs:1079,93.49)
--(axis cs:1079,93.49)
--(axis cs:2159,96.25)
--(axis cs:3240,96.57)
--(axis cs:5400,97.69)
--(axis cs:7559,97.64)
--cycle;

\addplot [thick, black, mark=halfsquare right*, mark size=1.25, mark options={solid}, opacity=0.4, fill opacity=0.8]
table {%
7560 97.97
5400 98.06
3240 97.55
2160 96.79
1080 92.78
};
\addlegendentry{TDDS}
\addplot [thick, darkgray155, dashed, mark=halfcircle*, mark size=1.25, mark options={solid}, opacity=0.4, fill opacity=0.8]
table {%
7560 97.04
5400 96.43
3240 95
2160 93.73
1080 90.69
};
\addlegendentry{Rand}
\addplot [thick, darkorange2551094, dashed, mark=*, mark size=1.25, mark options={solid}, opacity=0.4, fill opacity=0.8]
table {%
7560 97.59
5400 97.53
3240 96.45
2160 96.06
1080 92.94
};
\addlegendentry{BlindCS}
\legend{},
\end{axis}

\end{tikzpicture}

%% file: EuroSAT20.tex
\begin{tikzpicture}

\definecolor{darkgray155}{RGB}{155,155,155}
\definecolor{darkorange2551094}{RGB}{255,109,4}
\definecolor{dimgray85}{RGB}{85,85,85}
\definecolor{gainsboro229}{RGB}{239,239,239}
\definecolor{gainsboro247}{RGB}{253,253,253}
\definecolor{lightgray204}{RGB}{204,204,204}
\definecolor{steelblue52138189}{RGB}{52,138,189}

\begin{axis}[
width=5.33cm,
height=4.0cm,
y label style={at={(axis description cs:0.0, 0.5)}},
xtick={540,1080,1620,2700,3780},
axis background/.style={fill=gainsboro247},
axis line style={white},
legend cell align={left},
legend style={
	fill opacity=0.4,
	draw opacity=0.6,
  text opacity=1,
  at={(0.03,0.03)},
  anchor=south west,
  draw=lightgray204,
  fill=gainsboro247
},
log basis x={10},
tick align=outside,
tick pos=left,
x dir=reverse,
x grid style={gainsboro229},
xmajorgrids,
xmin=488.989303150116, xmax=4165.49193791811,
xmode=log,
xtick style={color=dimgray85},
y grid style={gainsboro229},
ylabel=\textcolor{white}{A},
ymajorgrids,
ymin=76, ymax=98,
ytick style={color=dimgray85},
ticklabel style={font=\tiny},
xtick style={draw=none},
ytick style={draw=none},
tick align=inside,
y label style={at={(axis description cs:0.195, 0.5)}},
ylabel=\textcolor{white}{.},
axis line style={gainsboro229},
x label style={at={(axis description cs:0.5, -0.03)}},
xlabel=\textcolor{white}{.},
title style={at={(0.5,0.06)},anchor=north},
title={\bf \textcolor{gray}{\scriptsize EuroSAT20}},
xticklabels={$\meanrelrand{\scriptstyle 10\%}{\scriptscriptstyle 540}$,$\meanrelrand{\scriptstyle 20\%}{\scriptscriptstyle 1,080}$,$\meanrelrand{\scriptstyle 30\%}{\scriptscriptstyle 1,620}$,$\meanrelrand{\scriptstyle 50\%}{\scriptscriptstyle ~~2,700}$,$\meanrelrand{\scriptstyle 70\%}{\scriptscriptstyle 3,780~~}$},
yticklabel style = {xshift=0.5ex},
xticklabel style = {yshift=0.5ex},
]
\path [fill=darkgray155, fill opacity=0.2, very thin]
(axis cs:3779,95.46)
--(axis cs:3779,94.82)
--(axis cs:2700,92.66)
--(axis cs:1620,88.82)
--(axis cs:1079,87.79)
--(axis cs:539,82.57)
--(axis cs:539,84.03)
--(axis cs:539,84.03)
--(axis cs:1079,88.23)
--(axis cs:1620,89.9)
--(axis cs:2700,93.84)
--(axis cs:3779,95.46)
--cycle;

\path [fill=darkorange2551094, fill opacity=0.2, very thin]
(axis cs:3779,96.65)
--(axis cs:3779,96.33)
--(axis cs:2700,95.23)
--(axis cs:1620,92.31)
--(axis cs:1079,91.1)
--(axis cs:539,76.48)
--(axis cs:539,84.3)
--(axis cs:539,84.3)
--(axis cs:1079,92.5)
--(axis cs:1620,92.89)
--(axis cs:2700,95.67)
--(axis cs:3779,96.65)
--cycle;

\path [fill=black, fill opacity=0.2, very thin]
(axis cs:3779,97.04)
--(axis cs:3779,96.84)
--(axis cs:2700,96.38)
--(axis cs:1620,93.3)
--(axis cs:1079,94.35)
--(axis cs:539,86.39)
--(axis cs:539,87.49)
--(axis cs:539,87.49)
--(axis cs:1079,95.05)
--(axis cs:1620,94.42)
--(axis cs:2700,96.52)
--(axis cs:3779,97.04)
--cycle;

\addplot [thick, black, mark=halfsquare right*, mark size=1.25, mark options={solid}, opacity=0.4, fill opacity=0.8]
table {%
3780 96.94
2700 96.45
1620 93.86
1080 94.7
540 86.94
};
\addlegendentry{\scriptsize TDDS}
\addplot [thick, darkgray155, dashed, mark=halfcircle*, mark size=1.25, mark options={solid}, opacity=0.4, fill opacity=0.8]
table {%
3780 95.14
2700 93.25
1620 89.36
1080 88.01
540 83.3
};
\addlegendentry{\scriptsize Random}
\addplot [thick, darkorange2551094, dashed, mark=*, mark size=1.25, mark options={solid}, opacity=0.4, fill opacity=0.8]
table {%
3780 96.49
2700 95.45
1620 92.60
1080 91.80
540 80.39
};
\addlegendentry{\scriptsize ZCore}
\legend{}
\end{axis}

\end{tikzpicture}

%% file: EuroSAT10.tex
\begin{tikzpicture}

\definecolor{darkgray155}{RGB}{155,155,155}
\definecolor{darkorange2551094}{RGB}{255,109,4}
\definecolor{dimgray85}{RGB}{85,85,85}
\definecolor{gainsboro229}{RGB}{239,239,239}
\definecolor{gainsboro247}{RGB}{253,253,253}
\definecolor{lightgray204}{RGB}{204,204,204}
\definecolor{steelblue52138189}{RGB}{52,138,189}

\begin{axis}[
width=5.33cm,
height=4.0cm,
xtick={270,540,810,1350,1890},
axis line style={white},
legend cell align={left},
legend style={
  fill opacity=0.8,
  draw opacity=1,
  text opacity=1,
  at={(0.03,0.03)},
  anchor=south west,
  draw=lightgray204,
  fill=gainsboro229
},
log basis x={10},
tick align=outside,
tick pos=left,
x dir=reverse,
x grid style={gainsboro229},
xmajorgrids,
xmin=244.021614456221, xmax=2082.36061847367,
xmode=log,
xtick style={color=dimgray85},
y grid style={gainsboro229},
ymajorgrids,
ymin=61, ymax=96,
ytick style={color=dimgray85},
ticklabel style={font=\tiny},
xtick style={draw=none},
ytick style={draw=none},
tick align=inside,
y label style={at={(axis description cs:0.22, 0.5)}},
ylabel=\textcolor{white}{.},
axis line style={gainsboro229},
x label style={at={(axis description cs:0.5, -0.03)}},
xlabel=\textcolor{white}{.},
title style={at={(0.5,0.06)},anchor=north},
title={\bf \textcolor{gray}{\scriptsize EuroSAT10}},
xticklabels={$\meanrelrand{\scriptstyle 10\%}{\scriptscriptstyle 270}$,$\meanrelrand{\scriptstyle 20\%}{\scriptscriptstyle 540}$,$\meanrelrand{\scriptstyle 30\%}{\scriptscriptstyle 810}$,$\meanrelrand{\scriptstyle 50\%}{\scriptscriptstyle ~~1,350}$,$\meanrelrand{\scriptstyle 70\%}{\scriptscriptstyle 1,890~~}$},
axis background/.style={fill=gainsboro247},
yticklabel style = {xshift=0.5ex},
xticklabel style = {yshift=0.5ex},
]

\path [fill=black, fill opacity=0.2, very thin]
(axis cs:269,76.76)
--(axis cs:269,72.72)
--(axis cs:539,84.89)
--(axis cs:810,88.89)
--(axis cs:1350,92.59)
--(axis cs:1889,94.53)
--(axis cs:1889,94.71)
--(axis cs:1889,94.71)
--(axis cs:1350,93.25)
--(axis cs:810,89.93)
--(axis cs:539,86.23)
--(axis cs:269,76.76)
--cycle;

\path [fill=darkgray155, fill opacity=0.2, very thin]
(axis cs:269,75.06)
--(axis cs:269,70.56)
--(axis cs:539,77.09)
--(axis cs:810,81.45)
--(axis cs:1350,86.88)
--(axis cs:1889,89.71)
--(axis cs:1889,90.99)
--(axis cs:1889,90.99)
--(axis cs:1350,88.22)
--(axis cs:810,84.67)
--(axis cs:539,80.85)
--(axis cs:269,75.06)
--cycle;

\path [fill=darkorange2551094, fill opacity=0.2, very thin]
(axis cs:269,66.72)
--(axis cs:269,61.2)
--(axis cs:539,79.09)
--(axis cs:810,84.92)
--(axis cs:1350,91.59)
--(axis cs:1889,93.48)
--(axis cs:1889,93.94)
--(axis cs:1889,93.94)
--(axis cs:1350,92.05)
--(axis cs:810,87.24)
--(axis cs:539,84.45)
--(axis cs:269,66.72)
--cycle;

\addplot [thick, black, mark=halfsquare right*, mark size=1.25, mark options={solid}, opacity=0.4, fill opacity=0.8]
table {%
270 74.74
540 85.56
810 89.41
1350 92.92
1890 94.62
};
\addlegendentry{TDDS}

\addplot [thick, darkgray155, dashed, mark=halfcircle*, mark size=1.25, mark options={solid}, opacity=0.4, fill opacity=0.8]
table {%
	270 72.81
	540 78.97
	810 83.06
	1350 87.55
	1890 90.35
};
\addlegendentry{Rand}

\addplot [thick, darkorange2551094, dashed, mark=*, mark size=1.25, mark options={solid}, opacity=0.4, fill opacity=0.8]
table {%
270 63.96
540 81.77
810 86.08
1350 91.82
1890 93.71
};
\addlegendentry{BlindCS}
\legend{},
\end{axis}

\end{tikzpicture}

%% file: nsample_accuracy_paper_vertical.tex
\begin{tikzpicture}

\definecolor{chocolate2267451}{RGB}{226,74,51}
\definecolor{darkgray155}{RGB}{155,155,155}
\definecolor{darkorange2551094}{RGB}{255,109,4}
\definecolor{dimgray85}{RGB}{85,85,85}
\definecolor{gainsboro229}{RGB}{239,239,239}
\definecolor{gainsboro247}{RGB}{253,253,253}
\definecolor{gray119}{RGB}{119,119,119}
\definecolor{lightgray204}{RGB}{204,204,204}
\definecolor{mediumpurple152142213}{RGB}{152,142,213}
\definecolor{sandybrown25119394}{RGB}{251,193,94}
\definecolor{steelblue52138189}{RGB}{52,138,189}
\definecolor{yellowgreen14218666}{RGB}{142,186,66}

\begin{axis}[
title style={at={(0.5,0.075)},anchor=north},
x label style={at={(axis description cs:1.08, -0.04)}},
width=8.75cm,
height=2.5cm,
axis background/.style={fill=gainsboro247},
axis line style={white},
log basis x={10},
tick align=outside,
tick pos=left,
x grid style={gainsboro229},
y grid style={gainsboro229},
xmajorgrids,
xmin=63.0957344480193, xmax=15848931.9246111,
xmode=log,
xtick style={color=dimgray85},
ymajorgrids,
ymin=63, ymax=66,
ytick style={color=dimgray85},
xtick={100, 1000,10000,100000,300000,1000000,3000000,10000000},
xticklabels={\tiny $\text{100}$, \tiny $\text{1,000}$, \tiny $\text{10\rm{K}}$, \tiny $\text{100\rm{K}}$, \tiny $\text{300\rm{K}}$, \tiny $\text{1\rm{M}}$, \tiny $\text{3\rm{M}}$, \tiny $\text{10\rm{M}}$},
ylabel=\textcolor{dimgray85}{\footnotesize Acc. (\%)},
ytick={63.0, 64.0, 65.0, 66},
yticklabels={\scriptsize 63.0, \scriptsize 64.0, \scriptsize 65.0, \scriptsize 66.0},
y label style={at={(axis description cs:0.06, 0.5)}},
yticklabel style = {xshift=0.5ex},
xticklabel style = {yshift=0.5ex},
ticklabel style={font=\tiny},
tick align=inside,
axis line style={gainsboro229},
xtick style={draw=none},
ytick style={draw=none},
]
\addplot [very thick, darkorange2551094, mark=*, mark size=1.25, mark options={solid}, opacity=0.4, fill opacity=0.8]
table {%
100 63.09
1000 64.98
10000 65.01
100000 65.05
300000 65.55
1000000 65.77
3000000 65.54
10000000 65.4
};
\end{axis}

\end{tikzpicture}

%% file: nsample_runtime_paper_vertical.tex
\begin{tikzpicture}

\definecolor{chocolate2267451}{RGB}{226,74,51}
\definecolor{darkgray155}{RGB}{155,155,155}
\definecolor{darkorange2551094}{RGB}{255,109,4}
\definecolor{dimgray85}{RGB}{85,85,85}
\definecolor{gainsboro229}{RGB}{239,239,239}
\definecolor{gainsboro247}{RGB}{253,253,253}
\definecolor{gray119}{RGB}{119,119,119}
\definecolor{lightgray204}{RGB}{204,204,204}
\definecolor{mediumpurple152142213}{RGB}{152,142,213}
\definecolor{sandybrown25119394}{RGB}{251,193,94}
\definecolor{steelblue52138189}{RGB}{52,138,189}
\definecolor{yellowgreen14218666}{RGB}{142,186,66}

\begin{axis}[
xlabel=\textcolor{dimgray85}{\footnotesize Number of Sample Iterations ($T$)},
title style={at={(0.5,0.075)},anchor=north},
width=8.75cm,
height=4.25cm,
axis background/.style={fill=gainsboro247},
axis line style={white},
log basis x={10},
tick align=outside,
tick pos=left,
x grid style={gainsboro229},
y grid style={gainsboro229},
xmajorgrids,
xmin=63.0957344480193, xmax=15848931.9246111,
xmode=log,
ymode=log,
xtick style={color=dimgray85},
ymajorgrids,
ymin=0.212, ymax=5122.61231311758,
ytick style={color=dimgray85},
xtick={100, 1000,10000,100000,300000,1000000,3000000,10000000},
xticklabels={\tiny $\text{100}$, \tiny $\text{1,000}$, \tiny $\text{10\rm{K}}$, \tiny $\text{100\rm{K}}$, \tiny $\text{300\rm{K}}$, \tiny $\text{1\rm{M}}$, \tiny $\text{3\rm{M}}$, \tiny $\text{10\rm{M}}$},
ylabel=\textcolor{dimgray85}{\footnotesize Runtime (\rm{s})},
ytick={0.365, 3.57, 9.27, 42.46, 115.36, 381.57, 1109.84, 3713.05},
yticklabels={\scriptsize 0.365, \scriptsize 3.57, \scriptsize 9.27, \scriptsize 42.46, \scriptsize 115.4, \scriptsize 381.6, \scriptsize $\text{1,110}$, \scriptsize $\text{3,713}$},
yticklabel style = {xshift=0.5ex},
xticklabel style = {yshift=0.5ex},
ticklabel style={font=\tiny},
tick align=inside,
axis line style={gainsboro229},
xtick style={draw=none},
ytick style={draw=none},
y label style={at={(axis description cs:0.06, 0.5)}},
x label style={at={(axis description cs:0.5, 0.1)}},
]
\addplot [very thick, darkgray155, mark=*, mark size=0.75, mark options={solid}, opacity=0.8, fill opacity=0.8]
table {%
100 0.365
1000 3.57
10000 9.27
100000 42.46
300000 115.36
1000000 381.57
3000000 1109.84
10000000 3713.05
};
\end{axis}

\end{tikzpicture}

%% file: imagenet_paper_zcore.tex
\begin{tikzpicture}

\definecolor{chocolate2267451}{RGB}{226,74,51}
\definecolor{darkgray155}{RGB}{155,155,155}
\definecolor{darkorange2551094}{RGB}{255,109,4}
\definecolor{dimgray85}{RGB}{85,85,85}
\definecolor{gainsboro229}{RGB}{239,239,239}
\definecolor{gainsboro247}{RGB}{253,253,253}
\definecolor{gray119}{RGB}{119,119,119}
\definecolor{lightgray204}{RGB}{204,204,204}
\definecolor{mediumpurple152142213}{RGB}{152,142,213}
\definecolor{sandybrown25119394}{RGB}{251,193,94}
\definecolor{steelblue52138189}{RGB}{52,138,189}
\definecolor{yellowgreen14218666}{RGB}{142,186,66}
\definecolor{neonpink}{RGB}{128,8,120}
\definecolor{lightpink255181184}{RGB}{255,181,184}
\definecolor{darkorange-clip}{RGB}{255,54,2}
\definecolor{darkorange-clip}{RGB}{200,175,4}
\definecolor{darkorange-res}{RGB}{200,0,0}


\begin{axis}[
title style={at={(0.5,0.075)},anchor=north},
width=8.9cm,
height=8.7cm,
axis line style={white},
legend cell align={left},
legend style={
  fill opacity=1,
  draw opacity=1,
  text opacity=1,
  at={(0.985,0.97)},
  anchor=north east,
  draw=lightgray204,
  fill=gainsboro247
},
log basis x={10},
tick align=inside,
title={\bf  \textcolor{gray}{ImageNet}},
x dir=reverse,
x grid style={gainsboro229},
y grid style={gainsboro229},
xmajorgrids,
xmin=121268.273641759, xmax=406053.315688034,
xmode=log,
ymajorgrids,
xtick={128116,256233,384350},
xticklabels={$\meanrelrand{\scriptscriptstyle 10\%}{\scriptscriptstyle 128,116~~}$,$\meanrelrand{\scriptscriptstyle 20\%}{\scriptscriptstyle 256,233}$,$\meanrelrand{\scriptscriptstyle 30\%}{\scriptscriptstyle 384,350}$},
ticklabel style={font=\scriptsize},
xtick style={draw=none},
ytick style={draw=none},
ylabel=\textcolor{white}{A},
axis background/.style={fill=gainsboro247},
axis line style={gainsboro229},
ymin=52, ymax=65,
xlabel=\textcolor{dimgray85}{$\xlabel{\text{\footnotesize \% of Dataset Selected for Coreset}}{\text{~/ Num.~of Training Examples in Coreset ($n$)}}$},
ylabel=\textcolor{dimgray85}{\footnotesize Accuracy (\%)},
x label style={at={(axis description cs:0.4475, 0.02)}},
y label style={at={(axis description cs:0.08, 0.5)}},
]
\addplot [ultra thick, darkorange-res, mark=halfcircle*, mark size=1.25, mark options={solid}, opacity=0.4, fill opacity=0.8, mark options={rotate=180}]
table {%
	384350 63.37
	256233 60.57
	128116 52.97
};
\addlegendentry{\scriptsize ZCore$_{\text{\tiny ResNet18}}$}
\addplot [ultra thick, darkorange2551094, mark=*, mark size=1.25, mark options={solid}, opacity=0.4, fill opacity=0.8]
table {%
	384350 64.43
	256233 61.31
	128116 53.99
};
\addlegendentry{\scriptsize ZCore$_{\text{\tiny ResNet18, CLIP}}$}
\addplot [ultra thick, darkorange-clip, dashed, mark=halfcircle*, mark size=1.25, mark options={solid}, opacity=0.4, fill opacity=0.8]
table {%
	384350 64.42
	256233 61.39
	128116 53.54
};
\addlegendentry{\scriptsize ZCore$_{\text{\tiny CLIP}}$}
\end{axis}

\end{tikzpicture}

%% file: dinov2_0.tex
\begin{tikzpicture}
\definecolor{chocolate2267451}{RGB}{50,50,50}
\definecolor{dimgray85}{RGB}{85,85,85}
\definecolor{gainsboro229}{RGB}{239,239,239}
\definecolor{gainsboro247}{RGB}{253,253,253}
\definecolor{darkorange2551094}{RGB}{255,109,4}
\definecolor{gray119}{RGB}{255,109,4}
\definecolor{mediumpurple152142213}{RGB}{142,186,66}
\definecolor{yellowgreen14218666}{RGB}{142,186,66}
\definecolor{steelblue52138189}{RGB}{52,138,189}
\definecolor{whitesmoke238}{RGB}{238,238,238}


\definecolor{lightgray204}{RGB}{204,204,204}

\begin{axis}[width=8.95cm,
	height=4.63cm,
	ymajorgrids,
	y grid style={gainsboro229},
	legend cell align={left},
	legend style={
		fill opacity=0,
		draw opacity=0,
		text opacity=1,
		at={(1.0,1.0)},
		anchor=north east,
		draw=lightgray204,
		fill=gainsboro247
	},
	axis background/.style={fill=gainsboro247},
	scaled y ticks=false,
	ylabel near ticks,
	ytick style={color=white},
	xtick pos=left,
	xticklabel style = {font=\scriptsize,yshift=0.5ex},
	yticklabel style = {font=\scriptsize,xshift=0.5ex},
	axis line style={white},
	tick align=outside,
	x grid style={white},
	xmajorgrids,
	xtick style={color=dimgray85},
	ymajorgrids,
xmin=-4.5, xmax=4,
ymin=0, ymax=7000,
ticklabel style={font=\tiny},
tick align=inside,
axis line style={gainsboro229},
ytick={2000,4000,6000},
yticklabels={2\textrm{K}, 4\textrm{K}, 6\textrm{K}},
y label style={at={(axis description cs:-0.04, 0.5)}},
yticklabel style = {xshift=0.2ex},
xticklabel style = {yshift=0.05ex},
ylabel=\textcolor{dimgray85} {\footnotesize Frequency},
]
\draw[draw=whitesmoke238,fill=chocolate2267451,fill opacity=0.5,very thin] (axis cs:-4.35493898391724,0) rectangle (axis cs:-4.03055553436279,9);
\addlegendimage{ybar,ybar legend,draw=whitesmoke238,fill=chocolate2267451,fill opacity=0.5,very thin}
\addlegendentry{\scriptsize DINOv2}

\draw[draw=whitesmoke238,fill=chocolate2267451,fill opacity=0.5,very thin] (axis cs:-4.03055553436279,0) rectangle (axis cs:-3.70617208480835,4);
\draw[draw=whitesmoke238,fill=chocolate2267451,fill opacity=0.5,very thin] (axis cs:-3.70617208480835,0) rectangle (axis cs:-3.38178863525391,27);
\draw[draw=whitesmoke238,fill=chocolate2267451,fill opacity=0.5,very thin] (axis cs:-3.38178863525391,0) rectangle (axis cs:-3.05740518569946,45);
\draw[draw=whitesmoke238,fill=chocolate2267451,fill opacity=0.5,very thin] (axis cs:-3.05740518569946,0) rectangle (axis cs:-2.73302173614502,112);
\draw[draw=whitesmoke238,fill=chocolate2267451,fill opacity=0.5,very thin] (axis cs:-2.73302173614502,0) rectangle (axis cs:-2.40863828659058,257);
\draw[draw=whitesmoke238,fill=chocolate2267451,fill opacity=0.5,very thin] (axis cs:-2.40863828659058,0) rectangle (axis cs:-2.08425483703613,471);
\draw[draw=whitesmoke238,fill=chocolate2267451,fill opacity=0.5,very thin] (axis cs:-2.08425483703613,0) rectangle (axis cs:-1.75987138748169,982);
\draw[draw=whitesmoke238,fill=chocolate2267451,fill opacity=0.5,very thin] (axis cs:-1.75987138748169,0) rectangle (axis cs:-1.43548793792725,1675);
\draw[draw=whitesmoke238,fill=chocolate2267451,fill opacity=0.5,very thin] (axis cs:-1.43548793792725,0) rectangle (axis cs:-1.1111044883728,2534);
\draw[draw=whitesmoke238,fill=chocolate2267451,fill opacity=0.5,very thin] (axis cs:-1.1111044883728,0) rectangle (axis cs:-0.786721038818359,3731);
\draw[draw=whitesmoke238,fill=chocolate2267451,fill opacity=0.5,very thin] (axis cs:-0.786721038818359,0) rectangle (axis cs:-0.462337589263916,5032);
\draw[draw=whitesmoke238,fill=chocolate2267451,fill opacity=0.5,very thin] (axis cs:-0.462337589263916,0) rectangle (axis cs:-0.137954139709472,6031);
\draw[draw=whitesmoke238,fill=chocolate2267451,fill opacity=0.5,very thin] (axis cs:-0.137954139709472,0) rectangle (axis cs:0.186429309844971,6488);
\draw[draw=whitesmoke238,fill=chocolate2267451,fill opacity=0.5,very thin] (axis cs:0.186429309844971,0) rectangle (axis cs:0.510812759399414,6773);
\draw[draw=whitesmoke238,fill=chocolate2267451,fill opacity=0.5,very thin] (axis cs:0.510812759399414,0) rectangle (axis cs:0.835196208953858,6142);
\draw[draw=whitesmoke238,fill=chocolate2267451,fill opacity=0.5,very thin] (axis cs:0.835196208953858,0) rectangle (axis cs:1.1595796585083,4609);
\draw[draw=whitesmoke238,fill=chocolate2267451,fill opacity=0.5,very thin] (axis cs:1.1595796585083,0) rectangle (axis cs:1.48396310806274,2721);
\draw[draw=whitesmoke238,fill=chocolate2267451,fill opacity=0.5,very thin] (axis cs:1.48396310806274,0) rectangle (axis cs:1.80834655761719,1399);
\draw[draw=whitesmoke238,fill=chocolate2267451,fill opacity=0.5,very thin] (axis cs:1.80834655761719,0) rectangle (axis cs:2.13273000717163,579);
\draw[draw=whitesmoke238,fill=chocolate2267451,fill opacity=0.5,very thin] (axis cs:2.13273000717163,0) rectangle (axis cs:2.45711345672607,248);
\draw[draw=whitesmoke238,fill=chocolate2267451,fill opacity=0.5,very thin] (axis cs:2.45711345672607,0) rectangle (axis cs:2.78149690628052,90);
\draw[draw=whitesmoke238,fill=chocolate2267451,fill opacity=0.5,very thin] (axis cs:2.78149690628052,0) rectangle (axis cs:3.10588035583496,31);
\draw[draw=whitesmoke238,fill=chocolate2267451,fill opacity=0.5,very thin] (axis cs:3.10588035583496,0) rectangle (axis cs:3.43026380538941,6);
\draw[draw=whitesmoke238,fill=chocolate2267451,fill opacity=0.5,very thin] (axis cs:3.43026380538941,0) rectangle (axis cs:3.75464725494385,4);
\draw[draw=whitesmoke238,fill=steelblue52138189,fill opacity=0.5,very thin] (axis cs:-4.35472000526212,0) rectangle (axis cs:-4.03035985185242,1984);
\addlegendimage{ybar,ybar legend,draw=whitesmoke238,fill=steelblue52138189,fill opacity=0.5,very thin}
\addlegendentry{\scriptsize Uniform}

\draw[draw=whitesmoke238,fill=steelblue52138189,fill opacity=0.5,very thin] (axis cs:-4.03035985185242,0) rectangle (axis cs:-3.70599969844273,1957);
\draw[draw=whitesmoke238,fill=steelblue52138189,fill opacity=0.5,very thin] (axis cs:-3.70599969844273,0) rectangle (axis cs:-3.38163954503304,2031);
\draw[draw=whitesmoke238,fill=steelblue52138189,fill opacity=0.5,very thin] (axis cs:-3.38163954503304,0) rectangle (axis cs:-3.05727939162334,1962);
\draw[draw=whitesmoke238,fill=steelblue52138189,fill opacity=0.5,very thin] (axis cs:-3.05727939162334,0) rectangle (axis cs:-2.73291923821365,1906);
\draw[draw=whitesmoke238,fill=steelblue52138189,fill opacity=0.5,very thin] (axis cs:-2.73291923821365,0) rectangle (axis cs:-2.40855908480396,2047);
\draw[draw=whitesmoke238,fill=steelblue52138189,fill opacity=0.5,very thin] (axis cs:-2.40855908480396,0) rectangle (axis cs:-2.08419893139427,2002);
\draw[draw=whitesmoke238,fill=steelblue52138189,fill opacity=0.5,very thin] (axis cs:-2.08419893139427,0) rectangle (axis cs:-1.75983877798457,2038);
\draw[draw=whitesmoke238,fill=steelblue52138189,fill opacity=0.5,very thin] (axis cs:-1.75983877798457,0) rectangle (axis cs:-1.43547862457488,2010);
\draw[draw=whitesmoke238,fill=steelblue52138189,fill opacity=0.5,very thin] (axis cs:-1.43547862457488,0) rectangle (axis cs:-1.11111847116519,2011);
\draw[draw=whitesmoke238,fill=steelblue52138189,fill opacity=0.5,very thin] (axis cs:-1.11111847116519,0) rectangle (axis cs:-0.786758317755494,1998);
\draw[draw=whitesmoke238,fill=steelblue52138189,fill opacity=0.5,very thin] (axis cs:-0.786758317755494,0) rectangle (axis cs:-0.4623981643458,2022);
\draw[draw=whitesmoke238,fill=steelblue52138189,fill opacity=0.5,very thin] (axis cs:-0.4623981643458,0) rectangle (axis cs:-0.138038010936108,1965);
\draw[draw=whitesmoke238,fill=steelblue52138189,fill opacity=0.5,very thin] (axis cs:-0.138038010936108,0) rectangle (axis cs:0.186322142473585,2027);
\draw[draw=whitesmoke238,fill=steelblue52138189,fill opacity=0.5,very thin] (axis cs:0.186322142473585,0) rectangle (axis cs:0.510682295883278,2015);
\draw[draw=whitesmoke238,fill=steelblue52138189,fill opacity=0.5,very thin] (axis cs:0.510682295883278,0) rectangle (axis cs:0.835042449292971,1949);
\draw[draw=whitesmoke238,fill=steelblue52138189,fill opacity=0.5,very thin] (axis cs:0.835042449292971,0) rectangle (axis cs:1.15940260270266,1984);
\draw[draw=whitesmoke238,fill=steelblue52138189,fill opacity=0.5,very thin] (axis cs:1.15940260270266,0) rectangle (axis cs:1.48376275611236,2044);
\draw[draw=whitesmoke238,fill=steelblue52138189,fill opacity=0.5,very thin] (axis cs:1.48376275611236,0) rectangle (axis cs:1.80812290952205,2056);
\draw[draw=whitesmoke238,fill=steelblue52138189,fill opacity=0.5,very thin] (axis cs:1.80812290952205,0) rectangle (axis cs:2.13248306293174,2027);
\draw[draw=whitesmoke238,fill=steelblue52138189,fill opacity=0.5,very thin] (axis cs:2.13248306293174,0) rectangle (axis cs:2.45684321634144,1953);
\draw[draw=whitesmoke238,fill=steelblue52138189,fill opacity=0.5,very thin] (axis cs:2.45684321634144,0) rectangle (axis cs:2.78120336975113,2081);
\draw[draw=whitesmoke238,fill=steelblue52138189,fill opacity=0.5,very thin] (axis cs:2.78120336975113,0) rectangle (axis cs:3.10556352316082,1961);
\draw[draw=whitesmoke238,fill=steelblue52138189,fill opacity=0.5,very thin] (axis cs:3.10556352316082,0) rectangle (axis cs:3.42992367657052,1995);
\draw[draw=whitesmoke238,fill=steelblue52138189,fill opacity=0.5,very thin] (axis cs:3.42992367657052,0) rectangle (axis cs:3.75428382998021,1975);
\draw[draw=whitesmoke238,fill=mediumpurple152142213,fill opacity=0.5,very thin] (axis cs:-4.00579060070183,0) rectangle (axis cs:-3.67815778958795,2);
\addlegendimage{ybar,ybar legend,draw=whitesmoke238,fill=mediumpurple152142213,fill opacity=0.5,very thin}
\addlegendentry{\scriptsize Gaussian}

\draw[draw=whitesmoke238,fill=mediumpurple152142213,fill opacity=0.5,very thin] (axis cs:-3.67815778958795,0) rectangle (axis cs:-3.35052497847407,7);
\draw[draw=whitesmoke238,fill=mediumpurple152142213,fill opacity=0.5,very thin] (axis cs:-3.35052497847407,0) rectangle (axis cs:-3.0228921673602,16);
\draw[draw=whitesmoke238,fill=mediumpurple152142213,fill opacity=0.5,very thin] (axis cs:-3.0228921673602,0) rectangle (axis cs:-2.69525935624632,73);
\draw[draw=whitesmoke238,fill=mediumpurple152142213,fill opacity=0.5,very thin] (axis cs:-2.69525935624632,0) rectangle (axis cs:-2.36762654513244,211);
\draw[draw=whitesmoke238,fill=mediumpurple152142213,fill opacity=0.5,very thin] (axis cs:-2.36762654513244,0) rectangle (axis cs:-2.03999373401857,479);
\draw[draw=whitesmoke238,fill=mediumpurple152142213,fill opacity=0.5,very thin] (axis cs:-2.03999373401857,0) rectangle (axis cs:-1.71236092290469,892);
\draw[draw=whitesmoke238,fill=mediumpurple152142213,fill opacity=0.5,very thin] (axis cs:-1.71236092290469,0) rectangle (axis cs:-1.38472811179081,1761);
\draw[draw=whitesmoke238,fill=mediumpurple152142213,fill opacity=0.5,very thin] (axis cs:-1.38472811179081,0) rectangle (axis cs:-1.05709530067693,2921);
\draw[draw=whitesmoke238,fill=mediumpurple152142213,fill opacity=0.5,very thin] (axis cs:-1.05709530067693,0) rectangle (axis cs:-0.729462489563058,4372);
\draw[draw=whitesmoke238,fill=mediumpurple152142213,fill opacity=0.5,very thin] (axis cs:-0.729462489563058,0) rectangle (axis cs:-0.401829678449181,5808);
\draw[draw=whitesmoke238,fill=mediumpurple152142213,fill opacity=0.5,very thin] (axis cs:-0.401829678449181,0) rectangle (axis cs:-0.0741968673353046,6772);
\draw[draw=whitesmoke238,fill=mediumpurple152142213,fill opacity=0.5,very thin] (axis cs:-0.0741968673353046,0) rectangle (axis cs:0.253435943778572,6772);
\draw[draw=whitesmoke238,fill=mediumpurple152142213,fill opacity=0.5,very thin] (axis cs:0.253435943778572,0) rectangle (axis cs:0.581068754892449,6290);
\draw[draw=whitesmoke238,fill=mediumpurple152142213,fill opacity=0.5,very thin] (axis cs:0.581068754892449,0) rectangle (axis cs:0.908701566006326,5119);
\draw[draw=whitesmoke238,fill=mediumpurple152142213,fill opacity=0.5,very thin] (axis cs:0.908701566006326,0) rectangle (axis cs:1.2363343771202,3637);
\draw[draw=whitesmoke238,fill=mediumpurple152142213,fill opacity=0.5,very thin] (axis cs:1.2363343771202,0) rectangle (axis cs:1.56396718823408,2330);
\draw[draw=whitesmoke238,fill=mediumpurple152142213,fill opacity=0.5,very thin] (axis cs:1.56396718823408,0) rectangle (axis cs:1.89159999934796,1397);
\draw[draw=whitesmoke238,fill=mediumpurple152142213,fill opacity=0.5,very thin] (axis cs:1.89159999934796,0) rectangle (axis cs:2.21923281046183,667);
\draw[draw=whitesmoke238,fill=mediumpurple152142213,fill opacity=0.5,very thin] (axis cs:2.21923281046183,0) rectangle (axis cs:2.54686562157571,297);
\draw[draw=whitesmoke238,fill=mediumpurple152142213,fill opacity=0.5,very thin] (axis cs:2.54686562157571,0) rectangle (axis cs:2.87449843268959,121);
\draw[draw=whitesmoke238,fill=mediumpurple152142213,fill opacity=0.5,very thin] (axis cs:2.87449843268959,0) rectangle (axis cs:3.20213124380346,37);
\draw[draw=whitesmoke238,fill=mediumpurple152142213,fill opacity=0.5,very thin] (axis cs:3.20213124380346,0) rectangle (axis cs:3.52976405491734,16);
\draw[draw=whitesmoke238,fill=mediumpurple152142213,fill opacity=0.5,very thin] (axis cs:3.52976405491734,0) rectangle (axis cs:3.85739686603122,3);
\draw[draw=whitesmoke238,fill=gray119,fill opacity=0.5,very thin] (axis cs:-4.31828506101976,0) rectangle (axis cs:-3.99620936756243,182);
\addlegendimage{ybar,ybar legend,draw=whitesmoke238,fill=gray119,fill opacity=0.5,very thin}
\addlegendentry{\scriptsize Triangular}

\draw[draw=whitesmoke238,fill=gray119,fill opacity=0.5,very thin] (axis cs:-3.99620936756243,0) rectangle (axis cs:-3.6741336741051,518);
\draw[draw=whitesmoke238,fill=gray119,fill opacity=0.5,very thin] (axis cs:-3.6741336741051,0) rectangle (axis cs:-3.35205798064777,795);
\draw[draw=whitesmoke238,fill=gray119,fill opacity=0.5,very thin] (axis cs:-3.35205798064777,0) rectangle (axis cs:-3.02998228719043,992);
\draw[draw=whitesmoke238,fill=gray119,fill opacity=0.5,very thin] (axis cs:-3.02998228719043,0) rectangle (axis cs:-2.7079065937331,1274);
\draw[draw=whitesmoke238,fill=gray119,fill opacity=0.5,very thin] (axis cs:-2.7079065937331,0) rectangle (axis cs:-2.38583090027577,1645);
\draw[draw=whitesmoke238,fill=gray119,fill opacity=0.5,very thin] (axis cs:-2.38583090027577,0) rectangle (axis cs:-2.06375520681844,1962);
\draw[draw=whitesmoke238,fill=gray119,fill opacity=0.5,very thin] (axis cs:-2.06375520681844,0) rectangle (axis cs:-1.7416795133611,2303);
\draw[draw=whitesmoke238,fill=gray119,fill opacity=0.5,very thin] (axis cs:-1.7416795133611,0) rectangle (axis cs:-1.41960381990377,2480);
\draw[draw=whitesmoke238,fill=gray119,fill opacity=0.5,very thin] (axis cs:-1.41960381990377,0) rectangle (axis cs:-1.09752812644644,2817);
\draw[draw=whitesmoke238,fill=gray119,fill opacity=0.5,very thin] (axis cs:-1.09752812644644,0) rectangle (axis cs:-0.775452432989105,3019);
\draw[draw=whitesmoke238,fill=gray119,fill opacity=0.5,very thin] (axis cs:-0.775452432989105,0) rectangle (axis cs:-0.453376739531773,3364);
\draw[draw=whitesmoke238,fill=gray119,fill opacity=0.5,very thin] (axis cs:-0.453376739531773,0) rectangle (axis cs:-0.131301046074441,3713);
\draw[draw=whitesmoke238,fill=gray119,fill opacity=0.5,very thin] (axis cs:-0.131301046074441,0) rectangle (axis cs:0.190774647382892,3825);
\draw[draw=whitesmoke238,fill=gray119,fill opacity=0.5,very thin] (axis cs:0.190774647382892,0) rectangle (axis cs:0.512850340840225,3618);
\draw[draw=whitesmoke238,fill=gray119,fill opacity=0.5,very thin] (axis cs:0.512850340840225,0) rectangle (axis cs:0.834926034297557,3335);
\draw[draw=whitesmoke238,fill=gray119,fill opacity=0.5,very thin] (axis cs:0.834926034297557,0) rectangle (axis cs:1.15700172775489,2915);
\draw[draw=whitesmoke238,fill=gray119,fill opacity=0.5,very thin] (axis cs:1.15700172775489,0) rectangle (axis cs:1.47907742121222,2617);
\draw[draw=whitesmoke238,fill=gray119,fill opacity=0.5,very thin] (axis cs:1.47907742121222,0) rectangle (axis cs:1.80115311466955,2298);
\draw[draw=whitesmoke238,fill=gray119,fill opacity=0.5,very thin] (axis cs:1.80115311466955,0) rectangle (axis cs:2.12322880812689,1935);
\draw[draw=whitesmoke238,fill=gray119,fill opacity=0.5,very thin] (axis cs:2.12322880812689,0) rectangle (axis cs:2.44530450158422,1543);
\draw[draw=whitesmoke238,fill=gray119,fill opacity=0.5,very thin] (axis cs:2.44530450158422,0) rectangle (axis cs:2.76738019504155,1212);
\draw[draw=whitesmoke238,fill=gray119,fill opacity=0.5,very thin] (axis cs:2.76738019504155,0) rectangle (axis cs:3.08945588849888,887);
\draw[draw=whitesmoke238,fill=gray119,fill opacity=0.5,very thin] (axis cs:3.08945588849888,0) rectangle (axis cs:3.41153158195622,541);
\draw[draw=whitesmoke238,fill=gray119,fill opacity=0.5,very thin] (axis cs:3.41153158195622,0) rectangle (axis cs:3.73360727541355,210);
\addplot [ultra thick, opacity=0.75, chocolate2267451, forget plot]
table {%
-4.35493898391724 3.26875445735056
-4.31418729427472 4.44971623952888
-4.2734356046322 5.69240342173251
-4.23268391498968 6.81739795914981
-4.19193222534716 7.61809174016914
-4.15118053570465 7.93587650884178
-4.11042884606213 7.72460638190447
-4.06967715641961 7.07641619465194
-4.02892546677709 6.21104071772928
-3.98817377713457 5.433563419051
-3.94742208749206 5.05698458217008
-3.90667039784954 5.3079709129012
-3.86591870820702 6.26944315405135
-3.8251670185645 7.90045958969179
-3.78441532892198 10.1051385962404
-3.74366363927947 12.7734786082889
-3.70291194963695 15.7539378703838
-3.66216025999443 18.8084010523213
-3.62140857035191 21.6408192191597
-3.58065688070939 24.021219178382
-3.53990519106688 25.9113341064168
-3.49915350142436 27.4712797658231
-3.45840181178184 28.9373536036955
-3.41765012213932 30.4934331326823
-3.3768984324968 32.2579419025449
-3.33614674285429 34.3745611319465
-3.29539505321177 37.0838609016789
-3.25464336356925 40.6842823967073
-3.21389167392673 45.4165428252563
-3.17313998428421 51.3831031875618
-3.1323882946417 58.564735346851
-3.09163660499918 66.8812730931532
-3.05088491535666 76.1997406659179
-3.01013322571414 86.275589303892
-2.96938153607162 96.7234568666591
-2.92862984642911 107.122679478062
-2.88787815678659 117.263211210308
-2.84712646714407 127.421141737207
-2.80637477750155 138.489364586209
-2.76562308785903 151.807779117952
-2.72487139821652 168.660778270837
-2.684119708574 189.613072660807
-2.64336801893148 214.027301201052
-2.60261632928896 240.105387083148
-2.56186463964644 265.555118272327
-2.52111295000393 288.612116560716
-2.48036126036141 308.905595079252
-2.43960957071889 327.74235516955
-2.39885788107637 347.719830214206
-2.35810619143385 371.899639966622
-2.31735450179134 402.90384381097
-2.27660281214882 442.246103403748
-2.2358511225063 490.054321496845
-2.19509943286378 545.169496765987
-2.15434774322126 605.526626189813
-2.11359605357875 668.751254998691
-2.07284436393623 732.897141770772
-2.03209267429371 797.118355134988
-1.99134098465119 861.954039682337
-1.95058929500867 928.988808829117
-1.90983760536616 999.938343587383
-1.86908591572364 1075.59073441897
-1.82833422608112 1155.32739354911
-1.7875825364386 1237.78309846039
-1.74683084679608 1322.35391102897
-1.70607915715357 1410.30226286927
-1.66532746751105 1504.26372003295
-1.62457577786853 1606.30058258779
-1.58382408822601 1716.03997274774
-1.54307239858349 1830.44481514429
-1.50232070894098 1945.39664798135
-1.46156901929846 2057.83605503908
-1.42081732965594 2166.90822138803
-1.38006564001342 2273.53599839287
-1.3393139503709 2379.24340355617
-1.29856226072839 2485.63807619186
-1.25781057108587 2595.05776615709
-1.21705888144335 2711.28906134761
-1.17630719180083 2838.79869368357
-1.13555550215831 2980.40344021255
-1.0948038125158 3135.24469426096
-1.05405212287328 3299.03634232494
-1.01330043323076 3466.56130987701
-0.972548743588241 3634.45655121937
-0.931797053945723 3802.34596239665
-0.891045364303205 3971.90069138678
-0.850293674660688 4144.69637045619
-0.809541985018169 4320.06011390245
-0.768790295375651 4493.99024917041
-0.728038605733133 4660.01287544193
-0.687286916090615 4812.03894911391
-0.646535226448097 4947.79967609809
-0.605783536805579 5070.47810795124
-0.565031847163061 5187.1484948826
-0.524280157520543 5305.14438558106
-0.483528467878025 5429.06625716486
-0.442776778235507 5560.04278049622
-0.402025088592989 5696.36499101013
-0.361273398950471 5833.76064822177
-0.320521709307953 5965.33274852684
-0.279770019665436 6082.96874582012
-0.239018330022917 6180.92073460334
-0.198266640380399 6259.14068022995
-0.157514950737881 6322.92777186445
-0.116763261095363 6378.584317297
-0.0760115714528453 6428.89204136632
-0.0352598818103269 6472.48711279617
0.00549180783219061 6507.27262080549
0.046243497474709 6534.27650207609
0.0869951871172265 6558.42027858159
0.127746876759745 6585.67568044486
0.168498566402263 6618.98033170473
0.209250256044781 6656.07165730713
0.250001945687299 6690.82071117875
0.290753635329817 6716.79154025247
0.331505324972335 6729.90641665916
0.372257014614853 6728.19186971316
0.413008704257371 6709.71742249904
0.453760393899889 6671.63613763698
0.494512083542407 6611.65404270732
0.535263773184925 6530.20677694814
0.576015462827443 6430.6864480349
0.616767152469961 6317.16298509501
0.657518842112479 6191.61006361365
0.698270531754997 6052.94928157922
0.739022221397515 5898.36876542619
0.779773911040033 5725.56943015896
0.820525600682551 5534.38642382814
0.861277290325069 5327.17910904282
0.902028979967588 5108.18681839838
0.942780669610105 4882.21694464742
0.983532359252624 4653.07843458537
1.02428404889514 4422.49782935789
1.06503573853766 4190.36058065036
1.10578742818018 3956.31290107771
1.1465391178227 3721.46181540919
1.18729080746521 3488.58809510286
1.22804249710773 3260.65322264244
1.26879418675025 3039.16172470405
1.30954587639277 2824.11196084909
1.35029756603529 2615.519524978
1.3910492556778 2414.77199409819
1.43180094532032 2224.31232897317
1.47255263496284 2045.98928143515
1.51330432460536 1879.74997973453
1.55405601424787 1723.86831950129
1.59480770389039 1576.35677221987
1.63555939353291 1436.25055286076
1.67631108317543 1303.82999163682
1.71706277281795 1179.91089595815
1.75781446246047 1064.97410064555
1.79856615210298 958.731633506174
1.8393178417455 860.255450338803
1.88006953138802 768.561039183258
1.92082122103054 683.410492862313
1.96157291067306 605.806165087275
2.00232460031557 537.552114745613
2.04307628995809 479.89605876296
2.08382797960061 432.226566655809
2.12457966924313 391.960753207434
2.16533135888565 355.757180502761
2.20608304852816 321.05449112331
2.24683473817068 286.841541983788
2.2875864278132 253.432511721181
2.32833811745572 221.800475885837
2.36908980709824 193.040529206538
2.40984149674075 168.074061357013
2.45059318638327 147.408231441919
2.49134487602579 130.879496245936
2.53209656566831 117.576558893507
2.57284825531083 106.154806401273
2.61359994495334 95.4345948856118
2.65435163459586 84.8705175602524
2.69510332423838 74.5687676393962
2.7358550138809 64.9352043458596
2.77660670352342 56.3091444707121
2.81735839316593 48.8251976744503
2.85811008280845 42.4601452651551
2.89886177245097 37.1019292893676
2.93961346209349 32.5721492725213
2.98036515173601 28.6413527269254
3.02111684137852 25.0711100231469
3.06186853102104 21.6646077923445
3.10262022066356 18.3009784093519
3.14337191030608 14.9545773056589
3.1841235999486 11.7044751256999
3.22487528959111 8.72203359204315
3.26562697923363 6.22182266822761
3.30637866887615 4.38251512431454
3.34713035851867 3.26866416061381
3.38788204816119 2.79273067142571
3.4286337378037 2.74311391458775
3.46938542744622 2.86991031155264
3.51013711708874 2.98194881214825
3.55088880673126 2.99737245695976
3.59164049637378 2.92316842731798
3.63239218601629 2.79238713299057
3.67314387565881 2.61360840677699
3.71389556530133 2.36555490156814
3.75464725494385 2.02736735747069
};
\addplot [very thick, opacity=0.75, steelblue52138189, forget plot]
table {%
-4.35472000526212 989.383249770604
-4.31397124227095 1108.23450684201
-4.27322247927978 1224.33495899253
-4.23247371628861 1335.22186710237
-4.19172495329745 1438.79576360709
-4.15097619030628 1533.43821997276
-4.11022742731511 1618.07707602408
-4.06947866432394 1692.19585248112
-4.02872990133278 1755.79191790912
-3.98798113834161 1809.29445092836
-3.94723237535044 1853.45731342396
-3.90648361235927 1889.2431704454
-3.86573484936811 1917.71369425862
-3.82498608637694 1939.93711318277
-3.78423732338577 1956.9196595962
-3.7434885603946 1969.56265917886
-3.70273979740344 1978.64296345667
-3.66199103441227 1984.81173648694
-3.6212422714211 1988.60546859177
-3.58049350842993 1990.4633569219
-3.53974474543877 1990.7464538162
-3.4989959824476 1989.75570059878
-3.45824721945643 1987.74761183726
-3.41749845646526 1984.94756147581
-3.3767496934741 1981.5611610524
-3.33600093048293 1977.784143333
-3.29525216749176 1973.81068004694
-3.25450340450059 1969.83947115155
-3.21375464150943 1966.07654143456
-3.17300587851826 1962.73367316536
-3.13225711552709 1960.02185197857
-3.09150835253593 1958.13991729038
-3.05075958954476 1957.25958344929
-3.01001082655359 1957.50887885332
-2.96926206356242 1958.9566062328
-2.92851330057126 1961.60051087266
-2.88776453758009 1965.36142382713
-2.84701577458892 1970.08480906551
-2.80626701159775 1975.55005818918
-2.76551824860659 1981.4867544109
-2.72476948561542 1987.59616830563
-2.68402072262425 1993.57559861669
-2.64327195963308 1999.14290367451
-2.60252319664192 2004.05867581074
-2.56177443365075 2008.14392243368
-2.52102567065958 2011.29172701909
-2.48027690766841 2013.47205932369
-2.43952814467725 2014.72959427455
-2.39877938168608 2015.17502394605
-2.35803061869491 2014.97088019552
-2.31728185570374 2014.31332062284
-2.27653309271258 2013.41166361798
-2.23578432972141 2012.46767241854
-2.19503556673024 2011.65664811861
-2.15428680373907 2011.11225248848
-2.11353804074791 2010.91660870123
-2.07278927775674 2011.09662181394
-2.03204051476557 2011.62667440208
-1.9912917517744 2012.4369954883
-1.95054298878324 2013.42622167347
-1.90979422579207 2014.47612285751
-1.8690454628009 2015.46627053223
-1.82829669980973 2016.28663270935
-1.78754793681857 2016.84664511095
-1.7467991738274 2017.08010876326
-1.70605041083623 2016.94612069839
-1.66530164784506 2016.42696816118
-1.6245528848539 2015.52435546808
-1.58380412186273 2014.25540767326
-1.54305535887156 2012.64961750667
-1.50230659588039 2010.74736590106
-1.46155783288923 2008.60000234907
-1.42080906989806 2006.27088696551
-1.38006030690689 2003.83641493121
-1.33931154391572 2001.38595195769
-1.29856278092456 1999.01981714163
-1.25781401793339 1996.8448933857
-1.21706525494222 1994.96800664711
-1.17631649195105 1993.48775193358
-1.13556772895989 1992.48582851908
-1.09481896596872 1992.01909766701
-1.05407020297755 1992.11347609728
-1.01332143998638 1992.76047469205
-0.972572676995217 1993.91677738011
-0.93182391400405 1995.50683728446
-0.891075151012882 1997.42813482035
-0.850326388021715 1999.55854080987
-0.809577625030547 2001.76515086666
-0.76882886203938 2003.91395912732
-0.728080099048213 2005.87975863324
-0.687331336057045 2007.5556451027
-0.646582573065877 2008.86144884022
-0.60583381007471 2009.75035561799
-0.565085047083543 2010.21295990921
-0.524336284092375 2010.27808510706
-0.483587521101208 2010.00994384144
-0.44283875811004 2009.50159259178
-0.402089995118873 2008.86510677317
-0.361341232127705 2008.2193791558
-0.320592469136538 2007.67682925173
-0.27984370614537 2007.33052384822
-0.239094943154202 2007.24320562998
-0.198346180163036 2007.43950951874
-0.157597417171868 2007.90225745627
-0.116848654180701 2008.57323098684
-0.076099891189533 2009.35830480604
-0.0353511281983652 2010.13635328511
0.00539763479280175 2010.77096729576
0.0461463977839696 2011.12376989678
0.0868951607751365 2011.0680064091
0.127643923766304 2010.50110297593
0.168392686757472 2009.35502586014
0.209141449748639 2007.60351547052
0.249890212739807 2005.2655950006
0.290638975730974 2002.40513839908
0.331387738722142 1999.12669178097
0.372136501713309 1995.56813122936
0.412885264704476 1991.89105534174
0.453634027695644 1988.2700008142
0.494382790686812 1984.88159581024
0.535131553677979 1981.8946186772
0.575880316669147 1979.46163579175
0.616629079660314 1977.71251599424
0.657377842651481 1976.74974933217
0.698126605642649 1976.64522684087
0.738875368633816 1977.43803476318
0.779624131624984 1979.13290428711
0.820372894616151 1981.69920377297
0.861121657607319 1985.07068145063
0.901870420598486 1989.14645092254
0.942619183589653 1993.79385079993
0.983367946580821 1998.85373043382
1.02411670957199 2004.14840256088
1.06486547256316 2009.49201471044
1.10561423555432 2014.70253479875
1.14636299854549 2019.61406228166
1.18711176153666 2024.08789775989
1.22786052452783 2028.02082209125
1.26860928751899 2031.34937414532
1.30935805051016 2034.04952101087
1.35010681350133 2036.13186648043
1.3908555764925 2037.63328560546
1.43160433948366 2038.60644762704
1.47235310247483 2039.10897777431
1.513101865466 2039.19395887922
1.55385062845717 2038.90311387302
1.59459939144833 2038.2634368091
1.6353481544395 2037.28739170443
1.67609691743067 2035.97621763127
1.71684568042184 2034.32547389667
1.757594443413 2032.33177993736
1.79834320640417 2029.99973312695
1.83909196939534 2027.34815342441
1.8798407323865 2024.41501276723
1.92058949537767 2021.26057796825
1.96133825836884 2017.96838972421
2.00208702136001 2014.6437343534
2.04283578435118 2011.40930428012
2.08358454734234 2008.39787377263
2.12433331033351 2005.74210578614
2.16508207332468 2003.56206783483
2.20583083631584 2001.95160933267
2.24657959930701 2000.96530976395
2.28732836229818 2000.60807689091
2.32807712528935 2000.82949682442
2.36882588828052 2001.52461543845
2.40957465127168 2002.54196861436
2.45032341426285 2003.6984996841
2.49107217725402 2004.79972696231
2.53182094024518 2005.66242232273
2.57256970323635 2006.13638567064
2.61331846622752 2006.12181829924
2.65406722921869 2005.57934814027
2.69481599220985 2004.53083509618
2.73556475520102 2003.05045828364
2.77631351819219 2001.24696904341
2.81706228118336 1999.23910421498
2.85781104417452 1997.12679495962
2.89855980716569 1994.96090921041
2.93930857015686 1992.7139069058
2.98005733314803 1990.25316267138
3.02080609613919 1987.318080128
3.06155485913036 1983.50173473277
3.10230362212153 1978.23780426092
3.1430523851127 1970.79400719564
3.18380114810386 1960.2740401627
3.22454991109503 1945.63081418203
3.2652986740862 1925.69427429092
3.30604743707737 1899.21687887133
3.34679620006853 1864.93863465565
3.3875449630597 1821.67133610206
3.42829372605087 1768.39850237635
3.46904248904204 1704.38387596418
3.5097912520332 1629.27791226083
3.55054001502437 1543.20924086403
3.59128877801554 1446.84736450189
3.63203754100671 1341.42439351107
3.67278630399787 1228.70750277697
3.71353506698904 1110.91965793641
3.75428382998021 990.613125440284
};
\addplot [very thick, opacity=0.75, mediumpurple152142213, forget plot]
table {%
-4.00579060070183 2.27754699303446
-3.96627709584387 2.33358213638156
-3.92676359098592 2.12110185040466
-3.88725008612796 1.73264947905236
-3.84773658127001 1.32727473812496
-3.80822307641205 1.07018025044111
-3.7687095715541 1.08704531031334
-3.72919606669614 1.44796846151257
-3.68968256183819 2.16890479816275
-3.65016905698023 3.21032728913657
-3.61065555212228 4.46443861562255
-3.57114204726432 5.742517292761
-3.53162854240637 6.79117458272803
-3.49211503754841 7.36464713376995
-3.45260153269046 7.34624601857606
-3.4130880278325 6.86045892943753
-3.37357452297455 6.29568933331495
-3.33406101811659 6.20147792910574
-3.29454751325864 7.10803431240736
-3.25503400840068 9.37029008395177
-3.21552050354273 13.1157324859673
-3.17600699868477 18.2927227227945
-3.13649349382682 24.7420589580298
-3.09697998896886 32.2143451673178
-3.05746648411091 40.3328672419593
-3.01795297925295 48.5823687783222
-2.978439474395 56.4035354281502
-2.93892596953704 63.3899378321723
-2.89941246467909 69.5024264934724
-2.85989895982113 75.2017993681158
-2.82038545496318 81.4376476974217
-2.78087195010522 89.4735070869149
-2.74135844524727 100.568739201525
-2.70184494038931 115.589815962039
-2.66233143553136 134.675990888009
-2.6228179306734 157.107210086799
-2.58330442581545 181.488383086929
-2.54379092095749 206.244840307665
-2.50427741609954 230.236954967398
-2.46476391124158 253.18755141728
-2.42525040638363 275.733948774509
-2.38573690152567 299.196808149712
-2.34622339666772 325.298090432967
-2.30670989180976 355.901246113301
-2.26719638695181 392.619812248137
-2.22768288209385 436.192580049417
-2.1881693772359 485.87212259547
-2.14865587237794 539.352085777617
-2.10914236751999 593.579137969639
-2.06962886266203 646.168648999844
-2.03011535780408 696.586680385779
-1.99060185294612 746.369155274597
-1.95108834808817 798.400408284283
-1.91157484323021 855.916896056815
-1.87206133837226 921.792889514389
-1.8325478335143 998.079401389177
-1.79303432865635 1085.5283075551
-1.75352082379839 1183.18808348638
-1.71400731894044 1288.48921116014
-1.67449381408248 1398.00527667854
-1.63498030922453 1508.56733814156
-1.59546680436657 1618.23926822765
-1.55595329950862 1726.89215164615
-1.51643979465066 1836.32259242569
-1.47692628979271 1949.86394764638
-1.43741278493475 2071.47796715116
-1.3978992800768 2204.53987976002
-1.35838577521884 2350.75830041481
-1.31887227036089 2509.66356292717
-1.27935876550293 2678.83746224838
-1.23984526064498 2854.72918882688
-1.20033175578702 3033.71969718431
-1.16081825092907 3213.10272194223
-1.12130474607111 3391.6818323316
-1.08179124121316 3569.6577000715
-1.0422777363552 3747.64942105626
-1.00276423149725 3925.30445875968
-0.963250726639292 4100.61130854909
-0.923737221781337 4270.83448376348
-0.884223716923382 4434.66763914077
-0.844710212065427 4593.85689603787
-0.805196707207472 4752.69852890259
-0.765683202349517 4915.56890766535
-0.726169697491562 5084.35653518366
-0.686656192633607 5257.66375495306
-0.647142687775652 5432.00234574409
-0.607629182917697 5603.59209223438
-0.568115678059742 5769.27106520344
-0.528602173201787 5926.28911488143
-0.489088668343832 6071.93434522749
-0.449575163485877 6203.92211307731
-0.410061658627922 6321.49904060359
-0.370548153769967 6426.32245481429
-0.331034648912012 6522.07952254873
-0.291521144054057 6612.5048021169
-0.252007639196102 6698.55466450759
-0.212494134338147 6776.46312354229
-0.172980629480192 6838.47744646512
-0.133467124622237 6876.64692422869
-0.0939536197642821 6887.68098150423
-0.0544401149063272 6875.58118026221
-0.0149266100483723 6850.07923739142
0.0245868948095831 6821.96785351318
0.064100399667538 6798.41022168756
0.103613904525493 6780.67761448754
0.143127409383448 6764.69252282262
0.182640914241403 6743.35478486829
0.222154419099358 6709.44384641241
0.261667923957313 6658.06785745171
0.301181428815267 6587.75052185654
0.340694933673222 6499.77833425707
0.380208438531177 6396.47584692698
0.419721943389132 6279.75895918823
0.459235448247088 6150.87100883219
0.498748953105043 6011.09254206436
0.538262457962998 5862.58198703717
0.577775962820953 5708.75324184925
0.617289467678908 5554.03736075367
0.656802972536862 5402.89919748308
0.696316477394817 5258.00327809302
0.735829982252772 5118.15143467397
0.775343487110727 4977.61568928498
0.814856991968682 4828.28860430981
0.854370496826638 4664.05015834721
0.893884001684593 4484.49363340516
0.933397506542548 4295.14013550551
0.972911011400503 4103.97693109458
1.01242451625846 3917.10951688362
1.05193802111641 3736.67688563436
1.09145152597437 3561.86640594192
1.13096503083232 3391.39418018721
1.17047853569028 3225.26374639272
1.20999204054823 3064.75345379894
1.24950554540619 2911.00737631118
1.28901905026414 2763.47402200885
1.3285325551221 2619.54467157793
1.36804605998005 2475.93915428964
1.40755956483801 2330.96846784346
1.44707306969596 2185.93478383767
1.48658657455392 2044.58896856848
1.52610007941187 1911.18895024756
1.56561358426983 1788.67160313798
1.60512708912778 1677.94309186435
1.64464059398574 1578.08157707389
1.68415409884369 1486.66978317243
1.72366760370165 1399.98826084139
1.7631811085596 1313.59819053137
1.80269461341756 1223.80074116542
1.84220811827551 1129.46080840222
1.88172162313347 1032.86408146807
1.92123512799142 938.729765426087
1.96074863284938 851.912622408948
2.00026213770733 775.304546934046
2.03977564256529 709.109272413645
2.07928914742324 651.503666939947
2.1188026522812 599.844812379886
2.15831615713915 551.612370992064
2.19782966199711 504.859882843805
2.23734316685506 458.398312902612
2.27685667171302 411.919561744749
2.31637017657097 366.019800131419
2.35588368142893 321.997145374219
2.39539718628688 281.445791752241
2.43491069114484 245.804579910538
2.47442419600279 215.983331838233
2.51393770086075 192.102660639342
2.5534512057187 173.400850890203
2.59296471057666 158.420866008363
2.63247821543461 145.497531418835
2.67199172029257 133.319569947363
2.71150522515052 121.225282011186
2.75101873000848 109.109565638127
2.79053223486643 97.1553528440172
2.83004573972439 85.6555828024693
2.86955924458234 74.9439315410186
2.9090727494403 65.2896558023024
2.94858625429825 56.7455760913717
2.98809975915621 49.1164472151518
3.02761326401416 42.1389176519103
3.06712676887212 35.7256453186219
3.10664027373007 30.0413753849535
3.14615377858803 25.3488577346798
3.18566728344598 21.7802982564452
3.22518078830394 19.2284741280592
3.26469429316189 17.4095435734751
3.30420779801985 16.0013353068845
3.3437213028778 14.7408656362331
3.38323480773576 13.4491059895709
3.42274831259371 12.0263609764696
3.46226181745167 10.4591849172424
3.50177532230962 8.827953091161
3.54128882716758 7.27684803945974
3.58080233202553 5.94046241338
3.62031583688349 4.87480923804914
3.65982934174144 4.04783525948068
3.6993428465994 3.39014740283716
3.73885635145735 2.85033769758449
3.77836985631531 2.40542775127876
3.81788336117326 2.03442492181491
3.85739686603122 1.70077570994352
};
\addplot [very thick, opacity=0.75, gray119, forget plot]
table {%
-4.31828506101976 89.7185109723821
-4.27782329048492 113.743216238214
-4.23736151995008 141.107582958252
-4.19689974941524 171.571558621507
-4.1564379788804 204.764026278394
-4.11597620834556 240.198305853442
-4.07551443781072 277.297795333225
-4.03505266727588 315.434597904252
-3.99459089674104 353.982216954635
-3.9541291262062 392.378881330449
-3.91366735567136 430.192041342816
-3.87320558513651 467.169541291244
-3.83274381460167 503.261698730576
-3.79228204406683 538.602808442424
-3.75182027353199 573.450266432635
-3.71135850299715 608.092128291432
-3.67089673246231 642.745377448999
-3.63043496192747 677.4731167332
-3.58997319139263 712.146349922338
-3.54951142085779 746.464754193606
-3.50904965032295 780.033772283147
-3.46858787978811 812.477794178924
-3.42812610925327 843.557137012274
-3.38766433871843 873.254442592361
-3.34720256818358 901.805154478623
-3.30674079764874 929.664088271924
-3.2662790271139 957.419826030966
-3.22581725657906 985.683854876402
-3.18535548604422 1014.98684769584
-3.14489371550938 1045.70880579778
-3.10443194497454 1078.05573972384
-3.0639701744397 1112.07892036453
-3.02350840390486 1147.71979141794
-2.98304663337002 1184.85872006808
-2.94258486283518 1223.34969024612
-2.90212309230034 1263.03313093232
-2.8616613217655 1303.73055310172
-2.82119955123065 1345.23279032967
-2.78073778069581 1387.29554643802
-2.74027601016097 1429.65162476968
-2.69981423962613 1472.04113033158
-2.65935246909129 1514.25275507216
-2.61889069855645 1556.16413873689
-2.57842892802161 1597.76869308667
-2.53796715748677 1639.17968680541
-2.49750538695193 1680.60807134005
-2.45704361641709 1722.31660181385
-2.41658184588225 1764.55802709268
-2.37612007534741 1807.50893588301
-2.33565830481257 1851.2128895789
-2.29519653427772 1895.54611516651
-2.25473476374288 1940.21549136986
-2.21427299320804 1984.79165381161
-2.1738112226732 2028.77100533829
-2.13334945213836 2071.65203493785
-2.09288768160352 2113.00701537876
-2.05242591106868 2152.53217045795
-2.01196414053384 2190.06751499602
-1.971502369999 2225.58865619208
-1.93104059946416 2259.18218441446
-1.89057882892932 2291.01979627126
-1.85011705839448 2321.34259402469
-1.80965528785964 2350.45813065917
-1.7691935173248 2378.74319516961
-1.72873174678995 2406.63978082696
-1.68826997625511 2434.63286258226
-1.64780820572027 2463.20601798261
-1.60734643518543 2492.7811723119
-1.56688466465059 2523.65719656022
-1.52642289411575 2555.96492608576
-1.48596112358091 2589.65202307055
-1.44549935304607 2624.50157084798
-1.40503758251123 2660.17725416861
-1.36457581197639 2696.28004949923
-1.32411404144155 2732.39991589622
-1.28365227090671 2768.15167765935
-1.24319050037187 2803.19451259136
-1.20272872983702 2837.24426240777
-1.16226695930218 2870.09203286855
-1.12180518876734 2901.63844877459
-1.0813434182325 2931.94169133318
-1.04088164769766 2961.26417065152
-1.00041987716282 2990.09419960425
-0.95995810662798 3019.1206089794
-0.919496336093139 3049.15079721735
-0.879034565558298 3080.98200246265
-0.838572795023457 3115.25395363645
-0.798111024488616 3152.32086849734
-0.757649253953776 3192.1778547643
-0.717187483418935 3234.46211497814
-0.676725712884094 3278.52830007111
-0.636263942349253 3323.57703005704
-0.595802171814413 3368.8020618227
-0.555340401279572 3413.51804775337
-0.514878630744731 3457.23740126533
-0.474416860209891 3499.67918886275
-0.43395508967505 3540.71143221613
-0.393493319140209 3580.24614043932
-0.353031548605368 3618.11910563854
-0.312569778070527 3653.99030166999
-0.272108007535687 3687.2943686192
-0.231646237000846 3717.25620786535
-0.191184466466005 3742.96904153943
-0.150722695931164 3763.51710077086
-0.110260925396323 3778.11637609896
-0.0697991548614825 3786.24535662413
-0.0293373843266425 3787.74144605848
0.0111243862081984 3782.84512352505
0.0515861567430393 3772.18155656448
0.0920479272778802 3756.67862475493
0.132509697812721 3737.43181089757
0.172971468347562 3715.53905224468
0.213433238882402 3691.93873090664
0.253895009417243 3667.2866097051
0.294356779952084 3641.89942807378
0.334818550486925 3615.77499929813
0.375280321021766 3588.67657955778
0.415742091556607 3560.25112010238
0.456203862091447 3530.14378006286
0.496665632626287 3498.07745697411
0.537127403161128 3463.88319418422
0.577589173695969 3427.4879178978
0.61805094423081 3388.88156259442
0.658512714765651 3348.09011716276
0.698974485300492 3305.17323451379
0.739436255835332 3260.24897062294
0.779898026370173 3213.53167439673
0.820359796905014 3165.35981582084
0.860821567439855 3116.19288742388
0.901283337974696 3066.56925179263
0.941745108509537 3017.03392051405
0.982206879044377 2968.05882251296
1.02266864957922 2919.98207164816
1.06313042011406 2872.98552864772
1.1035921906489 2827.11500329882
1.14405396118374 2782.33133377725
1.18451573171858 2738.56996472611
1.22497750225342 2695.78544444459
1.26543927278826 2653.9652694945
1.3059010433231 2613.11084111288
1.34636281385794 2573.19631800498
1.38682458439278 2534.12390379139
1.42728635492763 2495.69427050091
1.46774812546247 2457.60432150647
1.50820989599731 2419.47467649217
1.54867166653215 2380.9000790591
1.58913343706699 2341.51020381788
1.62959520760183 2301.0268750384
1.67005697813667 2259.30559493883
1.71051874867151 2216.35298693239
1.75098051920635 2172.31619595277
1.79144228974119 2127.44516388224
1.83190406027603 2082.03406873966
1.87236583081087 2036.35366135453
1.91282760134571 1990.59035995335
1.95328937188056 1944.80871076986
1.9937511424154 1898.94952399495
2.03421291295024 1852.86663260318
2.07467468348508 1806.39311004868
2.11513645401992 1759.41710581321
2.15559822455476 1711.94268331656
2.1960599950896 1664.11485724389
2.23652176562444 1616.19979462095
2.27698353615928 1568.52678372669
2.31744530669412 1521.41205361839
2.35790707722896 1475.09040595865
2.3983688477638 1429.67649462293
2.43883061829864 1385.16524624024
2.47929238883349 1341.46562083496
2.51975415936833 1298.45008645305
2.56021592990317 1255.99853561169
2.60067770043801 1214.02083894811
2.64113947097285 1172.45369216748
2.68160124150769 1131.23932440011
2.72206301204253 1090.30074607522
2.76252478257737 1049.52796383633
2.80298655311221 1008.78302691969
2.84344832364705 967.922617448109
2.88391009418189 926.829440701477
2.92437186471673 885.440690997219
2.96483363525157 843.763681177222
3.00529540578642 801.873621892986
3.04575717632126 759.894103844074
3.0862189468561 717.965231708566
3.12668071739094 676.206898122805
3.16714248792578 634.6855959654
3.20760425846062 593.392905955254
3.24806602899546 552.242494992122
3.2885277995303 511.089840513733
3.32898957006514 469.774668564182
3.36945134059998 428.180533371894
3.40991311113482 386.300252552781
3.45037488166966 344.291939713503
3.4908366522045 302.510038104127
3.53129842273935 261.500006078001
3.57176019327419 221.95347808675
3.61222196380903 184.630567717921
3.65268373434387 150.264414748469
3.69314550487871 119.46741491378
3.73360727541355 92.6575655280813
};
\end{axis}

\end{tikzpicture}

%% file: nembdim_accuracy_paper_vertical.tex
\begin{tikzpicture}

\definecolor{chocolate2267451}{RGB}{226,74,51}
\definecolor{darkgray155}{RGB}{155,155,155}
\definecolor{darkorange2551094}{RGB}{255,109,4}
\definecolor{dimgray85}{RGB}{85,85,85}
\definecolor{gainsboro229}{RGB}{239,239,239}
\definecolor{gainsboro247}{RGB}{253,253,253}
\definecolor{gray119}{RGB}{119,119,119}
\definecolor{lightgray204}{RGB}{204,204,204}
\definecolor{mediumpurple152142213}{RGB}{152,142,213}
\definecolor{sandybrown25119394}{RGB}{251,193,94}
\definecolor{steelblue52138189}{RGB}{52,138,189}
\definecolor{yellowgreen14218666}{RGB}{142,186,66}

\begin{axis}[
title style={at={(0.5,0.075)},anchor=north},
x label style={at={(axis description cs:1.08, -0.04)}},
width=8.75cm,
height=3cm,
axis background/.style={fill=gainsboro247},
axis line style={white},
log basis x={10},
tick align=outside,
tick pos=left,
x grid style={gainsboro229},
y grid style={gainsboro229},
xmajorgrids,
xmin=63.0957344480193, xmax=15848931.9246111,
xmode=log,
xtick style={color=dimgray85},
ymajorgrids,
ytick style={color=dimgray85},
xtick={100, 1000,10000,100000,300000,1000000,3000000,10000000},
xticklabels={\tiny $\text{100}$, \tiny $\text{1,000}$, \tiny $\text{10\rm{K}}$, \tiny $\text{100\rm{K}}$, \tiny $\text{300\rm{K}}$, \tiny $\text{1\rm{M}}$, \tiny $\text{3\rm{M}}$, \tiny $\text{10\rm{M}}$},
ylabel=\textcolor{dimgray85}{\footnotesize Acc. (\%)},
ytick={62, 63, 64, 65, 66},
yticklabels={\scriptsize 62.0, \scriptsize 63.0, \scriptsize 64.0, \scriptsize 65.0, \scriptsize 66.0},
y label style={at={(axis description cs:0.06, 0.45)}},
yticklabel style = {xshift=0.5ex},
xticklabel style = {yshift=0.5ex},
ticklabel style={font=\tiny},
tick align=inside,
axis line style={gainsboro229},
xtick style={draw=none},
ytick style={draw=none},
xtick={1,2,3,10,100},
xmin=0.8, xmax=125,
xticklabels={1,2,3,10,100},
ymin=61.5, ymax=66.5,
]
\addplot [very thick, darkorange2551094, mark=*, mark size=1.25, mark options={solid}, opacity=0.4, fill opacity=0.8]
table {%
1 64.59
2 65.77
3 65.10
10 63.27
100 62.20
};
\end{axis}

\end{tikzpicture}

%% file: nembdim_runtime_paper_vertical.tex
\begin{tikzpicture}

\definecolor{chocolate2267451}{RGB}{226,74,51}
\definecolor{darkgray155}{RGB}{155,155,155}
\definecolor{darkorange2551094}{RGB}{255,109,4}
\definecolor{dimgray85}{RGB}{85,85,85}
\definecolor{gainsboro229}{RGB}{239,239,239}
\definecolor{gainsboro247}{RGB}{253,253,253}
\definecolor{gray119}{RGB}{119,119,119}
\definecolor{lightgray204}{RGB}{204,204,204}
\definecolor{mediumpurple152142213}{RGB}{152,142,213}
\definecolor{sandybrown25119394}{RGB}{251,193,94}
\definecolor{steelblue52138189}{RGB}{52,138,189}
\definecolor{yellowgreen14218666}{RGB}{142,186,66}

\begin{axis}[
xlabel=\textcolor{dimgray85}{\footnotesize Number of Subspace Sampling Dimensions ($m$)},
title style={at={(0.5,0.075)},anchor=north},
width=8.75cm,
height=4cm,
axis background/.style={fill=gainsboro247},
axis line style={white},
log basis x={10},
tick align=outside,
tick pos=left,
x grid style={gainsboro229},
y grid style={gainsboro229},
xmajorgrids,
xmode=log,
ymode=log,
xtick style={color=dimgray85},
ymajorgrids,
ytick style={color=dimgray85},
ylabel=\textcolor{dimgray85}{\footnotesize Runtime (\rm{s})},
yticklabel style = {xshift=0.5ex},
xticklabel style = {yshift=0.5ex},
ticklabel style={font=\tiny},
tick align=inside,
axis line style={gainsboro229},
xtick style={draw=none},
ytick style={draw=none},
y label style={at={(axis description cs:0.06, 0.5)}},
x label style={at={(axis description cs:0.5, 0.125)}},
xtick={1,2,3,10,100},
ymin=250, ymax=18000,
xmin=0.8, xmax=125,
xticklabels={1,2,3,10,100},
ytick={250, 381.6, 570,1485,14235},
yticklabels={\scriptsize 250, \scriptsize 380, \scriptsize 570, \scriptsize $\text{1,485}$, \scriptsize $\text{14,235}$},
]
\addplot [very thick, darkgray155, mark=*, mark size=0.75, mark options={solid}, opacity=0.8, fill opacity=0.8]
table {%
1 345
2 382
3 570
10 1485
100 14235
};
\end{axis}

\end{tikzpicture}